\documentclass[lettersize,journal]{IEEEtran}

\usepackage[table, rgb]{xcolor}
\usepackage{amsmath,amsfonts}
\usepackage{algorithmic}
\usepackage{algorithm}
\usepackage{array}
\usepackage{textcomp}
\usepackage{stfloats}
\usepackage{url}
\usepackage{verbatim}
\usepackage{graphicx}
\usepackage{cite}
\usepackage{todonotes}
\usepackage{booktabs}
\usepackage{balance}
\usepackage{multirow}
\usepackage{color, colortbl}
\usepackage{subcaption} 
\usepackage{hyperref}
\usepackage{float}

\definecolor{mycellcolor}{rgb}{0.976, 0.906, 0.890}

\newcommand*\circled[1]{\tikz[baseline=(char.base)]{
            \node[shape=circle,draw,inner sep=0.1pt, line width=0.80pt] (char) {#1};}}

\newcommand{\XXX}{{\sc DriftLens}} 

\begin{document}

\title{Unsupervised Concept Drift Detection \\ from Deep Learning Representations in Real-time}


\author{Salvatore Greco, Bartolomeo Vacchetti, Daniele Apiletti, Tania Cerquitelli

}



\maketitle

\begin{abstract}
Concept drift is the phenomenon in which the underlying data distributions and statistical properties of a target domain change over time, leading to a degradation in model performance. Consequently, production models require continuous drift detection monitoring.
Most drift detection methods to date are supervised, relying on ground-truth labels. However, they are inapplicable in many real-world scenarios, as true labels are often unavailable. Although recent efforts have proposed unsupervised drift detectors, many lack the accuracy required for reliable detection or are too computationally intensive for real-time use in high-dimensional, large-scale production environments. Moreover, they often fail to characterize or explain drift effectively.

To address these limitations, we propose \XXX, an unsupervised framework for real-time concept drift detection and characterization. Designed for deep learning classifiers handling unstructured data, \XXX\ leverages distribution distances in deep learning representations to enable efficient and accurate detection.
Additionally, it characterizes drift by analyzing and explaining its impact on 
each label.
Our evaluation across classifiers and data-types demonstrates that \XXX\ (i) outperforms previous methods in detecting drift 
in 15/17 use cases; (ii) runs at least 5 times faster; (iii) produces drift curves that align closely with actual drift (correlation $\geq\!0.85$); 
(iv) effectively identifies representative drift samples as explanations. 
\end{abstract}

\vspace{-1mm}

\vspace{-1mm}
\smallskip
\noindent \textbf{Code:} \url{https://github.com/grecosalvatore/drift-lens}

\section{Introduction}
\label{sec:intro}
The basic assumption in deep learning is that training data mimics the real world. Yet, deep learning models are typically trained and evaluated on static datasets. However, the world is dynamic, and what the model learned during training may no longer be valid. The underlying data distribution and statistical properties of the target domain may change over time, leading to a degradation of the model's performance (or decay). 

As we will formally define in \S\ref{sec:problem-formulation}, 
changes in data distribution—known as \textit{concept drift}~\cite{learning-under-concept-drift-survey}—can compromise the reliability of deep learning models in real-world applications~\cite{wang2022measure}.  
To mitigate this, continuous monitoring of production models is crucial~\cite{10.14778/3565838.3565853}. Effective monitoring should not only provide \textit{early warnings} when drift occurs but also \textit{characterize} and \textit{explain} drift to assist humans in implementing \textit{adaptive measures} to maintain model performance on evolving data.

As we will discuss in \S\ref{sec:related-work}, a large body of research has focused on concept drift detection through \textit{supervised} methods, which rely on error rates or performance-based measures computed from ground-truth labels~\cite{learning-under-concept-drift-survey}. However, they exhibit limited applicability in many real-world applications where true labels are unavailable for newly processed data.

A parallel research effort has been devoted to \textit{unsupervised} concept drift detection~\cite{hinder2023things,hinder2023things-partb,An-overview-of-unsupervised-drift-detection-methods-survey,10.1007/978-3-031-20738-9_121}. 
Most methods compare the data stream with reference data using distribution distances, divergence measures, or statistical hypothesis tests. However, they are computationally intensive and
ineffective for real-time concept drift detection in deep learning models processing large-scale, high-dimensional, unstructured, and unlabeled data streams~\cite{werner2023examining}. In addition, they typically fail to characterize and explain drift, which is crucial to facilitate drift adaptation. 

In tackling this, the contribution of this paper is twofold:

\noindent (1) We propose \XXX~(\S\ref{sec:drift-lens}), a novel \textit{unsupervised} drift detection framework designed to detect \textit{whether} and \textit{when} drift occurs by exploiting distribution distances of deep learning representations from unstructured data, modeled as Gaussian distributions. In addition, \XXX\ performs \textit{drift characterization} by determining and \textit{explaining} the drift impact on each label. Thanks to its low complexity, it enables \textit{real-time} drift detection, regardless of data dimensionality or~volume.

\noindent
(2) We conduct a comprehensive evaluation on several deep learning classifiers for text, image, and audio~(\S\ref{sec:evaluation}), and we attempt to answer the following four research questions (RQs): 
\noindent\textbf{(RQ1)} \textit{To what extent can it detect drift of varying severity without relying on true labels?} \\
\noindent\textbf{(RQ2)} \textit{How can it be applied broadly and effectively across various data types, models, and classification tasks?} \\
\noindent\textbf{(RQ3)} \textit{How efficient is it at detecting  drift in near real-time?} \\
\noindent\textbf{(RQ4)} \textit{To what extent can it accurately model, characterize, and explain the presence of drift over time?}

\noindent
We found that \XXX\ (i) outperforms previous drift detectors in 15/17 use cases; (ii) is extremely fast, enabling real-time drift detection independent of data volumes ($\leq0.2$ seconds, at least 5 times faster than other detectors); (iii)~produces drift curves highly correlated with drift severity; (iv)~effectively identifies representative drift samples as~explanations.

We first review concept drift (\S\ref{sec:problem-formulation}) and existing methods (\S\ref{sec:related-work}), then discuss \XXX\  (\S\ref{sec:drift-lens}) and its evaluation~(\S\ref{sec:evaluation}).


\section{Problem Formulation}
\label{sec:problem-formulation}
Here, we define concept drift (\S \ref{subsec:concept-drift-definition}), possible drift patterns (\S \ref{subsec:concept-drift-patterns}), and our application scenario and challenges~(\S \ref{subsec:application-scenario}).

\subsection{Concept Drift Definition}
\label{subsec:concept-drift-definition}
Concept drift can be defined as the phenomenon in which the underlying data distributions and statistical properties of a target data domain change over time. Drift can significantly affect the performance of deep learning models deployed in real-world applications. Various terms have been proposed to refer to \textit{``concept drift''} \cite{10.1016/j.knosys.2022.108632}, such as data drift, dataset shift, covariate shift, prior probability shift, and concept shift.
Although each definition emphasizes a particular facet of drift, most works, similar to ours, broadly refer to all subcategories under the term \textit{``concept drift''}.  Formally, concept drift is defined as a change in the joint distribution between a time period $[0,t]$ and a time window $t\!+\!w$. Drift occurs in $t\!+\!w$ if: 
\begin{equation}
\small
    P_{[0,t]}(X,y) \neq P_{t+w}(X,y)
\end{equation}
Where $X$ and $y$ are the feature vectors and the target variable of each data instance $(x_i,y_i)$, and $P_t(X,y)$ is the joint probability. 
The time window $t+w$ can be defined as a period or instant based on how the data stream is processed.
The joint probability can be further decomposed as:
\begin{equation}
\small
    P_t(X,y) = P_t(y/X)P_t(X) = P_t(X/y)P_t(y) 
\end{equation}
Where, $P_t(X/y)$ is the class-conditional probability, $P_t(y/X)$ is the target labels posterior probability, $P_t(X)$  is the input data prior probability, and $P_t(y)$ is target labels prior probability. 
In classification tasks, concept drift can occur as a change in any of these terms: 
(1) $P(X)$: A drift in the input data. The marginal probability of the input features $X$ changes (also known as data drift, covariance drift, or virtual drift). 
(2) $P(y/X)$: The relationships or conditional probabilities of target labels given input features change, but the input features do not necessarily change (usually referred to as concept or real drift). 
(3) $P(y)$: A change in the output data---labels and their probabilities change (usually referred to as label drift).

This work focuses on detecting drift in scenarios where no ground truth labels are available for new data. Therefore, drift must be detected in an \textit{unsupervised} manner. Due to the lack of actual labels, the only possible drift that can be considered is the change in the prior probability of the features $P(X)$~\cite{hinder2023things}.

\subsection{Drift Patterns}
\label{subsec:concept-drift-patterns}
\begin{figure}
  \centering \includegraphics[width=0.49\textwidth]{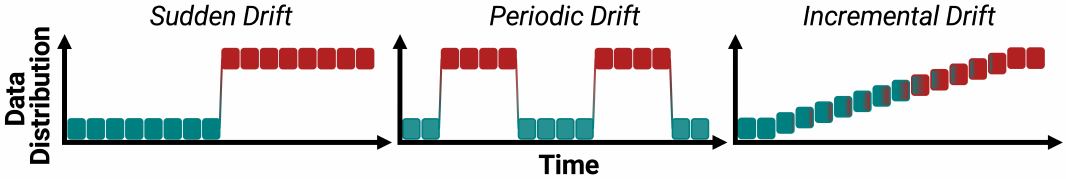}
  \caption{\textbf{Examples of drift patterns.}}
  \label{fig:drift-patterns}
\end{figure}

Concept drift can occur in several patterns (Figure \ref{fig:drift-patterns}):

\smallskip
\noindent
\textbf{Sudden/Abrupt Pattern} \ 
Drift can occur suddenly if the distribution of new samples changes rapidly. An example of sudden drift is a tweet topic classifier during the outbreak of the COVID-19 pandemic. At that point, the model was suddenly exposed to numerous text samples containing a new topic. Recognizing this scenario is crucial for initiating the retraining process and restoring the classifier's performance. 

\smallskip
\noindent
\textbf{Periodic/Recurrent Pattern} \ 
Drift occurs repeatedly after the first observed event, with a seasonality unknown during training. For example, in an election year, the language and topics of discussion on social media can change significantly, affecting the sentiment and context in which certain words or phrases are used. After the election, discussions return to normal, but during another major event, such as another election, they may change again. Another example of recurring distribution changes is sporting events like the Olympics.

\smallskip
\noindent
\textbf{Incremental Pattern} \ 
The transition between concepts occurs gradually over time. For example, in image or object classification for autonomous vehicles, the model may encounter new vehicle types, such as scooters (i.e., Personal Light Electric Vehicles), that were not part of the original training data. Initially, these new vehicles may appear rarely and sporadically. However, they gradually become more commonplace, eventually becoming a permanent feature of the streetscape.

\subsection{Application Scenario}
\label{subsec:application-scenario}
We design \XXX\ so that it can be effectively exploited in a variety of application scenarios characterized by: 

\noindent
(1) \textit{Unavailability of ground truth labels for new incoming data}. Many applications involve classifying large-scale data productions (e.g., social media content) through deep learning models. Usually, actual labels for newly processed samples are unavailable. Consequently, both drift detection and adaptation must be performed in an unsupervised manner.

(2) \textit{High data complexity and dimensionality}. Most applications use unstructured data like texts, images, and audio, which are characterized by high dimensionality and sparsity, complexity, and the lack of a fixed structure, such as columns. These factors can undermine the effectiveness and increase the complexity of drift detection and characterization techniques.

In such a target scenario, our desiderata for the drift detection method are the following:
(1) \textit{Fast detection}.
 Drift should be detected as soon as possible and not only when it occurs with high severity;
(2) \textit{Real-time detection}.
 Drift detection should have low complexity to be performed in real-time;
(3)~\textit{Drift characterization}.
 Information on drift patterns, the labels most affected by 
 changes, and an explanation should be provided. \XXX\ addresses all these desiderata.

\section{Related Works}
\label{sec:related-work}
There has been a significant effort in the field of concept drift detection \cite{learning-under-concept-drift-survey,An-overview-of-unsupervised-drift-detection-methods-survey,https://doi.org/10.1002/widm.1327,10.1016/j.knosys.2022.108632,Wang2024,10.14778/3594512.3594517}. Drift detection techniques can be classified into two macro-categories based on the true labels availability assumption: (1) \textit{supervised} and (2) \textit{unsupervised}. 

\smallskip
\noindent
\textbf{(1) Supervised Concept Drift Methods} \ Most of the previous drift detection techniques are \textit{supervised}~\cite{DDM,EDDM,LLDD,HDDM,FW-DDM,DELM,ADWIN,Arora2023,4053163,optAdwin,EnsembleDriftAdapt,9892066, SAND,10.1007/s10618-019-00656-w}.  
These methods typically rely on error rate-based measures or ensemble models to assess degradation in the error
rate over time (e.g., decrease in accuracy) \cite{9729470}.
However, they assume that the true labels are available with the new data or within a short period of time. In practice, the labels for new data are usually unavailable, and labeling them is very costly and time-consuming, limiting their applicability in real-world scenarios.

\smallskip
\noindent
\textbf{(2) Unsupervised Concept Drift Methods} \ In contrast, \textit{unsupervised} techniques do not require true labels~\cite{hinder2023things,An-overview-of-unsupervised-drift-detection-methods-survey,10.1007/978-3-031-20738-9_121,werner2023examining}. Our method falls into this category. As outlined by~\cite{hinder2023things}, unsupervised techniques can be further divided into: 
(2.i)~\textit{statistical-based}, 
(2.ii)~\textit{loss-based}, and 
(2.iii)~\textit{virtual classifier-based}.

\smallskip
\noindent
\textbf{(2.i) Statistical-based Methods} \ Most of the unsupervised techniques~\cite{JMLR:v13:gretton12a,Kolmogorov-Smirnov-Test,7745962,10.1145/3350489.3350494,DBLP:conf/bigdata/VenturaPACPBMM19,UDetect,Plover, 10.5555/1316689.1316707, STEPD,Rabanser2018FailingLA} rely on statistical hypothesis tests, two-sample tests, or divergence metrics between distributions to detect drift,  such as Maximum Mean Discrepancy \cite{JMLR:v13:gretton12a}, Kolmogorov-Smirnov \cite{massey1951kolmogorov}, Least-Squares Density Difference~\cite{7745962}, or Cramér–von Mises \cite{cramer1928composition}.
These techniques have two main advantages. Firstly, they can be applied similarly independently of the data type. Secondly, they do not require external models or resources.
Their main limitation is that they are usually computationally intensive and ineffective in detecting drift affecting deep learning models applied to large-scale, high-dimensional, and unstructured data stream. This can affect their runtime and drift prediction performance~\cite{werner2023examining}.

\smallskip
\noindent
\textbf{(2.ii) Loss-based Methods} \ These techniques exploit model loss functions to evaluate the similarity of new data points with previous ones, such as \cite{7368802,10.14778/3407790.3407837,DBLP:journals/corr/abs-2001-06386,10.1145/775047.775148}. 
Usually, they rely on autoencoders or are used in conjunction with supervised drift detectors. They are based on the assumption of a correlation between the increase in model losses and concept drift \cite{inproceedingshard}. However, they have two main limitations. Firstly, they require an external resource or model to perform drift detection. Secondly, being usually based on autoencoders, they are mainly suitable for computer vision models. Therefore, they are not agnostic to the data domain, and their usage on other data types requires specific implementations and design choices.

\smallskip
\noindent
\textbf{(2.iii) Virtual Classifier-based Methods} \ 
Similarly, these techniques \cite{10.1145/3357384.3358144, inproceedingsldd, 10.1007/978-3-540-68125-0_15} use classifiers to detect drift. The general idea is to divide data into two sets, before and after a certain moment in time. If the classifier achieves an accuracy higher than random in classifying samples into the two classes, it implies divergent data properties between the two class distributions, suggesting the presence of drift. The main limitation is the need to train and maintain another model to detect drift, whose implementation is domain- and task-specific.

\smallskip
\noindent
\textbf{Drift Explanation} \ Some techniques attempt to provide human-readable explanations for drift rather than only detecting when it occurs \cite{HINDER2023126640,hinder2022contrasting,ADAMS2023102177,263854,hinder2023things-partb}. For instance, \cite{HINDER2023126640} proposes a framework that trains a proxy model to classify or segment drifted samples \cite{9892374,hinder2021concept} and then leverages Explainable AI (XAI) \cite{10.1145/3236009,molnar2020interpretable} to interpret drift through the proxy model. Other techniques provide feature-wise representations to interpret drift \cite{9308627,10.1145/956750.956849,10.1007/s10618-018-0554-1}. However, they usually struggle with high-dimensional or non-semantic features, like unstructured data. 
Despite its importance, research on drift explanations remains limited, especially for unstructured data where inputs lack clear semantics and are high-dimensional \cite{HINDER2023126640}. 

\smallskip
\noindent
\textbf{Drift Adaptation and Incremental Learning} \ Some works are related to incremental learning for drift detection and adaptation \cite{10.1145/2523813}. However, in unsupervised settings, the adaptation to drift is
challenging due to the lack of annotated samples~\cite{hinder2023things}.  Drift adaptation is out of the scope of this paper. Instead, we focus only on the drift detection problem (\textit{monitoring}). 

\smallskip
This paper presents \XXX, an unsupervised statistical-based drift detection technique. It significantly extends the preliminary idea presented in \cite{driftlens} in two key aspects: (1) \textit{Methodology}: It enhances the original approach by complementing the \textit{per-label} with a \textit{per-batch} analysis, improving its general applicability across tasks. 
Additionally, it redefines the threshold estimation and provides drift explanations to better characterize drift; 
(2)~\textit{Evaluation}: It demonstrates its general applicability and effectiveness through a comprehensive evaluation across multiple data types, models, and baseline detectors. \XXX\ is also available in a tool with a user-friendly graphical interface~\cite{driftlensdemo}. 

\XXX\ differs from previous work in three key aspects. (1) Like other statistical-based methods, it is completely \textit{unsupervised}. Thus, it does not require any external model to detect drift and is \textit{data type agnostic}, unlike supervised, loss-based, and virtual classifier methods. (2) It exploits statistical distances that better scale with data dimensionality than other statistical-based methods, thereby enabling efficient \textit{real-time} drift detection in large data volumes. (3) It \textit{characterizes} drift by assessing its impact on each label and providing prototype-based \textit{explanations} to improve human understanding of drift.

\section{DriftLens}
\label{sec:drift-lens}
\XXX\ is an \textit{unsupervised} drift detection technique based on distribution distances within embeddings---internal dense representations generated by deep learning models. Designed specifically for unstructured data, it captures subtle patterns and relationships in embeddings to identify data drifts over time. It operates in an \textit{unsupervised} way as it does not require true labels in the data stream. However, it is tailored to detect concept drift on supervised models, such as classifiers. 

\begin{figure}
  \centering \includegraphics[width=0.49\textwidth]{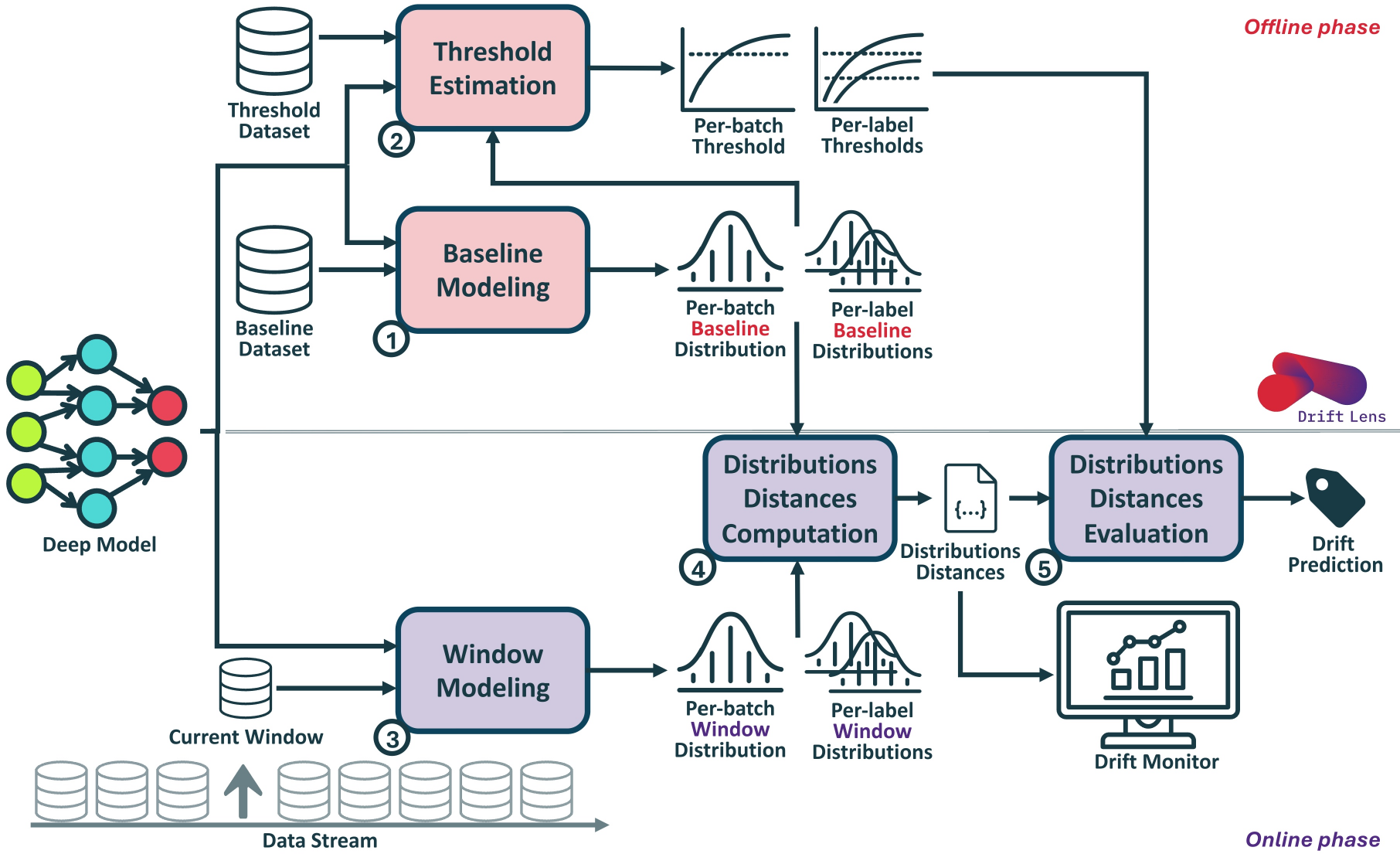}
\caption{\textbf{\XXX\ Framework.} In the \textit{offline} phase, it estimates the
reference distributions and distance thresholds from historical (training) data. Distributions are modeled as multivariate normal and are computed: (i) for the entire batch (\textit{per-batch}), and (ii) conditioned on the predicted label (\textit{per-label}). In the \textit{online} phase, it analyzes data streams in fixed windows, comparing new and reference distributions, and using thresholds to identify drift, visualized in a drift monitor.} 
  \label{fig:architecture}
\end{figure}

\XXX\ consists of an \textit{offline} and an \textit{online} phase (Figure~\ref{fig:architecture}). 
In the \textit{offline} phase, it estimates the \textit{reference distributions} and \textit{threshold values} from the historical dataset (e.g., training data). 
These distributions, called \textit{baseline}, represent the distribution of features (embeddings) of the concepts learned by the model during training, thus representing the absence of drift. 
In the \textit{online} phase, it processes the new data stream in fixed-size windows. Firstly, the distributions of the new data windows are estimated. Secondly, the distribution distances are computed with respect to the reference distributions. If the distance exceeds the threshold, a drift is predicted.

We first detail the data modeling (\S\ref{subsec:methodology-data-modeling}) and the distribution distance (\S\ref{subsec:methodology-distribution-distance}), performed similarly in both phases. We then describe the offline (\S\ref{subsec:methodology-offline-phase}) and online (\S\ref{subsec:methodology-online-phase}) phases. Finally, we discuss drift explanations and adaptation (\S\ref{subsec:methodology-drift-explanations}).

\subsection{Data Modeling: Embedding Distributions Estimation}
\label{subsec:methodology-data-modeling}
Consider a deep learning classifier designed to distinguish between a set of class labels $L$.
The classifier typically consists of an encoder $\phi(X)$ that transforms sparse, complex, and high-dimensional raw input data into a dense, lower-dimensional latent representation (embedding) through a series of nonlinear transformations parameterized by the learned weights $W$. 
The embedding representation is then processed by one or more fully connected layers to predict the final class label.

To simplify data distribution modeling, \XXX\ exploits the internal representations (embeddings) generated by the deep learning model instead of raw inputs. This reduces noise and complexity, especially for unstructured data types like images, text, and audio, which are often complex, noisy, and non-Gaussian. In contrast, embeddings are lower-dimensional and capture more structured information. 

\begin{figure}
  \centering \includegraphics[width=0.49\textwidth]{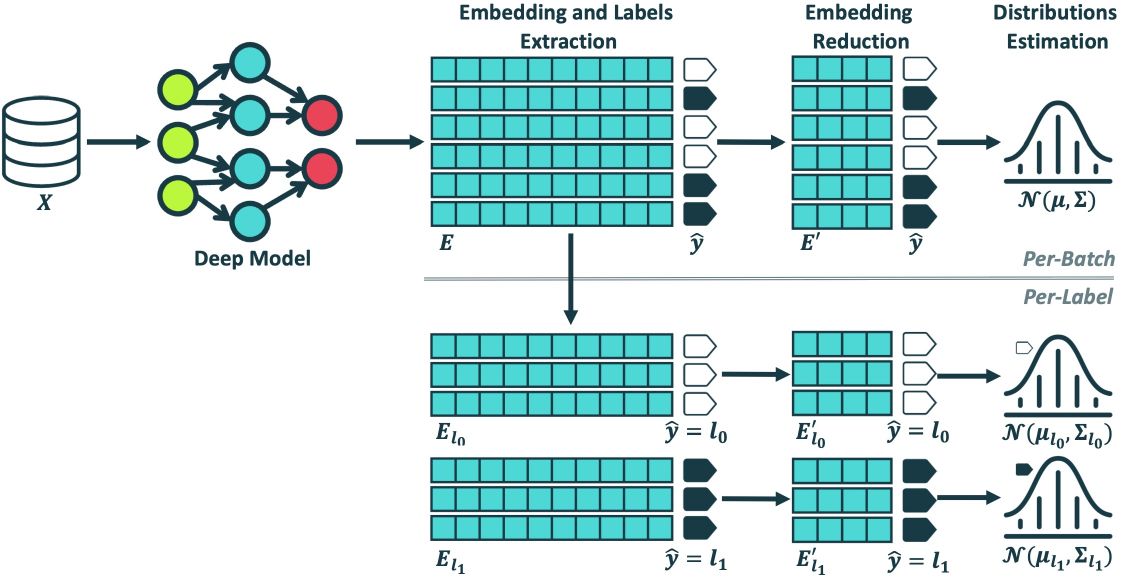}
  \caption{\textbf{\XXX\ Data Modeling.} Given a deep learning model and a set of data $X$, it estimates the multivariate normal embedding distributions. It first extracts data embeddings and predicted labels from the model and reduces the embedding dimensionality. It then estimates (i) the \textit{per-batch} distribution by computing the mean vector $\mu$ and covariance matrix $\Sigma$ over all embeddings, and (ii) label-specific distributions by grouping embeddings by label (e.g., $l_0$ and $l_1$), reducing the dimensionality, and computing the \textit{per-label} $\mu_l$~and~$\Sigma_l$,~$\forall l~\in~L$.}
  \label{fig:data-modeling}
\end{figure}

As shown in Figure~\ref{fig:data-modeling}, \XXX\ estimates two types of embedding distributions: (1) the overall distribution of the entire data (\textit{per-batch}), and (2) $|L|$ separate distributions, one for each predicted class (\textit{per-label}). It assumes that embeddings follow multivariate normal distributions, making Gaussianity a more plausible approximation given the simpler relationships in embedding space~\cite{6472238}. This approach is data type-independent, ensuring broad applicability in diverse domains. 

Formally, during data modeling, a batch of data $X$ (historical data or a data stream window) is fed into the deep learning model to extract the embedding matrix $E=\phi(X) \in \mathbb{R}^{m \times d}$ and predicted labels $\hat{y} \in \mathbb{R}^{m}$, where $m$ is the number of samples in the set and $d$ is the dimensionality of the embedding layer. 
Embeddings are also grouped by predicted label, resulting in $|L|$ matrices $E_l \in \mathbb{R}^{m_l \times d}, \forall l \in L$, where $m_l$ is the number of samples predicted with label~$l$.

In deep learning classifiers, the embedding space dimensionality $d$ is usually large. 
To estimate the multivariate normal distributions, we need to compute the mean vector $\mu$ and the covariance matrix $\Sigma$ of the embedding vectors.
However, the covariance matrix requires at least $d$ linearly independent vectors to be full rank.
Otherwise, it may contain complex numbers that could affect the distribution distance computation (see~\S\ref{subsec:methodology-distribution-distance}).
This issue is more pronounced when estimating the multivariate normal distribution for each label, as we need $d$ linearly independent vectors predicted with each label.
This is usually not an issue when modeling the baseline, which is computed over a usually large historical dataset. 
In contrast, in the online phase, estimating the per-batch distribution requires $d$ while the per-label distributions $d \times |L|$ linearly independent vectors in each data stream window.
Therefore, the fixed size used to divide the data stream into windows must be very large. To solve this, \XXX\ performs an embedding dimensionality reduction applying a principal component analysis (PCA).

\begin{equation}
\small
E' = PCA(E); \quad E'_{l} = PCA(E_{l}), \forall l \in L
\end{equation}

This leads to the reduced embedding matrices $E' \in \mathbb{R}^{m \times d'}$ and $E'_l \in \mathbb{R}^{m_l \times d'_l}, \forall l \in L$. $d'$ and $d'_l$ are user-defined parameters that determine the number of principal components for the per-batch and per-label reductions, with $0 < d', d'_l  \le d$. 
For the per-batch, $d'$ can be set to a value up to $d' = min(m_w, d)$, where $m_w$ is the window size used in the online phase and $d$ is the embedding dimensionality. Instead, $d'_l$ also depends on the number of labels.
A reasonable value for a balanced data stream is close to $d'_l = min(m_w/|L|, d)$.  

Finally, the reduced embedding matrices $E'$ and $E'_{l}$ are used to estimate the \textit{per-batch} and the $|L|$ \textit{per-label} multivariate normal distributions (equations~\ref{eq:normal-distribution-per-batch} and \ref{eq:normal-distribution-per-label}). 
\begin{equation}
\label{eq:normal-distribution-per-batch}
\small
P(\phi(x); W)\sim \mathcal{N}(\mu,\,\Sigma)
\end{equation}
\begin{equation}
\small
\label{eq:normal-distribution-per-label}
P(\phi(x) \ | \ \hat{y} = l; W)\sim \mathcal{N}(\mu_l,\,\Sigma_l), \forall l \in L
\end{equation}

Where $\mu \in \mathbb{R}^{d'}$ and $\Sigma \in \mathbb{R}^{d' \times d'}$ are the mean vector and the covariance matrix of the \textit{per-batch} multivariate normal distribution, and each $\mu_l \in \mathbb{R}^{d'_l}$ and $\Sigma_l \in \mathbb{R}^{d'_l \times d'_l}$ represents the \textit{per-label} multivariate normal distribution of each label $l \in L$, separately. The multivariate normal distributions are straightforward and fast to estimate because they can be fully characterized by the mean vector and the covariance matrix. The main advantage of estimating the distributions as multivariate normal is that they can be represented with dimensionality $d'$ and $d'_l$, regardless of the number of samples in the reference and new windows. This allows the method to scale well even with large amounts of data, enabling drift detection in real-time regardless of the data volumes.

\subsection{Distributions Distance Computation: Frechét Distance}
\label{subsec:methodology-distribution-distance}
\XXX\ uses the Frechét distance~\cite{DOWSON1982450} to calculate the distance between two multivariate normal distributions. The Frechét distance, also known as the Wasserstein-2 distance~\cite{wasserstein1969markov}, has been widely used in deep learning to measure the distances between the distributions of models' features but in very different scenarios~\cite{NIPS2017_8a1d6947}. \XXX\ uses the Frechét distance to measure the distances between the embedding distributions of a baseline (historical) and the new windows in the data stream, and we call it \textit{Frechét Drift Distance} ($FDD$). 

Given a multivariate reference normal distribution $b$ (e.g., the baseline computed in the \textit{offline} phase) characterized by a mean vector $\mu_b$ and a covariance matrix $\Sigma_b$, and the multivariate normal distribution of a new data window $w$, characterized by $\mu_w$ and $\Sigma_w$, the $FDD$ is calculated as: 
\begin{equation}
\label{eqn:FDD-score-batch}
\small
FDD(b,w) = ||\mu_b- \mu_w||^2_2 + Tr\bigg(\Sigma_b + \Sigma_w - 2\sqrt{\Sigma_b \Sigma_w}\bigg)
\end{equation}

$FDD\in [0, \infty]$. The higher the $FDD$, the greater the distance and the more likely the drift.
It captures changes in the mean (center) and diagonal elements of the covariance (spread) between two distributions. The former results from the L2 norm of the difference between the mean vectors $||\mu_b-\mu_w||^2_2$. The latter is computed by $Tr(\Sigma_b + \Sigma_w - 2\sqrt{\Sigma_b \Sigma_w})$, which is a generalization of the squared difference between the standard deviations in a one-dimensional space. 
As a result, $FDD$ can detect subtle drifts that affect both the center and spread of distributions.
Although \XXX\ uses the $FDD$ as default metric, it allows flexibility in its choice~\cite{kullback1951information,61115,noauthor_2018-zc,3142ae09-8e70-3b5c-a340-fa8eafc77ee5}. An ablation study of alternative metrics is presented in Appendix~\ref{apx:evaluation-distance-metric}.

\XXX\ computes a single \textit{per-batch} distance by computing the $FDD$ score between the baseline ($\mu_b$, $\Sigma_b$) and the new window $(\mu_w, \Sigma_w)$ \textit{per-batch} distributions, and $|L|$ \textit{per-label} distances by computing the distance of the distributions $(\mu_{b,l}, \Sigma_{b,l})$ and $(\mu_{w,l}, \Sigma_{w,l})$ for each label $l \in L$ separately.

\subsection{Offline Phase: Baseline and Threshold Estimation}
\label{subsec:methodology-offline-phase}
In the \textit{offline} phase (steps \circled{1} -- \circled{2} in Figure \ref{fig:architecture}), \XXX\ estimates: \circled{1} the probability distribution of a reference (historical) dataset representing the concepts learned by the model during training (\textit{baseline}), and \circled{2} distance \textit{thresholds} to distinguish between normal and 
drift-indicative distances, which will be used during the \textit{online} phase.
This phase is executed once to instantiate \XXX, after the monitored model has been trained, evaluated, and is ready for deployment.

\smallskip
\noindent
\textbf{Baseline Estimation} \ 
It estimates the reference distributions (distributions in the absence of drift) by performing the data modeling (\S\ref{subsec:methodology-data-modeling}) of the baseline (historical) dataset.

Formally, given a baseline dataset $X_b$, containing $m_b$ samples, the entire baseline dataset $X_b$ is fed into the deep learning model to extract the embedding vectors $E_b = \phi(X_b) \in \mathbb{R}^{m_b \times d}$ and estimate the predicted labels $\hat{y}_b$. Then, the \textit{per-batch} PCA is fitted over the entire set of vectors, and $|L|$ different PCAs are fitted, grouping the embedding vectors according to the predicted labels.
The embedding vectors are then reduced both for the \textit{per-batch} and \textit{per-label}, to obtain the reduced embedding matrices $E'_{b} \in \mathbb{R}^{m_b \times d'}$, and $E'_{b,l} \in \mathbb{R}^{m_{b,l} \times d'_l}, \forall l \in L$.
These matrices are used to estimate the baseline \textit{per-batch} and \textit{per-label} multivariate normal distributions. 
The \textit{per-batch} distribution is fully characterized by the baseline \textit{per-batch} mean vector $\mu_b \in \mathbb{R}^{d'}$ and the covariance matrix $\Sigma_b \in \mathbb{R}^{d' \times d'}$, which are obtained by calculating the mean and covariance over the entire set of reduced embedding vectors in the baseline dataset.
The \textit{per-label} distributions are obtained by computing the $|L|$ mean vectors $\mu_{b,l} \in \mathbb{R}^{d'_l}$ and the covariance matrices $\Sigma_{b,l} \in \mathbb{R}^{d'_l \times d'_l}$ on the reduced embedding vectors, grouped by predicted labels. Note that regardless of the dimensionality of the reference set (baseline) $m_b$, the estimated reference distributions are characterized by vectors of size $d'$ and $d'_l$. Thus, drift detection in new windows is not influenced by~$m_b$.

\smallskip
\noindent
\textbf{Threshold Estimation} \ It estimates the maximum possible distance ($FDD$) a window without drift can reach.
It takes in input a threshold dataset $X_{th}$, the window size $m_w$ (equal to the one 
used in the online phase), the baseline, and a parameter $n_{th}$ defining the number of windows randomly sampled from the threshold dataset. 
$n_{th}$ should be large to better estimate the maximum distance considered without drift. In our experiments (\S\ref{subsec:experimental-use-cases}), we set $n_{th}\!=\!10{,}000$, and show that its variations have a small impact (see Appendix~\ref{apx:evaluation-sensitivity}).

Specifically, \XXX\, randomly samples, with replacement, $n_{th}$ windows from the threshold dataset $X_{th}$, each containing~$m_{w}$ inputs. For each window, it performs the data modeling and computes the \textit{per-batch} and \textit{per-label} distribution distances from the baseline distributions (equation~\ref{eqn:FDD-score-batch}). Therefore, $n_{th}$ distribution distances for the entire batch and each label are computed. Finally, the distribution distances are sorted in descending order. 
The first element contains the maximum distance that a window of data considered without drift can achieve. Distances that exceed this value are potential warnings of drift. However, there are potentially outlier distances due to the large number of randomly sampled windows. Therefore, \XXX\ provides a parameter to define the threshold sensitivity $T_\alpha \in [0,1]$. This parameter removes the $T_\alpha\%$ left tail of the sorted distances (in descending order) to remove outliers. 
The final thresholds $T$ and $T_l, \forall l \in L,$ are set to the maximum distance after removing the $T_\alpha\%$ of the largest distances. The higher the value of $T_\alpha$, the lower the threshold values and the higher the sensitivity to possible drift and false alarm. In our experiments (see \S\ref{subsec:experimental-use-cases}), we use $T_\alpha = 0.01$ (removing $1\%$) as the default threshold sensitivity.

\smallskip
\noindent
\textbf{Choice of the Reference Dataset} \ 
The reference dataset must consist of \textit{historical} data representing what the model learned during training. Since we operate after training, we assume the availability of historical data used for training and testing, typically split into train/val/test or train/test sets. 
Training data can be used as the baseline to represent the concepts learned by the model, modeling the absence of drift. For the threshold estimation---maximum distribution distance without compromising model performance---a reasonable choice is the data used for evaluation (e.g., test or validation sets, or their combination). Alternatively, windows from the data stream can be used. Due to the absence of true labels, human experts must ensure that these windows represent distributions without drift.

\subsection{Online Phase: Drift Detection Over Time}
\label{subsec:methodology-online-phase}
The \textit{online} phase (steps \circled{3}–\circled{5} in Figure~\ref{fig:architecture}) continuously detects drift in the monitored model when deployed and applied to a data stream. \XXX\ splits the data stream into fixed-size windows, whose size is defined by the parameter~$m_w$, and applies the same process to each window.

 \smallskip
\noindent
\textbf{Drift Detection in a Window} \ Given a new data window $X_w$ containing $m_w$ samples, \circled{3} the data modeling is processed by (i) extracting the embedding $E_w =\phi(X_w) \in \mathbb{R}^{m_w \times d}$ and predicted labels $\hat{y}_w \in \mathbb{R}^{m_w}$, (ii) performing the embedding dimensionality reduction to obtain $E'_w \in \mathbb{R}^{m_w \times d'}$ and $E'_{w,l} \in \mathbb{R}^{m_{w,l} \times d'_l}, \forall l \in L$, and (iii) estimating the \textit{per-batch} and \textit{per-label} multivariate normal distributions. The PCA models fitted during the offline phase are reused in this step. The \textit{per-batch} multivariate normal distribution is obtained by computing the mean vector $\mu_w \in \mathbb{R}^{d'}$ and the covariance matrix $\Sigma_w \in \mathbb{R}^{d' \times d'}$ over the entire set of reduced embeddings in the window. 
For the \textit{per-label}, the reduced embedding vectors are grouped by predicted labels, and the mean and covariance are computed separately on each subset, resulting in $|L|$ multivariate normal distributions characterized by $\mu_{w,l} \in \mathbb{R}^{d'_l}$ and $\Sigma_{w,l} \in \mathbb{R}^{d'_l \times d'_l}$.
Next, \circled{4} the \textit{per-batch} and the \textit{per-label} $FDD$ distances between the window and baseline distributions are computed using equation \ref{eqn:FDD-score-batch}.  If these distances exceed the thresholds, drift is predicted \circled{5}. The \textit{per-batch} distance detects if the entire window is affected by drift, while the \textit{per-label} distance characterizes drift and identifies the labels most affected.

\smallskip
\noindent
\textbf{Drift Monitoring Over Time} \
Once \XXX\ processes a new window, the $FDD$ distribution distances are added to the drift monitor, which shows them separately for the \textit{per-batch} and \textit{per-label}. A warning symbol is added to the plot when distances exceed the corresponding threshold. The drift monitor provides valuable insights to understand (i) \textit{when}, \textit{whether} drift
occurs, and its \textit{severity}, (ii) \textit{which labels} are the most affected by drift, and (iii) the drift \textit{pattern} over time.

\begin{figure}
  \centering 
  \includegraphics[width=0.424\textwidth]{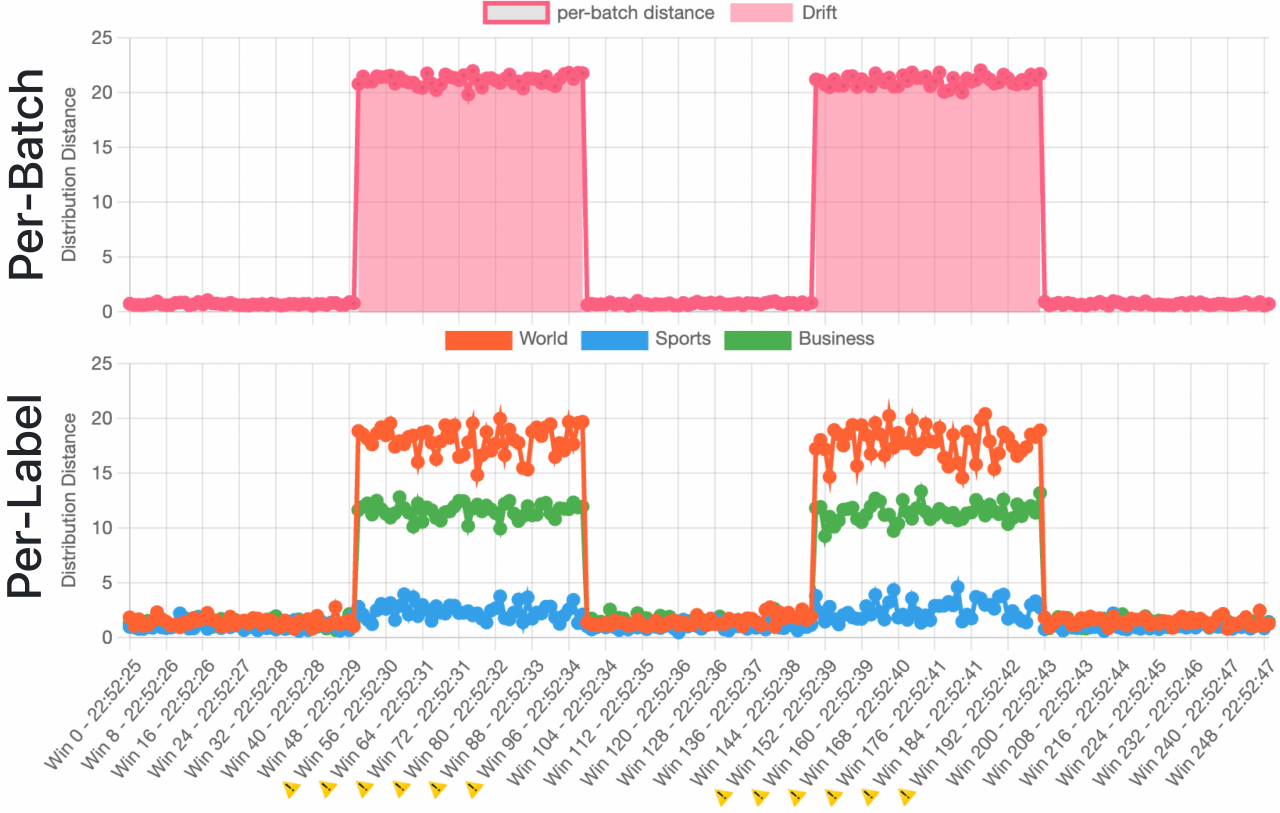}
  \caption{\textbf{Drift monitor example}} 
  \label{fig:monitor}
\end{figure}

Figure \ref{fig:monitor} shows a drift monitor example. 
The top chart shows the \textit{per-batch}, whereas the bottom chart shows the \textit{per-label} distribution distances over time. The \textit{x-axis} reports timestamps and window identifiers. The \textit{y-axis} shows the $FDD$ distances. When a drift is detected, the area under the curve is filled, and a warning is displayed on the x-tick. This monitor reveals that drift first occurs after 50 windows with high severity and reoccurs periodically. The label \textit{World} is most affected, followed by \textit{Business}, while \textit{Sports} is minimally affected.

\subsection{Drift Explanation, Characterization, and Adaptation}
\label{subsec:methodology-drift-explanations}
\smallskip
\noindent
\textbf{Drift Explanation} \ \XXX\ produces drift explanations to enhance human understanding of drift. Although the objective is related to XAI~\cite{molnar2020interpretable,10.1145/3236009}, it differs in focus. 
XAI aims to explain classifier predictions, whereas drift explainability aims
to explain the underlying causes of drift~\cite{HINDER2023126640}.
Since \XXX\ operates on classifiers for high-dimensional, high-volume unstructured data streams, without true labels, the explanation method must meet these requirements. 
(1)~\textit{Post-hoc}: It should apply to models already trained, without modifying the training process. (2)~\textit{Model-agnostic}: It should work with any model, regardless of its architecture.
(3)~\textit{Unsupervised}: It should not rely on true labels or predefined concepts. (4)~\textit{Data type-independent}: It should work uniformly across data types.

To meet these requirements, \XXX\ produces example-based explanations \cite{molnar2020interpretable}, which are intuitive to humans \cite{ijcai2019p876,10.1145/3287560.3287574}, including to understand drift~\cite{HINDER2023126640}. Drift explanations are generated on request for one or more drifted windows rather than continuously.
Drift explanations identify representative examples (\textit{prototypes}) of inputs processed by the model to illustrate the \textit{“concepts”} associated with each label and a batch of data. In the absence of drift, we expect these prototypes to closely resemble those in the historical data. In contrast, when drift is present, the drifting samples introduce new concepts specific to a label or the entire batch.

Given one or multiple concatenated windows, \XXX\ first exploits a clustering algorithm to group samples by embedding representations, considering the entire batch and each label separately. 
The objective of clustering is to group samples to isolate drifted samples into one or more pure clusters. To this end, it executes the K-Means algorithm multiple times, for $k \in [2, K_{\max}]$ on the embedding matrices $E$ and $E_l, \forall l \in L$, where $K_{\max}$ is a user-defined parameter. 
The optimal clustering partition---the best $k$---is determined by maximizing the Silhouette metric, 
and may differ between the per-batch and each label.  For a given label or batch, the K-Means 
produces $k$ clusters, each representing a distinct concept relevant to classification. The samples nearest to the centroids of these clusters serve as representative examples (\textit{prototypes}), effectively summarizing the primary characteristics of the group. To improve prototype interpretability, a user-defined parameter, $\text{top-}n$, is introduced, specifying the $n$ closest samples to each centroid based on the 
Euclidean distance of the embeddings.  
This 
enhances interpretability by allowing users to observe multiple representative samples per cluster. Examining a single example may not be sufficient to understand the common features, whereas visualizing multiple similar instances provides more comprehensive insight into the shared characteristics within each cluster.  Prototypes provide valuable insight to interpret the causes of drift (see \S\ref{subsec:evaluation-explanations}). 

\smallskip
\noindent
\textbf{Drift Characterization and Adaptation} \ In unsupervised data streams, the absence of annotated samples makes automatic drift adaptation challenging \cite{hinder2023things}. As a result, as with most unsupervised detectors, \XXX\ does not automatically adapt the model when drift is detected. Instead, human-in-the-loop is required for adaptation, necessitating the annotation of new data, or at least a subset, by domain experts (or an oracle). 
However,  drift characterization enables human experts to gain deeper insight into drift and take informed adaptive measures.

(1) The \textit{per-label drift analysis} helps users focus on the labels most affected by drift, simplifying both the understanding and annotation process. By narrowing the scope of analysis, it reduces complexity and enables users to direct their efforts where they are most needed. This enhances the interpretability of drift patterns and reduces the annotation workload, making it easier to adapt models to evolving data distributions.

(2) The \textit{prototype-based drift explanations} help users better understand the nature of drift and how it affects each label and the entire batch. These prototype examples can reveal whether a new class label should be introduced due to the emergence of concepts that differ significantly from those in the training data. Alternatively, they may suggest that the meaning of an existing label is evolving over time, indicating the need for further training to maintain accuracy. 

(3) The \textit{drift monitor over time} can reveal whether the drift is stabilizing, progressively increasing, or recurring. 
(i)~If drift occurs \textit{suddenly and remains constant}, it indicates that the deployed model is outdated, with new distributions consistently replacing the previous ones. 
Therefore, the model should be updated or replaced by fine-tuning it on the new distributions or retraining it from scratch, incorporating historical and new data. (ii) If drift occurs with a \textit{periodic} pattern, two strategies are possible. The model can be retrained to perform well across both distributions (similar to sudden drift), or an additional model can be trained and deployed alongside the original. When drift occurs, the outdated model can be swapped out for the one best suited to the new distribution. Once data return to their normal state, the system can switch back to the original model. (iii) If drift increases \textit{incrementally}, the model can be gradually fine-tuned to the new distributions, or retraining can be postponed until drift stabilizes.

Once the model is adapted (or retrained) to the new incoming distributions, \XXX\ can be updated by simply re-executing the offline phase to reset the baseline and threshold.

\begin{table*}[]
\centering
\scriptsize
\caption{\textbf{Overview of the experimental use cases.} Use cases are partitioned into groups based on the dataset and task. The training labels, the way the drift is simulated, and the split of the dataset are given in the description for each group. Within each group, different deep learning models are considered, and the corresponding F1 scores obtained on the test set are given. }
\label{tab_uses_cases}
\scalebox{.845}{
\begin{tabular}{cccclcl}
\toprule
\textbf{Data Type}  & \multicolumn{1}{c}{\textbf{Dataset}}   & \textbf{Task}        & \multicolumn{1}{c}{\textbf{Use Case}} & \multicolumn{1}{l}{\textbf{Models}} & \multicolumn{1}{c}{\textbf{F1}}  & \multicolumn{1}{c}{\textbf{Description}}  \\ \toprule
Text &  \multicolumn{1}{c}{Ag News}  & Topic  & 1.1\phantom{*} & BERT & 0.98 & \multirow{2}{*}{\begin{tabular}[c]{@{}l@{}}\textbf{Training labels}:  \textit{World}, \textit{Business}, and \textit{Sport}\\ \textbf{Drift}: Simulated with one new class label: \textit{Science/Technology}\end{tabular}} \\
 & \multirow{2}{*}{\cite{ag-news}}  & Detection & 1.2* & DistilBERT & 0.97 &  \\
 &  &  & 1.3* & RoBERTa & 0.98 &  \textbf{Data split}: \textsc{Historical} \{$59{,}480$ train; $5{,}700$ test\} - \textsc{Data Stream} \{$30{,}520$ without drift; $31{,}900$ drifted\}  \\ \midrule
Text &  20 Newsgroups & Topic  & 2.1\phantom{*}  & BERT & 0.88  & \multirow{2}{*}{\begin{tabular}[c]{@{}l@{}} \textbf{Training labels}: \textit{Technology}, \textit{Sale-Ads}, \textit{Politics}, \textit{Religion}, \textit{Science} (5 macro-labels)\\ \textbf{Drift}: Simulated with one new macro-label \textit{Recreation}: composed of \textit{baseball}, \textit{hockey}, \textit{motorcycles}, \textit{autos} subtopics \end{tabular} }  \\
 & \multirow{2}{*}{\cite{misctwentynewsgroups113}}  & Detection  & 2.2* & DistilBERT & 0.87 & \\
 & & & 2.3* & RoBERTa & 0.88 & \textbf{Data split}: \textsc{Historical} \{$5{,}080$ train; $3{,}387$ test\} - \textsc{Data Stream} \{$5{,}560$ without drift; $3{,}655$ drifted\} \\ \midrule
Text & 20 Newsgroups & Topic  & 3\phantom{*}  & BERT & 0.87 & \textbf{Training labels}: Training on 6 labels. The 5 macro-labels: \textit{Technology}, \textit{Sale-Ads}, \textit{Politics}, \textit{Religion}, \textit{Science}, \\
 & \multirow{2}{*}{\cite{misctwentynewsgroups113}} &  Detection & &  & & and a subset of the macro-label \textit{Recreation}, composed of subtopics: \textit{baseball} and \textit{hockey}  \\
  & &   & &  &  & \textbf{Drift}: Simulated with another subset of the macro-label \textit{Recreation}: \textit{motorcycles} and \textit{autos} subtopics \\
 & & & &  &  & \textbf{Data split}: \textsc{Historical} \{$5{,}744$ train; $3{,}829$ test\} - \textsc{Data Stream} \{$6{,}304$ without drift; $1{,}805$ drifted\} \\ \midrule

Text & Bias in Bios & Occupation  & 4\phantom{*}  & BERT & 0.94 & \textbf{Training labels}: \textit{Professor}, \textit{Physician}, \textit{Attorney}, \textit{Photographer}, \textit{Journalist} (male gender); \textit{Nurse} (female gender) \\
 & \multirow{2}{*}{\cite{10.1145/3287560.3287572}} &  Classification & &  &  & \textbf{Drift}: Simulated with bios from same labels but opposite gender (male nurses and female for all others occupations) \\
  & &   & &  &  & \textbf{Data split}: \textsc{Historical} \{$39{,}270$ train; $15{,}108$ test\} - \textsc{Data Stream} \{$94{,}184$ without drift; $106{,}242$ drifted\} \\ \midrule
 Image & MNIST & Image & 5.1\phantom{*}  & ViT & 0.98 &  \multirow{2}{*}{\begin{tabular}[c]{@{}l@{}}  \textbf{Training labels}: Digits from $0$ to $7$ \\ \textbf{Drift}: Simulated with two new digits: Digits $8$ and $9$ \end{tabular}} \\
 & \multirow{2}{*}{\cite{6296535}} & Classification  & 5.2* & VGG16 &  0.95 \\
 & & & &  & & \textbf{Data split}: \textsc{Historical} \{$35{,}000$ train; $8{,}000$ test\} - \textsc{Data Stream} \{$12{,}000$ without drift; $13{,}000$ drifted\}  \\ \midrule
Image &  Intel Image  & Image & 6.1\phantom{*} & ViT & 0.90 & \multirow{2}{*}{\begin{tabular}[c]{@{}l@{}} \textbf{Training labels}: \textit{Forest}, \textit{Glacier}, \textit{Mountain}, \textit{Building}, \textit{Street} \\ \textbf{Drift}: Simulated with one new class label: \textit{Sea} \end{tabular}} \\
 &  \multirow{2}{*}{\cite{intelimage}} & Classification & 6.2* & VGG16 & 0.89 \\
&   &  &  &  &   & \textbf{Data split}: \textsc{Historical} \{$6{,}000$ train; $4{,}000$ test\} - \textsc{Data Stream} \{$4{,}256$ without drift; $2{,}780$ drifted\}  \\ \midrule
Image &  STL-10  & Image & 7.1\phantom{*}  & ViT & 0.96 &  \multirow{2}{*}{\begin{tabular}[c]{@{}l@{}}  \textbf{Training labels}: \textit{Airplane}, \textit{Bird}, \textit{Car}, \textit{Cat}, \textit{Deer}, \textit{Dog}, \textit{Horse}, \textit{Monkey}, \textit{Ship} \\ \textbf{Drift}: Simulated with one new class label: \textit{Truck} \end{tabular}} \\
 & \multirow{2}{*}{\cite{stl}}  & Classification  & 7.2* & VGG16 &  0.82 \\
 &   & & &  & & \textbf{Data split}: \textsc{Historical} \{$5{,}850$ train; $2{,}925$ test\} - \textsc{Data Stream} \{$2{,}925$ without drift; $1{,}300$ drifted\}  \\ \midrule
Image &  STL-10  & Image & 8\phantom{*} & ViT & 0.90   & \multirow{2}{*}{\begin{tabular}[c]{@{}l@{}} \textbf{Training labels}: \textit{Airplane}, \textit{Bird}, \textit{Car}, \textit{Cat}, \textit{Deer}, \textit{Dog}, \textit{Horse}, \textit{Monkey}, \textit{Ship}, \textit{Truck} \\ \textbf{Drift}: Simulated by introducing \textit{blur} in the images within the same labels  \end{tabular}} \\
 & \multirow{2}{*}{\cite{stl}}   & Classification &  &   \\
 & &  & & & & \textbf{Data split}: \textsc{Historical} \{$6{,}500$ train; $3{,}250$ test\} - \textsc{Data Stream} \{$3{,}250$ without drift; $3{,}250$ drifted\} \\ \midrule
Image &  FairFace  & Gender & 9\phantom{*} & ViT & 0.95   & \multirow{2}{*}{\begin{tabular}[c]{@{}l@{}} \textbf{Training labels}: \textit{Male}, \textit{Female} - (White, Black, Middle Eastern, Latino-Hispanic, Indian races) \\ \textbf{Drift}: Introduced with images from same labels but different race (East Asian and Southeast Asian races)   \end{tabular}} \\
 & \multirow{2}{*}{\cite{karkkainenfairface}}  & Identification &  &  \\
 &  & & & & & \textbf{Data split}: \textsc{Historical} \{$27{,}984$ train; $7{,}313$ test\} - \textsc{Data Stream} \{$30{,}000$ without drift; $32{,}401$ drifted\} \\ \midrule
Audio &  Common Voice  & Gender & 10\phantom{*} & Wav2Vec & 0.91   & \multirow{2}{*}{\begin{tabular}[c]{@{}l@{}} \textbf{Training labels}: \textit{Male}, \textit{Female} - (US and UK accent) \\ \textbf{Drift}: Introduced with audios from same labels but different accent (Australian, Canadian, Scottish)   \end{tabular}} \\
 & \multirow{2}{*}{\cite{common-voice}}  & Identification &  &  &  \\
 &  & & & & & \textbf{Data split}: \textsc{Historical} \{$70{,}578$ train; $9{,}951$ test\} - \textsc{Data Stream} \{$29{,}556$ without drift; $42{,}697$ drifted\} \\ \bottomrule
\end{tabular}
}
\end{table*}

\section{Evaluation}
\label{sec:evaluation}
We assess \XXX' effectiveness in detecting drift across text, image, and audio classifiers (introduced in \S\ref{subsec:experimental-use-cases}) by evaluating drift detection performance (\S\ref{subsec:evaluation-accuracy}) and execution time (\S\ref{subsec:evaluation-complexity}). 
We also examine its ability to characterize drift trends over time (\S\ref{subsec:evaluation-characterization}) and provide explanations (\S\ref{subsec:evaluation-explanations}). 
Parameter sensitivity is analyzed in Appendix~\ref{apx:evaluation-sensitivity}, 
and a distribution distance ablation study is conducted in Appendix~\ref{apx:evaluation-distance-metric}.

\subsection{Experimental Settings}
\label{subsec:experimental-use-cases}
The 17 experimental use cases are summarized in Table \ref{tab_uses_cases}. 

\noindent
\textbf{Deep learning Classifiers} \ 
We experiment with BERT \cite{BERT}, DistilBERT \cite{distilbert}, and RoBERTa \cite{roberta} deep learning models for NLP classification, VGG16~\cite{simonyan2014very} and Vision Transformer (ViT)~\cite{DBLP:journals/corr/abs-2010-11929} for computer vision, and Wav2Vec \cite{wav2vec} for audio.

\smallskip
\noindent
\textbf{Datasets} \ We train NLP models for topic detection using Ag News \cite{ag-news} and 20 Newsgroups \cite{misctwentynewsgroups113}, and for occupation classification with Bias in Bios \cite{10.1145/3287560.3287572} datasets. In computer vision, we use MNIST \cite{6296535}, Intel Image~\cite{intelimage}, STL-10~\cite{stl}, and FairFace~\cite{karkkainenfairface} for image classification. In the audio domain, we use Common Voice~\cite{common-voice} for speaker gender classification.

\smallskip
\noindent
\textbf{Embedding Extraction} \ For BERT, DistilBERT, RoBERTa, and ViT models, we extract the embedding of the [CLS] token from the last hidden layer (dimensionality $d\!=\!768$). 
For VGG16, we extract and flatten the last convolutional layer, resulting in $d\!=\!512$, $d\!=\!8{,}192$, and $d\!=\!4{,}608$ for use cases 5.2, 6.2, and 7.2, respectively. 
For Wav2Vec, we extract and average all the embeddings of the last hidden layer ($d  = 768$).

\smallskip
\noindent
\textbf{Data Split} \ Each dataset has been split into four non-overlapping subsets, as described in the last column of Table~\ref{tab_uses_cases}. Two subsets are used to fine-tune and evaluate the classifiers (train and test splits). These two subsets represent \textit{historical} data that can be used by drift detectors as reference data. The third subset is reserved for generating windows in the \textit{data stream}, simulating new, unseen data with a similar distribution to the training set (without drift). 
The fourth subset is designated for simulating drift. Together, these last two subsets form the data stream for drift detection experiments. Importantly, the historical and data stream subsets are disjoint.

\smallskip
\noindent
\textbf{Drift Simulation} \ We simulate different drift sources, as described in 
Table~\ref{tab_uses_cases}. In use cases 1-3 and 5-7, drift is simulated by introducing a new unknown class label, obtained by removing samples from a label during training and by presenting them in the data stream. 
For instance, in use case~1, all $31{,}900$ samples for the class \textit{Science/Tech} are used to simulate drift. The remaining samples from classes \textit{World}, \textit{Business}, and \textit{Sport} are divided into $59{,}480$ and $5{,}700$ to fine-tune and evaluate the model (historical data), while $30{,}520$ to simulate data stream samples without drift. 
In use case 3, the same dataset is used as in use case 2, but drift is more subtle, as only a subset of a class is used to simulate the drift by exploiting the hierarchical categorization of labels. 
In use case 8, we simulate data drift by altering the input features through image blurring. Gaussian blur is applied with a radius of 2 to a circular patch covering $D_{\%}$ (drift percentage) of pixels. 
In use cases 4, 9, and 10, drift is simulated with biased classifiers trained on underrepresented protected features. 
Drift occurs by presenting samples from subgroups with the same labels but different features, exposing the classifier's bias.
In use case 4, we build upon previous research showing that occupation classifiers can exhibit gender bias \cite{ravfogel-etal-2020-null,10.1145/3686924}, and we trained a classifier on training bios where all nurses were female while individuals in other occupations were male. In the data stream, drift is introduced by mixing non-drift with drifted samples, where nurses are male and individuals in other occupations are female.
In use cases 9 and 10, drift is simulated as racial bias by presenting images from different races or audio with English accents other than those used during training.

\smallskip
\noindent
\textbf{Windows Generation} \ When generating windows of the data stream, we use samples—either with or without drift—that the model has never encountered during training or testing, whose dimensionalities are specified in the last column of Table~\ref{tab_uses_cases} (\textsc{Data Stream}). 
Windows are sampled with replacement due to dimensionality constraints (samples may belong to more windows).
When generating a window without drift, we randomly select a balanced sample by class labels from the new data without drift.  
These samples are previously unseen by the model but share the same distribution as the training data.
When generating windows containing $D_{\%}$ of drift, we randomly select $D_{\%}$ of the window size ($m_w$) from the drifted samples, while the remaining $(100 - D)_{\%}$ are balanced samples from the new unseen data without drift.
Use case 8 is the only exception where a $D_{\%}$ of pixels over all images are blurred.

\smallskip
\noindent
\textbf{DriftLens Configuration} \ In the \textit{offline} phase, the entire historical training and test sets are used for baseline and threshold estimation, respectively, with sizes shown in the last column of Table~\ref{tab_uses_cases} (\textsc{Historical}). 
The number of principal components for \textit{per-batch} and \textit{per-label} embedding are $d' = 150$ and $d'_l = 75$, except for use case 10, where $d' = d'_l = 25$ ensures full-rank matrices. The number of windows for threshold estimation is $n_{th} = 10{,}000$, with sensitivity $T_{\alpha} = 0.01$.

\smallskip
\noindent
\textbf{Baseline Detectors} \
We compare \XXX\ with four unsupervised statistical-based detectors from previous work: 
Maximum Mean Discrepancy\footnote{{\scriptsize \url{https://docs.seldon.io/projects/alibi-detect/en/stable/cd/methods/mmddrift.html}}} (MMD) \cite{Rabanser2018FailingLA,JMLR:v13:gretton12a},
Kolmogorov-Smirnov\footnote{{\scriptsize\url{https://docs.seldon.io/projects/alibi-detect/en/stable/cd/methods/ksdrift.html}}} (KS) \cite{Rabanser2018FailingLA,Kolmogorov-Smirnov-Test}, 
Least-Squares Density Difference\footnote{{\scriptsize \url{https://docs.seldon.io/projects/alibi-detect/en/stable/cd/methods/lsdddrift.html}}} (LSDD) \cite{7745962}, and 
Cramér-von Mises\footnote{{\scriptsize \url{https://docs.seldon.io/projects/alibi-detect/en/stable/cd/methods/cvmdrift.html}}} (CVM) \cite{cramer1928composition}. 
We use the implementation provided by the Alibi Detect library~\cite{alibi-detect}. 
We keep the default parameters configuration, and we use embedding vectors as input. 
The \textit{p-value} to discriminate drifted and non-drifted distributions is set to 0.05 as the default. 
However, as we will discuss in \S\ref{subsec:evaluation-complexity}, the running time of some detectors is significantly influenced by the reference set dimensionality~$m_b$. 
Given our focus on real-time drift detection, and to ensure a fair comparison, we constrain the reference window dimensionality. Specifically, we limit the execution time for each detector to predict drift in a window to a maximum of 30 seconds, estimated using $m_w=1{,}000$ and $d=1{,}000$.\footnote{Executed on an Apple M1 MacBook Pro 13 2020 with 16GB of RAM.} To this end, we set the maximum reference window size $m_b \approx 14{,}000$ for MMD and $m_b \approx 8{,}500$ for LSDD. In contrast, we used the full training set for KS and CVM due to their faster execution time. 
When the training set size exceeds these constraints, the reference set is created by randomly sampling a balanced subset equal to the maximum size.
The only exception is the 20 Newsgroups dataset, where we used $m_b \approx 1{,}700$ for use case 2 and $m_b \approx 2{,}050$ for use case 3 across all detectors to ensure a balanced reference set.

\subsection{Drift Detection Performance Evaluation}
\label{subsec:evaluation-accuracy}
This evaluation assesses the effectiveness and general applicability of \XXX\ in detecting drifted windows of varying severity across models, data types, and tasks (RQ1 and RQ2 in \S\ref{sec:intro}). We approach drift detection as a binary classification task---predict whether a window contains drift.

\begin{table*}[h!]
\scriptsize
\centering
\caption{\textbf{Drift detection performance evaluation.} For \XXX, \textsc{MMD} \cite{Rabanser2018FailingLA}, \textsc{KS}~\cite{Rabanser2018FailingLA}, \textsc{LSDD}~\cite{7745962}, \textsc{CVM} \cite{cramer1928composition} detectors \cite{alibi-detect}, and each window size $m_w$ are reported (i) the \textit{accuracy} separately per drift percentage $D_{\%}\!\in\!\{0\%, 5\%, 10\%, 15\%, 20\%\}$, and (ii) the \textit{Harmonic Drift Detection} mean $H_{DD}$. Accuracy is computed over 100 windows and averaged by repeating 5 runs. The best-performing detector for each use case based on the overall $H_{DD}$ is highlighted, and per window size is in~bold.}
\label{tab:drift-detection-experiments}

\begin{subtable}[t]{\textwidth}
\centering
\caption{\textbf{Larger data volumes:} Data stream window sizes $m_w \in \{500, 1000, 2000\}$}
\label{tab:drift-detection-experiments-larger}
\scalebox{.8355}{
\begin{tabular}{clclllllllllllllllllllll}
\toprule

\multirow{5}{*}{\textbf{Use}} & & \multirow{5}{*}{\textbf{Drift}} & & \multicolumn{6}{c}{\textit{$m_w = 500$}} & \multicolumn{1}{c}{\textit{}} & \multicolumn{6}{c}{\textit{$m_w = 1000$}} & \multicolumn{1}{c}{\textit{}} & \multicolumn{6}{c}{\textit{$m_w = 2000$}} \\ \cline{5-10} \cline{12-17} \cline{19-24} 
 & & &  & & & & & & & & & & & & & & & & & & \\
 & &  &  & \multicolumn{5}{c}{\textbf{Drift Percentage $D_{\%}$}} &  & \multicolumn{1}{c}{\textbf{}} &  \multicolumn{5}{c}{\textbf{Drift Percentage $D_{\%}$}} & \multicolumn{1}{c}{\textbf{}} & & \multicolumn{5}{c}{\textbf{Drift Percentage $D_{\%}$}} \\
 \textbf{Case} & & \multicolumn{1}{c}{\textbf{Detector}} & & 0\% & 5\% & 10\% & 15\% & 20\% & $H_{DD}$ & & 0\% & 5\% & 10\% & 15\% & 20\% & $H_{DD}$ & & 0\% & 5\% & 10\% & 15\% & 20\% & $H_{DD}$ \\ \cline{5-9} \cline{12-16} \cline{19-23} 
 & & & & & & & & & & & & & & & & & & & & & & \\ \toprule

\multirow{2}{*}{1.1 } 
 & & \multicolumn{1}{l}{MMD} & & 1.00 & 0.00 & 0.10 & 0.96 & 1.00 & \textit{0.68} & & 1.00 & 0.00 & 0.83 & 1.00 &  1.00 & \textit{0.83} & & 1.00 & 0.06 & 1.00 & 1.00 & 1.00 & \textit{0.87} \\
 & & \multicolumn{1}{l}{KS} & & 1.00 & 0.01 & 0.21 & 0.98 & 1.00 & \textit{0.71} & & 0.99 & 0.03 & 0.95 & 1.00 & 1.00 & \textit{0.85} & & 1.00 & 0.38 & 1.00 & 1.00 & 1.00 & \textit{0.91} \\
Ag News & & \multicolumn{1}{l}{LSDD} & & 1.00 & 0.00 & 0.13 & 0.93 & 1.00 & \textit{0.68} & & 1.00 & 0.00 & 0.83 & 1.00 &   1.00 & \textit{0.83} & & 1.00 & 0.12 & 1.00 & 1.00 & 1.00 &  \textit{0.88} \\
\multirow{2}{*}{BERT} & & \multicolumn{1}{l}{CVM} & & 1.00 & 0.01 & 0.22 & 0.99 & 1.00 & \textit{0.71} & & 0.99 & 0.02 & 0.99 & 1.00 & 1.00 & \textit{0.86} & & 1.00 & 0.44 & 1.00 & 1.00 & 1.00 & \textit{0.92} \\
 & & \multicolumn{1}{l}{\cellcolor{mycellcolor}\XXX} & \cellcolor{mycellcolor} & \cellcolor{mycellcolor}0.99 & \cellcolor{mycellcolor}0.83 & \cellcolor{mycellcolor}1.00 & \cellcolor{mycellcolor}1.00 & \cellcolor{mycellcolor}1.00 & \cellcolor{mycellcolor}\textit{\textbf{0.97}} & \cellcolor{mycellcolor} & \cellcolor{mycellcolor}1.00 & \cellcolor{mycellcolor}0.98 & \cellcolor{mycellcolor}1.00 & \cellcolor{mycellcolor}1.00 &   \cellcolor{mycellcolor}1.00 & \cellcolor{mycellcolor}\textit{\textbf{1.00}} & \cellcolor{mycellcolor} & \cellcolor{mycellcolor}0.97 & \cellcolor{mycellcolor}1.00 & \cellcolor{mycellcolor}1.00 & \cellcolor{mycellcolor}1.00 & \cellcolor{mycellcolor}1.00 & \cellcolor{mycellcolor}\textit{\textbf{0.98}} \\  \midrule

\multirow{2}{*}{4} 
 & & \multicolumn{1}{l}{MMD} & & 1.00 & 0.00 & 0.94 & 1.00 & 1.00 & \textit{0.85} & & 1.00 & 0.00 & 1.00 & 1.00 &  1.00 & \textit{0.86} & & 1.00 & 0.89 & 1.00 & 1.00 & 1.00 & \textit{0.99} \\
 & & \multicolumn{1}{l}{KS} & & 1.00 & 0.01 & 0.70 & 1.00 & 1.00 & \textit{0.81} & & 1.00 & 0.22 & 1.00 & 1.00 & 1.00 & \textit{0.89} & & 0.97 & 0.99 & 1.00 & 1.00 & 1.00 & \textit{0.98} \\
Bias in Bios & & \multicolumn{1}{l}{LSDD} & & 1.00 & 0.00 & 0.95 & 1.00 & 1.00 & \textit{0.85} & & 1.00 & 0.01 & 1.00 & 1.00 &   1.00 & \textit{0.86} & & 1.00 & 0.99 & 1.00 & 1.00 & 1.00 &  \textbf{\textit{1.00}} \\
\multirow{2}{*}{BERT} & & \multicolumn{1}{l}{CVM} & & 1.00 & 0.00 & 0.47 & 1.00 & 1.00 & \textit{0.77} & & 1.00 & 0.13 & 1.00 & 1.00 & 1.00 & \textit{0.88} & & 0.98 & 0.96 & 1.00 & 1.00 & 1.00 & \textit{0.98} \\
 & & \multicolumn{1}{l}{\cellcolor{mycellcolor}\XXX} & \cellcolor{mycellcolor} & \cellcolor{mycellcolor}0.97 & \cellcolor{mycellcolor}0.53 & \cellcolor{mycellcolor}1.00 & \cellcolor{mycellcolor}1.00 & \cellcolor{mycellcolor}1.00 & \cellcolor{mycellcolor}\textit{\textbf{0.93}} & \cellcolor{mycellcolor} & \cellcolor{mycellcolor}0.98 & \cellcolor{mycellcolor}0.96 & \cellcolor{mycellcolor}1.00 & \cellcolor{mycellcolor}1.00 &   \cellcolor{mycellcolor}1.00 & \cellcolor{mycellcolor}\textit{\textbf{0.99}} & \cellcolor{mycellcolor} & \cellcolor{mycellcolor}0.98 & \cellcolor{mycellcolor}1.00 & \cellcolor{mycellcolor}1.00 & \cellcolor{mycellcolor}1.00 & \cellcolor{mycellcolor}1.00 & \cellcolor{mycellcolor}\textit{0.99} \\  \midrule
 \multirow{2}{*}{5.1}                
 & & \multicolumn{1}{l}{MMD}  & & 1.00 & 0.00 & 0.01 & 0.90 & 1.00 & \textit{0.65} & & 1.00 & 0.00 & 0.64 & 1.00 & 1.00 & \textit{0.80} & & 1.00 & 0.00 & 1.00 & 1.00 & 1.00 & \textit{0.86} \\
 & & \multicolumn{1}{l}{KS}  & & 1.00 & 0.01 & 0.31 & 0.99 & 1.00 & \textit{0.73} & & 1.00 & 0.07 & 1.00 & 1.00 & 1.00 & \textit{0.87} & & 1.00 & 0.84 & 1.00 & 1.00 & 1.00 & \textbf{\textit{0.98}} \\
MNIST  & & \multicolumn{1}{l}{LSDD} &  & 1.00 & 0.00 & 0.00 & 0.77 & 1.00 & \textit{0.61} & & 1.00 & 0.00 & 0.50 & 1.00 & 1.00 & \textit{0.77} & & 1.00 & 0.00 & 1.00 & 1.00 & 1.00 & \textit{0.86} \\
 \multirow{2}{*}{ViT}  & & \multicolumn{1}{l}{CVM}  &  & 1.00 & 0.00 & 0.29 & 1.00 & 1.00 & \textit{0.73} & & 1.00 & 0.03 & 1.00 & 1.00 & 1.00 & \textit{0.86} & & 1.00 & 0.83 & 1.00 & 1.00 & 1.00 & \textbf{\textit{0.98}} \\
 & & \multicolumn{1}{l}{\cellcolor{mycellcolor}\XXX} & \cellcolor{mycellcolor} & \cellcolor{mycellcolor}1.00 & \cellcolor{mycellcolor}0.13 & \cellcolor{mycellcolor}1.00 & \cellcolor{mycellcolor}1.00 & \cellcolor{mycellcolor}1.00 & \cellcolor{mycellcolor}\textbf{\textit{0.88}} & \cellcolor{mycellcolor} & \cellcolor{mycellcolor}1.00 & \cellcolor{mycellcolor}0.43 & \cellcolor{mycellcolor}1.00 & \cellcolor{mycellcolor}0.58 & \cellcolor{mycellcolor}0.85 & \cellcolor{mycellcolor}\textbf{\textit{0.92}} & \cellcolor{mycellcolor} & \cellcolor{mycellcolor}1.00 & \cellcolor{mycellcolor}0.85 & \cellcolor{mycellcolor}1.00 & \cellcolor{mycellcolor}1.00 & \cellcolor{mycellcolor}1.00 & \cellcolor{mycellcolor}\textbf{\textit{0.98}} \\ \midrule
\multirow{2}{*}{9}                
 & & \multicolumn{1}{l}{MMD}  & & 0.75 & 0.38 & 0.61 & 0.75 & 0.92 & \textbf{0.71} & & 0.15 & 0.95 & 0.99 & 1.00 & 1.00 & \textit{0.26} & & 0.01 & 1.00 & 1.00 & 1.00 & 1.00 & \textit{0.01} \\
 & & \multicolumn{1}{l}{KS}  & & 0.39 & 0.67 & 0.77 & 0.89 & 0.94 & \textit{0.52} & & 0.00 & 1.00 & 1.00 & 1.00 & 1.00 & \textit{0.00} & & 0.00 & 1.00 & 1.00 & 1.00 & 1.00 & \textit{0.00} \\
FairFace & & \multicolumn{1}{l}{LSDD} &  & 0.85 & 0.30 & 0.46 & 0.71 & 0.89 & \textit{0.70} & & 0.31 & 0.89 & 0.98 & 1.00 & 1.00 & \textit{0.47} & & 0.00 & 1.00 & 1.00 & 1.00 & 1.00 & \textit{0.00} \\
 \multirow{2}{*}{ViT}  & & \multicolumn{1}{l}{CVM}  &  & 0.42 & 0.64 & 0.78 & 0.88 & 0.96 & \textit{0.56} & & 0.03 & 0.99 & 1.00 & 1.00 & 1.00 & \textit{0.05} & & 0.00 & 1.00 & 1.00 & 1.00 & 1.00 & \textit{0.00} \\
 & & \multicolumn{1}{l}{\cellcolor{mycellcolor}\XXX} & \cellcolor{mycellcolor} & \cellcolor{mycellcolor}1.00 & \cellcolor{mycellcolor}0.02 & \cellcolor{mycellcolor}0.07 & \cellcolor{mycellcolor}0.27 & \cellcolor{mycellcolor}0.64 & \cellcolor{mycellcolor}\textit{0.40} & \cellcolor{mycellcolor} & \cellcolor{mycellcolor}1.00 & \cellcolor{mycellcolor}0.04 & \cellcolor{mycellcolor}0.19 & \cellcolor{mycellcolor}0.65 & \cellcolor{mycellcolor}0.95 & \cellcolor{mycellcolor}\textbf{\textit{0.63}} & \cellcolor{mycellcolor} & \cellcolor{mycellcolor}1.00 & \cellcolor{mycellcolor}0.06 & \cellcolor{mycellcolor}0.48 & \cellcolor{mycellcolor}0.96 & \cellcolor{mycellcolor}1.00 & \cellcolor{mycellcolor}\textbf{\textit{0.77}} \\ \midrule
 
\multirow{2}{*}{10}                
 & & \multicolumn{1}{l}{MMD}  & & 0.14 & 0.93 & 0.99 & 1.00 & 1.00 & \textit{0.24} & & 0.00 & 1.00 & 1.00 & 1.00 & 1.00 & \textit{0.00} & & 0.00 & 1.00 & 1.00 & 1.00 & 1.00 & \textit{0.00} \\
 & & \multicolumn{1}{l}{KS}  & & 0.10 & 0.95 & 0.97 & 0.99 & 1.00 & \textit{0.17} & & 0.00 & 1.00 & 1.00 & 1.00 & 1.00 & \textit{0.00} & & 0.00 & 1.00 & 1.00 & 1.00 & 1.00 & \textit{0.00} \\
CommonVoice & & \multicolumn{1}{l}{LSDD} &  & 0.02 & 0.98 & 1.00 & 1.00 & 1.00 & \textit{0.05} & & 0.00 & 1.00 & 1.00 & 1.00 & 1.00 & \textit{0.00} & & 0.00 & 1.00 & 1.00 & 1.00 & 1.00 & \textit{0.00} \\
 \multirow{2}{*}{Wav2Vec}  & & \multicolumn{1}{l}{CVM}  &  & 0.00 & 1.00 & 1.00 & 1.00 & 1.00 & \textit{0.00} & & 0.00 & 1.00 & 1.00 & 1.00 & 1.00 & \textit{0.00} & & 0.00 & 1.00 & 1.00 & 1.00 & 1.00 & \textit{0.00} \\
 & & \multicolumn{1}{l}{\cellcolor{mycellcolor}\XXX} & \cellcolor{mycellcolor} & \cellcolor{mycellcolor}0.97 & \cellcolor{mycellcolor}0.07 & \cellcolor{mycellcolor}0.17 & \cellcolor{mycellcolor}0.27 & \cellcolor{mycellcolor}0.42 & \cellcolor{mycellcolor}\textbf{\textit{0.38}} & \cellcolor{mycellcolor} & \cellcolor{mycellcolor}0.93 & \cellcolor{mycellcolor}0.19 & \cellcolor{mycellcolor}0.38 & \cellcolor{mycellcolor}0.58 & \cellcolor{mycellcolor}0.85 & \cellcolor{mycellcolor}\textbf{\textit{0.65}} & \cellcolor{mycellcolor} & \cellcolor{mycellcolor}0.84 & \cellcolor{mycellcolor}0.42 & \cellcolor{mycellcolor}0.75 & \cellcolor{mycellcolor}0.95 & \cellcolor{mycellcolor}0.99 & \cellcolor{mycellcolor}\textbf{\textit{0.81}} \\ \bottomrule
\end{tabular}
}
\end{subtable}

\vspace{1em}

\begin{subtable}[t]{\textwidth}
\centering
\caption{\textbf{Smaller data volumes:} Data stream window sizes $m_w \in \{250, 500, 1000\}$}
\label{tab:drift-detection-experiments-smaller}
\scalebox{.8355}{
\begin{tabular}{clclllllllllllllllllllll}
\toprule

\multirow{5}{*}{\textbf{Use}} & & \multirow{5}{*}{\textbf{Drift}} & & \multicolumn{6}{c}{\textit{$m_w = 250$}} & \multicolumn{1}{c}{\textit{}} & \multicolumn{6}{c}{\textit{$m_w = 500$}} & \multicolumn{1}{c}{\textit{}} & \multicolumn{6}{c}{\textit{$m_w = 1000$}} \\ \cline{5-10} \cline{12-17} \cline{19-24} 
 & & &  & & & & & & & & & & & & & & & & & & \\
 & &  &  & \multicolumn{5}{c}{\textbf{Drift Percentage $D_{\%}$}} &  & \multicolumn{1}{c}{\textbf{}} &  \multicolumn{5}{c}{\textbf{Drift Percentage $D_{\%}$}} & \multicolumn{1}{c}{\textbf{}} & & \multicolumn{5}{c}{\textbf{Drift Percentage $D_{\%}$}} \\
 \textbf{Case} & & \multicolumn{1}{c}{\textbf{Detector}} & & 0\% & 5\% & 10\% & 15\% & 20\% & $H_{DD}$ & & 0\% & 5\% & 10\% & 15\% & 20\% & $H_{DD}$ & & 0\% & 5\% & 10\% & 15\% & 20\% & $H_{DD}$ \\ \cline{5-9} \cline{12-16} \cline{19-23} 
 & & & & & & & & & & & & & & & & & & & & & & \\ \toprule

\multirow{2}{*}{2.1} 
 & &  \multicolumn{1}{l}{MMD} & & 0.83 & 0.33 & 0.74 & 0.98 & 1.00 & \textit{0.80} & & 0.06 & 1.00 & 1.00 & 1.00 & 1.00 & \textit{0.11} & & 0.00 & 1.00 & 1.00 & 1.00 & 1.00 & \textit{0.00} \\
 & & \multicolumn{1}{l}{KS} & & 0.00 & 1.00 & 1.00 & 1.00 & 1.00 & \textit{0.00} & & 0.00 & 1.00 & 1.00 & 1.00 & 1.00 & \textit{0.00} & & 0.00 & 1.00 & 1.00 & 1.00 & 1.00 &  \textit{0.00} \\
20Newsgroup & & \multicolumn{1}{l}{LSDD} & & 1.00 & 0.01 & 0.04 & 0.19 & 0.48 & \textit{0.31} & & 0.98 & 0.14 & 0.41 & 0.76 & 0.96 & \textit{0.72} & & 0.61 & 0.65 & 0.93 & 1.00 & 1.00 & \textit{0.72} \\
 \multirow{2}{*}{BERT}  & & \multicolumn{1}{l}{CVM} & & 0.00 & 1.00 & 1.00 & 1.00 & 1.00 & \textit{0.00} & & 0.00 & 1.00 & 1.00 & 1.00 & 1.00 & \textit{0.00} & & 0.00 & 1.00 & 1.00 & 1.00 & 1.00 & \textit{0.00} \\
 & & \multicolumn{1}{l}{\cellcolor{mycellcolor}\XXX} & \cellcolor{mycellcolor} & \cellcolor{mycellcolor}0.92 & \cellcolor{mycellcolor}0.28 & \cellcolor{mycellcolor}0.68 & \cellcolor{mycellcolor}0.96 & \cellcolor{mycellcolor}1.00 & \cellcolor{mycellcolor}\textbf{\textit{0.81}} & \cellcolor{mycellcolor} & \cellcolor{mycellcolor}0.89 & \cellcolor{mycellcolor}0.42 & \cellcolor{mycellcolor}0.92 & \cellcolor{mycellcolor}1.00 & \cellcolor{mycellcolor}1.00 & \cellcolor{mycellcolor}\textbf{\textit{0.86}} & \cellcolor{mycellcolor} & \cellcolor{mycellcolor}0.84 & \cellcolor{mycellcolor}0.78 & \cellcolor{mycellcolor}1.00 & \cellcolor{mycellcolor}1.00 & \cellcolor{mycellcolor}1.00 & \cellcolor{mycellcolor}\textbf{\textit{0.89}} \\ \midrule
\multirow{2}{*}{3} 
& & \multicolumn{1}{l}{MMD} & & 1.00 & 0.00 & 0.04 & 0.59 & 0.98 & \textit{0.57} & & 0.99 & 0.08 & 0.76 & 1.00 &   1.00 & \textit{0.83} & & 0.48 & 0.89 & 1.00 & 1.00 & 1.00 & \textit{0.61} \\
& & \multicolumn{1}{l}{KS} & & 0.01 & 1.00 & 1.00 & 1.00 & 1.00 & \textit{0.02} & & 0.00 & 1.00 & 1.00 & 1.00 &   1.00 & \textit{0.00} & & 0.00 & 1.00 & 1.00 & 1.00 & 1.00 & \textit{0.00} \\
20Newsgroup & & \multicolumn{1}{l}{LSDD} & & 1.00 & 0.00 & 0.00 & 0.03 & 0.35 & \textit{0.17} & & 1.00 & 0.00 & 0.02 & 0.55 & 0.99 & \textit{0.56} & & 1.00 & 0.01 & 0.41 &  1.00 &  1.00 & \textit{0.75} \\
 \multirow{2}{*}{BERT}  & & \multicolumn{1}{l}{CVM} & & 0.00 & 1.00 & 1.00 & 1.00 & 1.00 & \textit{0.00} & & 0.00 & 1.00 & 1.00 & 1.00 &     1.00 & \textit{0.00} & & 0.00 & 1.00 & 1.00 & 1.00 & 1.00  & \textit{0.00}  \\
& & \multicolumn{1}{l}{\cellcolor{mycellcolor}\XXX} & \cellcolor{mycellcolor} & \cellcolor{mycellcolor}0.97 & \cellcolor{mycellcolor}0.21 & \cellcolor{mycellcolor}0.65 & \cellcolor{mycellcolor}0.96 & \cellcolor{mycellcolor}1.00 & \cellcolor{mycellcolor}\textbf{\textit{0.82}} & \cellcolor{mycellcolor} & \cellcolor{mycellcolor}0.98 & \cellcolor{mycellcolor}0.35 & \cellcolor{mycellcolor}0.94 & \cellcolor{mycellcolor}1.00 & \cellcolor{mycellcolor}1.00 & \cellcolor{mycellcolor}\textbf{\textit{0.89}} & \cellcolor{mycellcolor} & \cellcolor{mycellcolor}0.98 & \cellcolor{mycellcolor}0.63 & \cellcolor{mycellcolor}1.00 & \cellcolor{mycellcolor}1.00 &   \cellcolor{mycellcolor}1.00 & \cellcolor{mycellcolor}\textbf{\textit{0.94}} \\ \midrule

\multirow{2}{*}{6.1}
& & \multicolumn{1}{l}{MMD} & & 1.00 & 0.00 & 0.00 & 0.41 & 0.99 & \textit{0.52} & & 1.00 & 0.00 & 0.21 & 1.00 &  1.00 & \textit{0.71} & & 1.00 & 0.01 & 0.97 & 1.00 & 1.00  & \textit{0.85} \\
& & \multicolumn{1}{l}{KS} & & 1.00 & 0.01 & 0.04 & 0.53 & 1.00 & \textit{0.56} & & 1.00 & 0.03 & 0.43 & 1.00 & 1.00 & \textit{0.76} &  & 0.99 & 0.15 & 0.99 & 1.00 & 1.00 &   \textit{0.88} \\
Intel Image & & \multicolumn{1}{l}{LSDD} & & 1.00 & 0.00 & 0.00 &  0.08 & 0.69 & \textit{0.32} & & 1.00 & 0.00 & 0.02 & 0.71 & 1.00 & \textit{0.60} & & 1.00 & 0.01 & 0.49 & 1.00 & 1.00 & \textit{0.77}  \\
 \multirow{2}{*}{ViT} & & \multicolumn{1}{l}{CVM} & & 1.00 & 0.01 & 0.04 & 0.53 & 1.00 & \textit{0.56} & & 1.00 & 0.03 & 0.47 & 1.00 & 1.00 & \textit{0.77} & & 1.00 & 0.20 & 0.99 & 1.00 & 1.00 & \textit{\textit{0.89}} \\
& & \multicolumn{1}{l}{\cellcolor{mycellcolor}\XXX} & \cellcolor{mycellcolor} & \cellcolor{mycellcolor}0.95 & \cellcolor{mycellcolor}0.37 & \cellcolor{mycellcolor}1.00 & \cellcolor{mycellcolor}1.00 & \cellcolor{mycellcolor}1.00 & \cellcolor{mycellcolor}\textbf{\textit{0.88}} & \cellcolor{mycellcolor} & \cellcolor{mycellcolor}0.96 & \cellcolor{mycellcolor}0.57 & \cellcolor{mycellcolor}1.00 & \cellcolor{mycellcolor}1.00 &   \cellcolor{mycellcolor}1.00 & \cellcolor{mycellcolor}\textbf{\textit{0.93}} & \cellcolor{mycellcolor} & \cellcolor{mycellcolor}0.96 & \cellcolor{mycellcolor}0.75 & \cellcolor{mycellcolor}1.00 & \cellcolor{mycellcolor}1.00 & \cellcolor{mycellcolor}1.00 &  \cellcolor{mycellcolor}\textbf{0.95}    \\ \midrule

\multirow{2}{*}{7.1} 
& & \multicolumn{1}{l}{MMD} & & 1.00 & 0.00 & 0.06 & 1.00 & 1.00 & \textit{0.68} & & 1.00 & 0.00 & 1.00 & 1.00 & 1.00 & \textit{0.86} & & 1.00 & 0.10 & 1.00 & 1.00 & 1.00 & \textit{0.87} \\
 & & \multicolumn{1}{l}{KS} & & 0.97 & 0.12 & 0.37 & 0.98 & 1.00 & \textit{0.76}  & & 0.82 & 0.53 & 0.98 & 1.00 & 1.00 & \textit{0.85} & & 0.40 & 0.99 & 1.00 & 1.00 & 1.00 &  \textit{0.57} \\
STL-10 & & \multicolumn{1}{l}{LSDD} & & 1.00 & 0.00 & 0.01 & 1.00 & 1.00 & \textit{0.67} & & 1.00 & 0.00 & 0.99 & 1.00 & 1.00 & \textit{0.86} & & 1.00 & 0.09 & 1.00 & 1.00 & 1.00 & \textit{0.87} \\
 \multirow{2}{*}{ViT}  & & \multicolumn{1}{l}{CVM} & & 0.96 & 0.18 & 0.43 & 0.98 & 1.00 & \textit{0.77} & & 0.74 & 0.65 & 0.99  & 1.00 & 1.00 & \textit{0.82} & & 0.24 & 1.00 & 1.00 & 1.00 & 1.00 & \textit{0.38} \\
 & & \multicolumn{1}{l}{\cellcolor{mycellcolor}\XXX} & \cellcolor{mycellcolor} & \cellcolor{mycellcolor}0.96 & \cellcolor{mycellcolor}0.82 & \cellcolor{mycellcolor}1.00 & \cellcolor{mycellcolor}1.00 & \cellcolor{mycellcolor}1.00 & \cellcolor{mycellcolor}\textbf{\textit{0.96}} & \cellcolor{mycellcolor} & \cellcolor{mycellcolor}0.96 & \cellcolor{mycellcolor}1.00 & \cellcolor{mycellcolor}1.00 & \cellcolor{mycellcolor}1.00 & \cellcolor{mycellcolor}1.00 & \cellcolor{mycellcolor}\textbf{\textit{0.98}} & \cellcolor{mycellcolor} & \cellcolor{mycellcolor}0.98 & \cellcolor{mycellcolor}1.00 & \cellcolor{mycellcolor}1.00 & \cellcolor{mycellcolor}1.00 & \cellcolor{mycellcolor}1.00 & \cellcolor{mycellcolor}\textbf{\textit{0.99}} \\ \midrule
 \multirow{2}{*}{8} 
 & & \multicolumn{1}{l}{\cellcolor{mycellcolor}MMD}  & \cellcolor{mycellcolor} & \cellcolor{mycellcolor}1.00 & \cellcolor{mycellcolor}1.00 & \cellcolor{mycellcolor}1.00 & \cellcolor{mycellcolor}1.00 & \cellcolor{mycellcolor}1.00 & \cellcolor{mycellcolor}\textbf{\textit{1.00}} & \cellcolor{mycellcolor} & \cellcolor{mycellcolor}1.00 & \cellcolor{mycellcolor}1.00 & \cellcolor{mycellcolor}1.00 & \cellcolor{mycellcolor}1.00 &  \cellcolor{mycellcolor}1.00 & \cellcolor{mycellcolor}\textbf{\textit{1.00}} & \cellcolor{mycellcolor} & \cellcolor{mycellcolor}1.00 & \cellcolor{mycellcolor}1.00 & \cellcolor{mycellcolor}1.00 & \cellcolor{mycellcolor}1.00 & \cellcolor{mycellcolor}1.00 & \cellcolor{mycellcolor}\textbf{\textit{1.00}} \\
 & & \multicolumn{1}{l}{KS}   & & 0.95 & 1.00 & 1.00 & 1.00 & 1.00 & \textit{0.97} & & 0.65 & 1.00 & 1.00 & 1.00 &  1.00 & \textit{0.79} & & 0.12 & 1.00 & 1.00 & 1.00 & 1.00 & \textit{0.21} \\
STL-10 & & \multicolumn{1}{l}{LSDD} & & 1.00 & 0.04 & 0.16 & 0.48 & 0.80 & \textit{0.54} & & 1.00 & 1.00 & 1.00 & 1.00 &  1.00 & \textbf{\textit{1.00}} & & 1.00 & 1.00 & 1.00 & 1.00 & 1.00 & \textbf{\textit{1.00}} \\
 \multirow{2}{*}{ViT}  & & \multicolumn{1}{l}{CVM}  & & 0.94 & 1.00 & 1.00 & 1.00 & 1.00 & \textit{0.97} & & 0.60 & 1.00 & 1.00 & 1.00 &  1.00 & \textit{0.75} & & 0.07 & 1.00 & 1.00 & 1.00 & 1.00 & \textit{0.13} \\
 & & \multicolumn{1}{l}{\XXX} & & 0.95 & 1.00 & 1.00 & 1.00 & 1.00 & \textit{0.97} & & 0.93 & 1.00 & 1.00 & 1.00 &  1.00 & \textit{0.96} & & 0.92 & 1.00 & 1.00 & 1.00 & 1.00 & \textit{0.96} \\ 
 \bottomrule
\end{tabular}
}
\end{subtable}

\end{table*}

\smallskip
\noindent
\textbf{Evaluation Metrics} \ We measure the \textit{accuracy} in predicting drift ($A_{D_{\%}}$) with different degrees of severity $D_{\%}\!\in\!\{0\%, 5\%, 10\%, 15\%, 20\%\}$. Windows without drift ($D_{\%}\!=\!0\%$) are used to measure type I errors (false alarm)---no drift, but has been detected. If a window contains any percentage of drifted examples, 
the ground truth is set to 1; otherwise, it is set to 0. For each $D_{\%}$, the accuracy is computed as the mean over 5 independent runs of the accuracy computed on 100 windows with a fixed size of $m_w$.\footnote{At each run the threshold of \XXX\ is re-estimated and the reference set of the other detectors is re-sampled (if exceeds the maximum size).} For instance, $A_{20{\%}}$ is computed by averaging five accuracies, each computed over 100 windows containing $m_w$ samples of which $20{\%}$ are drifted. 
However, since the predictions of drifted and non-drifted windows are closely related, a technique that always predicts drift is unreliable. Therefore, we compute the \textit{Harmonic Drift Detection} ($H_{DD}$), as the harmonic mean between the accuracy for non-drifted windows ($A_{0\%}$) and the average accuracy across windows with different drift severities~($\bar{A}_{\text{drift}}$):
\begin{equation}
\small
    H_{DD} = \frac{2}{\frac{1}{A_{0\%}} + \frac{1}{\bar{A}_{\text{drift}}}}
\end{equation}
\vspace{-2mm}
\noindent Where:
\begin{equation}
\small
    \bar{A}_{\text{drift}} = \frac{A_{5\%} + A_{10\%} + A_{15\%} + A_{20\%}}{4}
\end{equation}

\begin{figure*}[!h]
  \centering \includegraphics[width=0.982\textwidth]{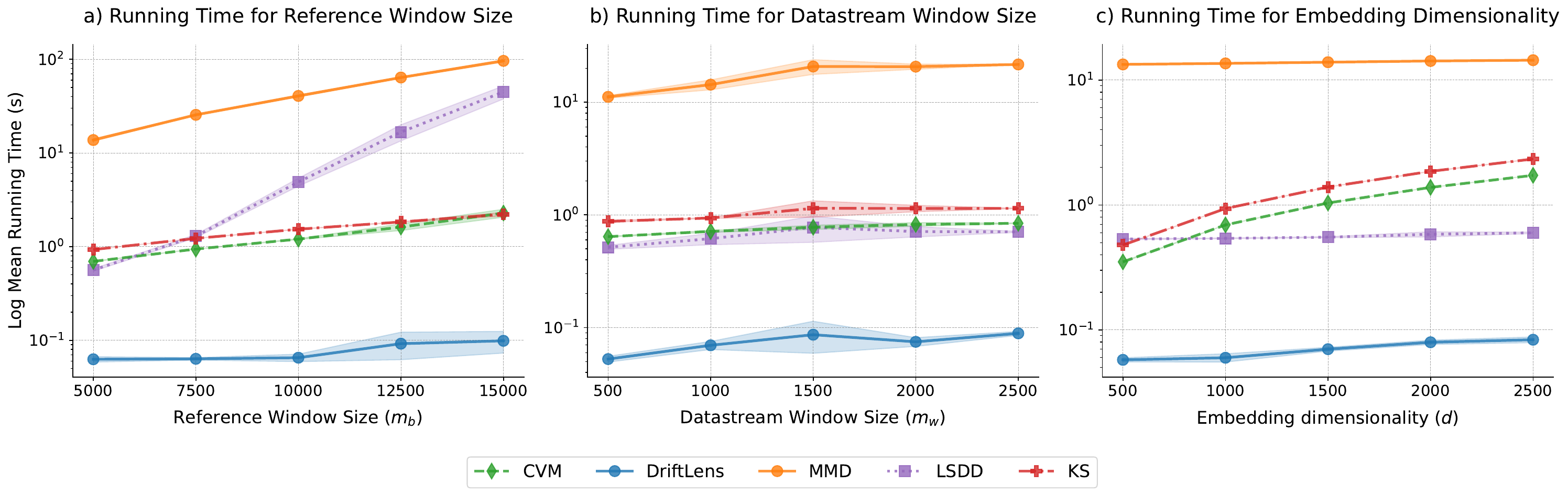}
  \caption{\textbf{Running time evaluation.} 
  Mean (\(\pm\) std) of running time (log scale) for predicting drift with each detector, while varying one dimension at a time: (a) reference window size \( m_b \), (b) data stream window size \( m_w \), and (c) embedding dimensionality~\( d \). The other parameters are fixed at \( m_b = 5{,}000 \), \( m_w = 1{,}000 \), and \( d = 1{,}000 \). Mean and std are computed over five runs.
  }
  \label{fig:running-time-comparison-log}
\end{figure*}

\smallskip
\noindent
\textbf{Results} \ 
Table~\ref{tab:drift-detection-experiments} reports: (i)~the mean \textit{accuracy} broken down by drift percentage ($D_{\%}$), and (ii) the $H_{DD}$ score, for each use case, drift detector, and window size ($m_w$). 
Results within the same use case but with a different classifier (marked with~* in Table~\ref{tab_uses_cases}) are reported in Appendix \ref{apx:more-experiment}. 
Table~\ref{tab:drift-detection-experiments-larger} uses larger data windows than Table~\ref{tab:drift-detection-experiments-smaller} as datasets contain more samples. 

For \textit{larger} data volumes (Table~\ref{tab:drift-detection-experiments-larger}), \XXX\ achieves better drift prediction performance on all use cases independently of the data type and window size, except for use cases 9 with $m_w\!=\!500$ and 4 with $m_w\!=\!2000$. For use cases 1.1 and~4, it is particularly effective in detecting drift, with $H_{DD}\!\geq\!~0.93$.  
\XXX\ is the only detector that consistently achieves reliable results across all use cases ($H_{DD}\!\geq\!0.38$) and that performs reliably in the audio domain (use case~10). Notably, it is capable of detecting drift and outperforming other detectors, even when drift manifests as classifier bias (use cases 4, 9,~10).

In these use cases with larger datasets and window sizes, \textsc{DriftLens} achieves an average $H_{DD}$ score of $0.82$, outperforming MMD ($0.57$), KS ($0.56$), LSDD ($0.57$), and CVM ($0.55$). Therefore, 
it significantly outperforms the best detector by an average margin of $0.25$ in $H_{DD}$. Results are consistent across deep learning classifiers (Table \ref{tab:drift-detection-experiments-appendix-larger} in Appendix \ref{apx:more-experiment}). 
Overall, \XXX\ is the best detector on 8 out of 8 experimental use cases with larger datasets and window sizes.

For \textit{smaller} data volume (Table~\ref{tab:drift-detection-experiments-smaller}), \XXX\ is still the most effective overall, except for use case 8, in which MDD and LSDD achieve slightly better $H_{DD}$. 
Interestingly, all detectors detect drift simulated with blur (use case 8). However, KS and CVM exhibit a large number of false positives, especially for larger window sizes. 
Notably, \XXX\ ($H_{DD} \geq 0.81$) and LSDD ($H_{DD} \geq~0.17$) are the only detectors consistently achieving reliable performance across all use cases and window sizes. In contrast, the other detectors exhibit some use cases with unreliable performance ($H_{DD} = 0$).

In these use cases with smaller datasets and window sizes, \XXX\ achieves an average $H_{DD}$ score of $0.92$, outperforming MMD ($0.69$), KS ($0.42$), LSDD ($0.66$), and CVM ($0.40$). Therefore, it significantly outperforms the best detector by $0.23$ in $H_{DD}$. Results are consistent across deep learning classifiers (refer to Table \ref{tab:drift-detection-experiments-appendix-smaller} in Appendix \ref{apx:more-experiment}). 
Overall, \XXX\ is the best detector on 7 out of 9 experimental use cases with smaller datasets and window sizes.

\smallskip
\noindent
\textbf{Summary of findings} \ \XXX\ is highly effective in detecting drift (RQ1 in \S\ref{sec:intro}), outperforming other detectors in 15/17 use cases. It is the only reliable method across all use cases, making it the most effective and widely applicable across models, data types, and data volumes (RQ2 in \S\ref{sec:intro}).

\begin{figure}
  \centering \includegraphics[width=0.495\textwidth]{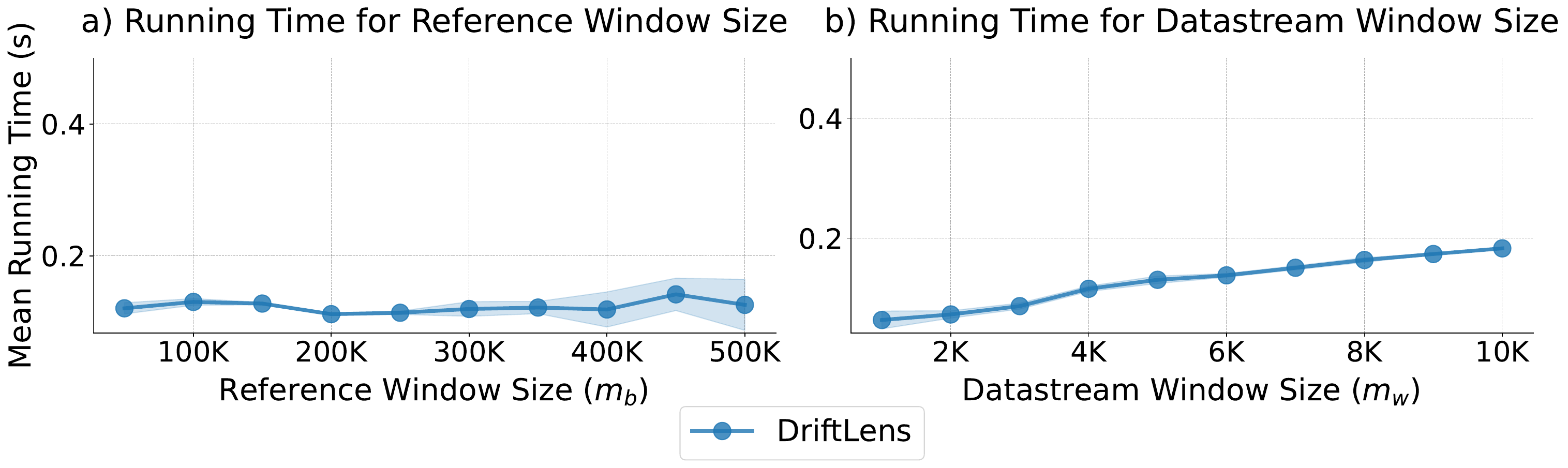}
  \caption{\XXX\ mean running time in seconds as the reference $m_b$ (a) and data stream $m_w$ (b) window sizes change.}
  \label{fig:running-time-driftlens}
\end{figure}

\subsection{Drift Detection Complexity Evaluation} 
\label{subsec:evaluation-complexity}
This evaluation assesses the efficiency of \XXX\ in performing near real-time drift detection (RQ3 in~\S\ref{sec:intro}).

\smallskip
\noindent
\textbf{Evaluation Metrics} \ We measure the mean running time in seconds required to predict drift, 
given the embeddings extracted.
Experiments are executed on an Apple M1 MacBook Pro 13 2020 with 16GB of RAM. Due to the high dimensionalities tested, we use synthetically generated embeddings.

\smallskip
\noindent
\textbf{Results} \ Figure~\ref{fig:running-time-comparison-log} shows the running time in seconds on a logarithmic scale of each 
detector by varying (a) the reference window size $m_b$, (b) the data stream window size $m_w$, and (c)~the embedding dimensionality $d$.  
One dimension at a time is varied while keeping the other fixed at: $m_w\!=\!1{,}000$, $d\!=\!1{,}000$, and  $m_b\!=\!5{,}000$. 
\XXX\ outperforms all 
detectors in terms of efficiency, running at least 5 times faster. As the dimensionalities increase, 
its execution time increases negligibly, unlike other techniques that are significantly affected by the window sizes and embedding dimensionality.

Figure \ref{fig:running-time-driftlens} shows \XXX' running time when dealing with large data volumes. The reference window is increased up to $500{,}000$ samples, and the data stream window to $10{,000}$. The other detectors do not work with these dimensionalities on the experimental hardware. When varying the reference window size $m_b$, the data stream window size is kept fixed to $m_w\!=\!5{,}000$. 
When varying the data stream window size~$m_w$, the reference window size is fixed to $m_b\!=\!500{,}000$. Figure~\ref{fig:running-time-driftlens}\mbox{-(a)} confirms that its running time is almost independent of the reference window size, as in the online phase, only the pre-computed reference distributions are used ($\mu_b$, $\mu_{b,l}$ and $\Sigma_b$, $\Sigma_{b,l}$ with dimensionalities $d'$, $d'_l$). 
Figure~\ref{fig:running-time-driftlens}-(b) shows that its running time increases almost negligibly with data stream window size, enabling real-time 
detection in high-throughput data streams. Notably, the running time is always $\leq0.2s$.

\smallskip
\noindent
\textbf{Summary of findings} \ \XXX\ is the fastest detector, enabling real-time drift detection regardless of data volumes in the reference set or data stream (answering RQ3 in \S\ref{sec:intro}).

\begin{figure*}
    \centering
    \begin{subfigure}[b]{0.328\textwidth}
        \centering
        \includegraphics[width=\textwidth]{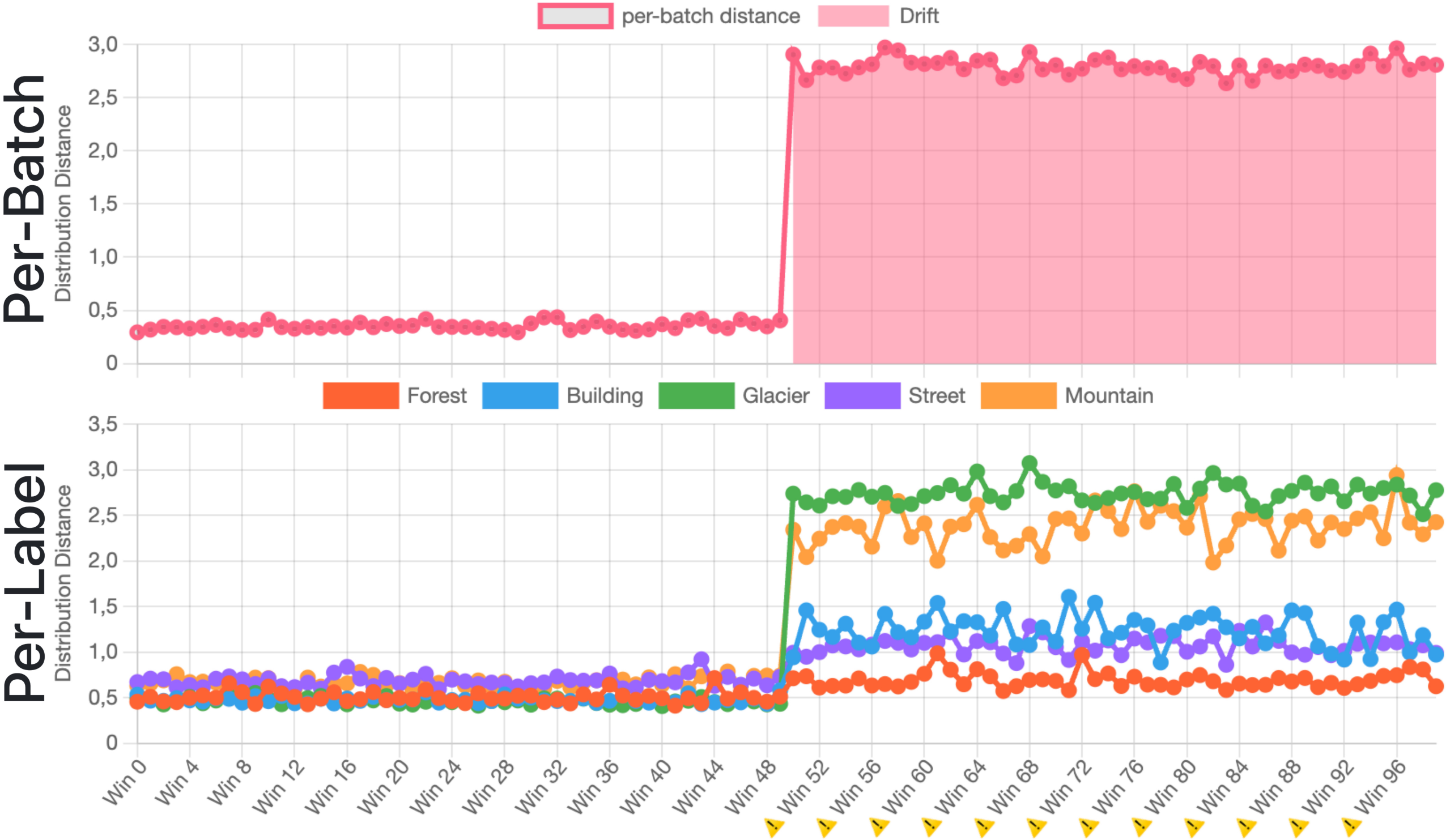}
        \caption{Sudden drift.}
        \label{fig:sub1}
    \end{subfigure}
    \hfill 
    \begin{subfigure}[b]{0.328\textwidth}
        \centering
        \includegraphics[width=\textwidth]{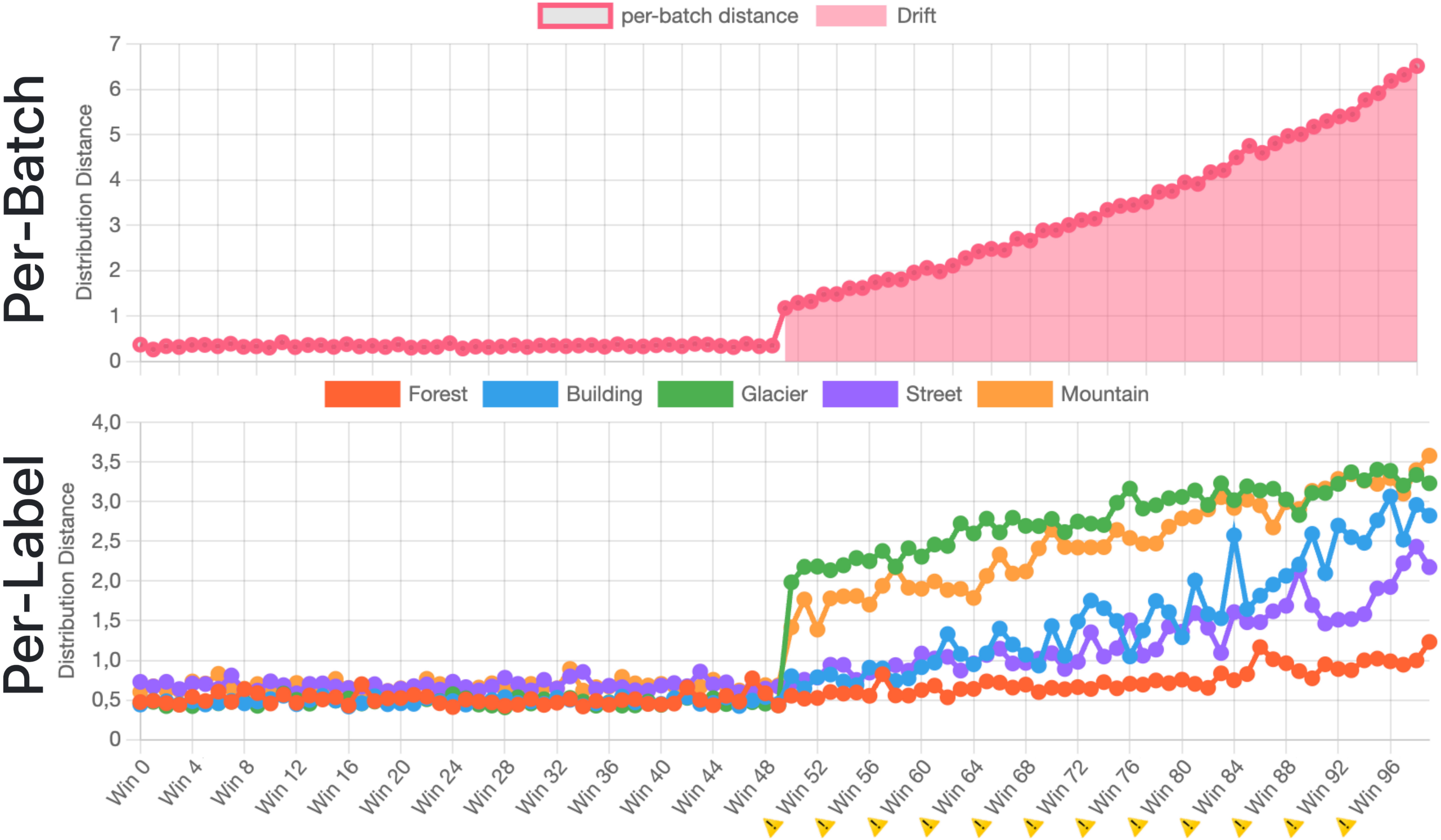}
        \caption{Incremental drift.}
        \label{fig:sub2}
    \end{subfigure}
    \hfill 
    \begin{subfigure}[b]{0.328\textwidth}
        \centering
        \includegraphics[width=\textwidth]{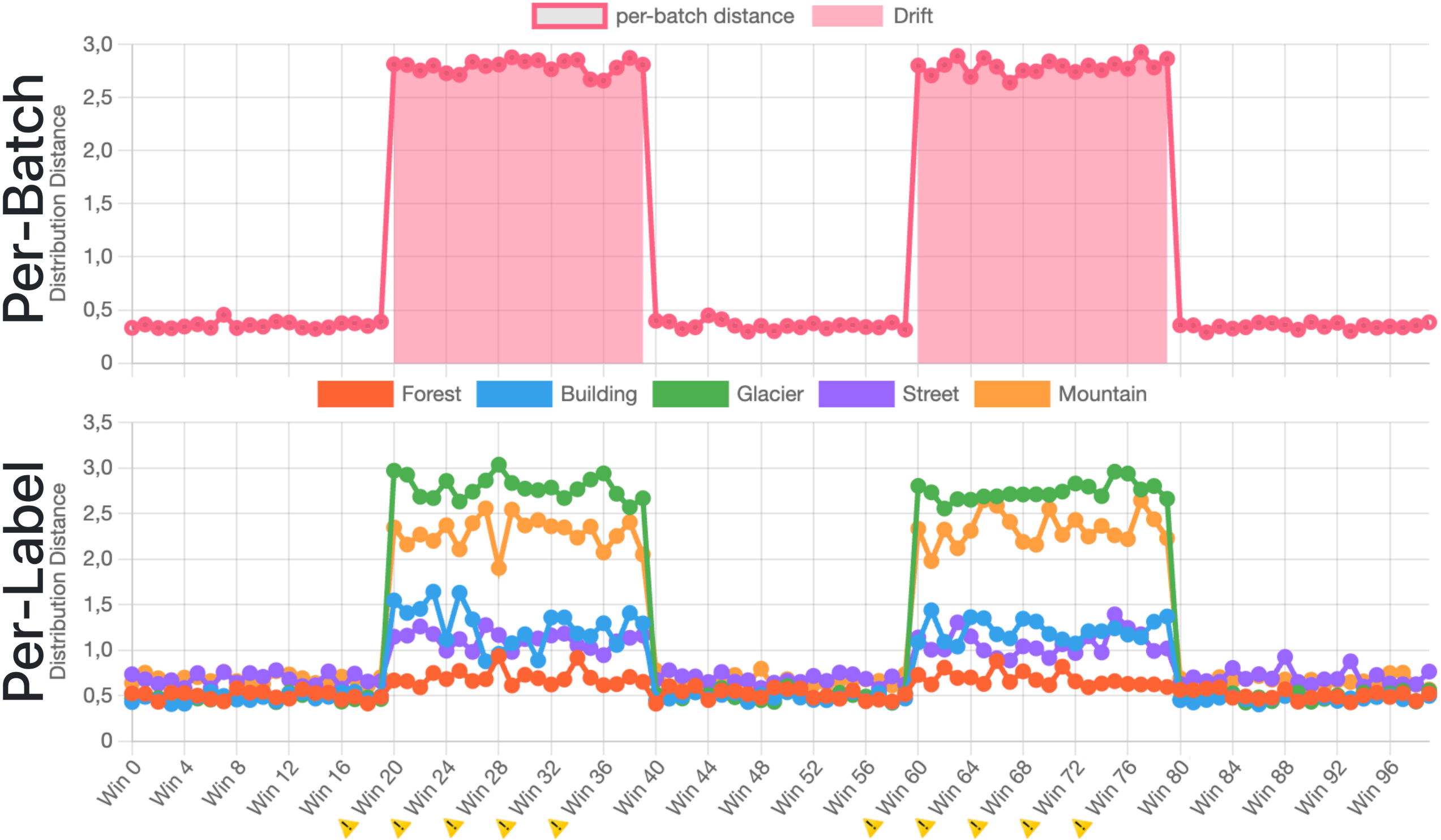}
        \caption{Periodic drift.}
        \label{fig:sub3}
    \end{subfigure}
    \caption{Drift patterns qualitative evaluation for \textit{use case 6.1} (ViT - Intel Image). Drift is simulated with a new class (\textit{Sea}).}
    \label{fig:drift-patterns-usecase6}
\end{figure*}

\begin{figure}
  \centering \includegraphics[width=0.49\textwidth]{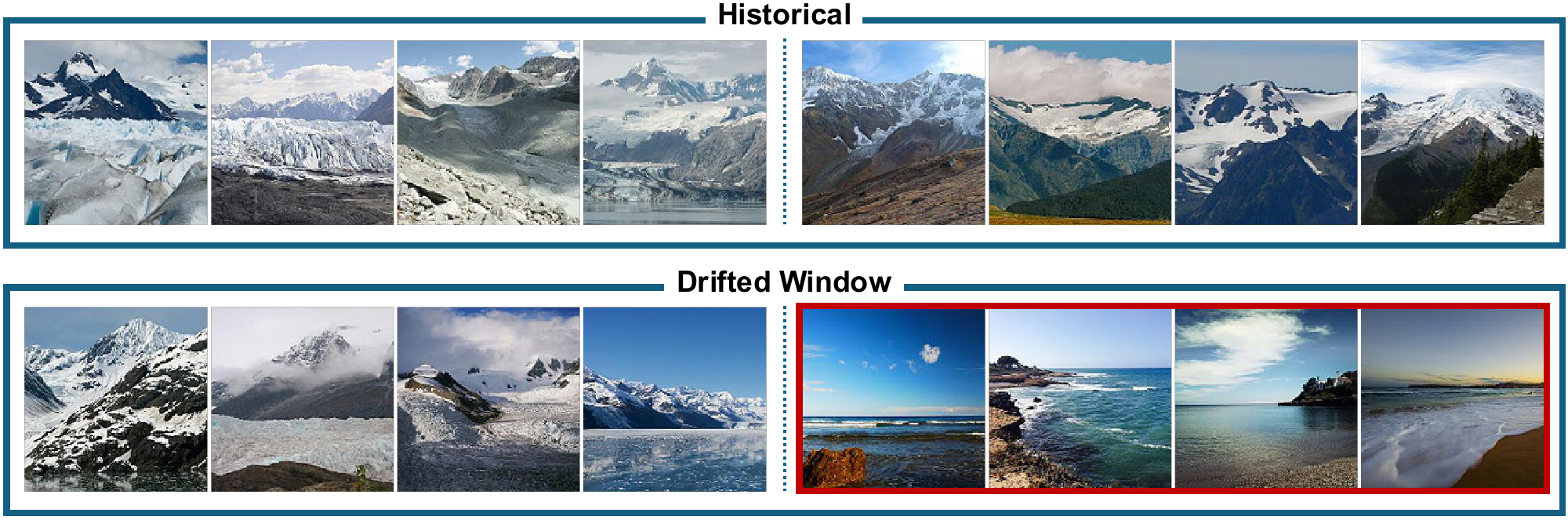}
  \caption{Drift explanation for class \textit{Glacier} in use case 6.1.}
  \label{fig:explanation-1-glaciar}
\end{figure}

\subsection{Drift Curves Evaluation}
\label{subsec:evaluation-characterization}
This evaluation assesses the ability of \XXX\ to represent and characterize the drift curve over time (RQ4 in \S\ref{sec:intro}).

\smallskip
\noindent
\textbf{Evaluation Metrics} \  We use the Spearman coefficient~\cite{ref1} to measure the correlation between the \textit{per-batch} ($FDD$) distances over time and the curve of injected drift. 
This coefficient evaluates the monotonic relationship between two variables---they tend to move in the same direction---and is suitable for non-linear patterns (e.g., sudden or periodic changes). 
It ranges from -1 (perfect negative) to +1 (perfect positive), with 0 indicating no monotonic relationship. The injected drift curve is composed of 0 in windows without drift, and the percentage of drift ($D_{\%}$) in windows containing some drift. We also qualitatively evaluate examples of drift curves.

\smallskip
\noindent
\textbf{Quantitative Results} \ Table \ref{tab:drift-patterns-evaluation} reports the mean (\(\pm\) std) Spearman correlation, computed across all use cases and averaged over 5 runs. Data streams are generated by randomly sampling 100 windows, each with $m_w\!=\!1{,}000$ samples. 
In the sudden pattern, drift occurs after 50 windows and remains constant with a percentage of $D_{\%}\!=\!40\%$. In the incremental pattern, drift occurs after 50 windows with $D_{\%}\!=\!20\%$ and increases by $\Delta D_{\%}\!=\!1\%$ after each window.  In the periodic pattern, 20 windows without drift and 20 windows containing $D_{\%}\!=\!40\%$ of drift reoccur periodically.  
Table \ref{tab:drift-patterns-evaluation} reveals that the \textit{per-batch} ($FDD$) distances are highly correlated with the generated drift patterns. The 
correlation exceeds $0.85$ for all drift patterns, demonstrating the ability of \XXX\ 
to correctly characterize and model the drift trend over time.

\begin{figure}
  \centering \includegraphics[width=0.49\textwidth]{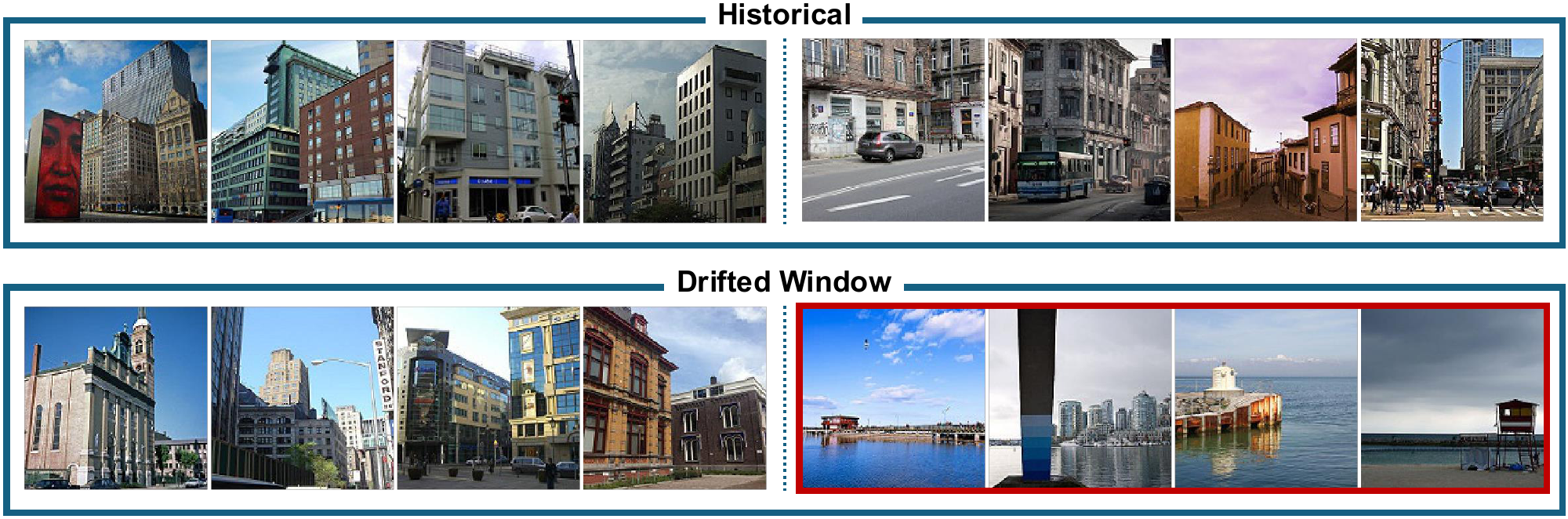}
  \caption{Drift explanation for class \textit{Building} in use case 6.1.}
  \label{fig:explanation-1-building}
\end{figure}

\begin{table}
\small
\centering
\caption{\textbf{Drift curve evaluation.} Spearman correlation between drift severity and \textit{per-batch} distribution distance ($FDD$), per drift pattern, averaged across all use cases.}
\label{tab:drift-patterns-evaluation}
\scalebox{0.9}{
\begin{tabular}{cccc}
\toprule
\textbf{Drift Pattern}  & \textit{Sudden} & \textit{Incremental} & \textit{Recurrent} \\ \midrule
\textit{Avg. Spearman Corr.}  & $0.88 \pm .02$ & $0.99 \pm .01$ & $0.85 \pm .00$ \\ \midrule
\end{tabular}}
\end{table}

\smallskip
\noindent
\textbf{Qualitative Results} \ Figure \ref{fig:drift-patterns-usecase6} shows the generated monitors for the (a) \textit{sudden}, (b) \textit{incremental}, and (c) \textit{periodic} 
patterns in use case 6.1, with the previous settings. The \textit{per-batch} and \textit{per-label} drift curves are coherent with the generated patterns. 
The labels \textit{Glacier} (green) and \textit{Mountain} (orange) are the most affected by drift, likely due to the newly injected class (\textit{Sea}) being classified under these labels. 
In contrast, \textit{Building} (blue) and \textit{Street} (purple) show moderate impact, and \textit{Forest} (red) is nearly unaffected.
The drift curves provide valuable insights to characterize drift by showing the most impacted labels, and the drift pattern. Additional examples are reported in Appendix~\ref{apx:drift-characterization}.

\smallskip
\noindent
\textbf{Summary of findings} \ \XXX\ generates drift curves over time that closely align with drift severity, and the \textit{per-label} curves help characterize drift (partially answering RQ4 in~\S\ref{sec:intro}).

\subsection{Drift Explanation Evaluation}
\label{subsec:evaluation-explanations}
This evaluation assesses the effectiveness of \XXX\ in generating \textit{per-label} drift explanations (RQ4 in \S\ref{sec:intro}). 

\smallskip
\noindent
\textbf{Evaluation Metrics} \  The \textit{per-label} drift explanations aim to identify 
prototype samples that caused drift in a label. To extract drifted prototypes from clusters' centroids, clustering must separate non-drifted samples into distinct groups while forming one or more predominantly drifted clusters. Therefore, we measure the \textit{purity} of clustering, which quantifies the extent to which each cluster contains samples from a single class. In our context, instead of the class label, we use two categories: (1) drifted and (2) non-drifted samples. 
Given a set $X_{l}=X_{l}^{D} \cup X_{l}^{N}$ of $m_{l}$ samples predicted with the label~$l$, where $X_{l}^{D}$ are the drifted and $X_{l}^{N}$ the non-drifted samples, the K-Means  divides $X_{l}$ into $k$ clusters. Purity is computed~as: 
\vspace{-2mm}
\begin{equation}
\label{eq:purity}
\small
\text{Purity} = \frac{1}{m_l} \sum_{i=1}^{k} \max \left( |C_i \cap X_{l}^{D}|, |C_i \cap X_{l}^{N}| \right)
\end{equation}
Where \( C_i \) is the set of samples in cluster \( i \),
the terms \( |C_i \cap X_{l}^{D}| \) and \( |C_i \cap X_{l}^{N}| \) represent the number of drifted and non-drifted samples in cluster \( i \), 
and \( \max (|C_i \cap X_{l}^{D}|, |C_i \cap X_{l}^{N}|) \) selects the majority category (drift or no-drift) in each cluster.
High purity suggests that clustering successfully isolates drifted samples into one or more clusters, ensuring that prototypes extracted from cluster centroids provide meaningful drifted examples. We also qualitatively discuss examples of drift explanations.

\smallskip
\noindent
\textbf{Quantitative Results} \ Table \ref{tab:drift-explanations-purity} reports the clustering purity for the \textit{per-label} drift explanations. We focus on the label most affected by drift---where drifted samples are more likely to be classified. Specifically, we analyze use cases (label) 1.1 (\textit{World}), 2.1 (\textit{Science}), 4 (\textit{Nurse}), 6.1 (\textit{Glacier}), and 7.1 (\textit{Ship}), based on findings in Appendix \ref{apx:drift-characterization}.
For each drift severity $D_{\%}\!\in\!\{10\%, 15\%, 20\% \}$ and window size $m_w\!\in\!\{1000, 2000, 4000 \}$, purity is calculated and averaged over 100 windows. In use cases 2.1 and 7.1, there are not enough samples for $m_w\!=\!4000$. K-Means is executed multiple times for $k \in [2, 10]$ with the best $k$ selected based on the Silhouette. 
Table \ref{tab:drift-explanations-purity} shows good clustering performance in all use cases and drift levels, with purity ranging from $0.73$ to $0.92$.  These results highlight that clustering effectively isolates normal and drifted samples, which is crucial to identifying drifted prototypes. 

\begin{table}
\caption{\textbf{Drift explanation evaluation.} \textit{Purity} score for \textit{per-label} clustering of the most affected label, separately per window size $m_w$ and drift percentage $D\!\in\!\{10\%,15\%,20\%\}$.}
\label{tab:drift-explanations-purity}
\centering
\small
\scalebox{0.72}{
\begin{tabular}{ccccclccclccc}
\toprule
\multicolumn{1}{c}{\multirow{3}{*}{\begin{tabular}[c]{@{}c@{}} \textbf{Use} \\ \textbf{case} \end{tabular}}} & \multirow{3}{*}{\textbf{Label}} & \multicolumn{3}{c}{$m_w = 1000$} & & \multicolumn{3}{c}{$m_w = 2000$} & & \multicolumn{3}{c}{$m_w = 4000$} \\
\multicolumn{1}{c}{}                                                        &             & 10\%  & 15\%  & 20\%  & & 10\%  & 15\%  & 20\% & & 10\%  & 15\%  & 20\%  \\
\cline{3-5} \cline{7-9} \cline{11-13} 
 & & & & & & & & & & \\
\toprule 
\textit{1.1} & \textit{World} & 0.92  & 0.88  & 0.85 &  & 0.92  & 0.88  & 0.84 & & 0.92  & 0.88  & 0.84 \\ 
\textit{2.1} & \textit{Science} & 0.84  & 0.82  & 0.80 &  & 0.83  & 0.82  & 0.81 & & -  & -  & - \\
\textit{4} & \textit{Nurse} & 0.80  & 0.76  & 0.73 &  & 0.80  & 0.76  & 0.73 & & 0.81  & 0.77  & 0.74 \\
\textit{6.1} & \textit{Glacier} & 0.80  & 0.79  & 0.78 &  & 0.79  & 0.79  & 0.78 & & 0.79  & 0.79  & 0.78 \\
\textit{7.1} & \textit{Ship} & 0.87  & 0.82  & 0.79 &  & 0.87  & 0.83  & 0.79 & & -  & -  & - \\
\bottomrule
\end{tabular}}
\end{table}

\smallskip
\noindent
\textbf{Qualitative Results} \
Figures \ref{fig:explanation-1-glaciar} and \ref{fig:explanation-1-building} show the \textit{per-label} drift explanations for use case 6.1, 
for the classes \textit{Glacier} (highly affected) and \textit{Building} (moderately affected), 
generated with 
$m_w\!=\!2000$ and $20\%$ of drift. 
The closest 4 samples to the centroids are extracted as prototypes.
In Figure~\ref{fig:explanation-1-glaciar}, historical prototypes for \textit{Glacier} are represented by two clusters: winter glaciers (top-left) and summer glaciers (top-right). 
When drift occurs, the drifted window is characterized by two clusters: glacier during winter and summer (bottom-left) and general sea images, which are responsible for drift (bottom-right). Similarly, in Figure \ref{fig:explanation-1-building}, \textit{historical} prototypes for the class \textit{Building} are characterized by two clusters of images: buildings only (top-left) and over streets (top-right). When drift occurs, the drifted window is characterized by two clusters: general buildings (bottom-left) and buildings over the sea, which are responsible for drift (bottom-right). These prototypes can help humans understand the drift nature, 
facilitating the adaptation.
Additional qualitative examples are discussed in Appendix~\ref{apx:drift-characterization}.

\smallskip
\noindent
\textbf{Summary of findings} \XXX\ provides explanations useful to characterize and understand drift (answering RQ4~in~\S\ref{sec:intro}).

\section{Conclusion}
This paper presents \XXX, an unsupervised drift detection and characterization framework to continuously monitor drift in deep learning classifiers for unstructured, unlabeled data streams. 
Our evaluation shows that (i) it is more effective and efficient in detecting drift than state-of-the-art techniques, (ii) it has a short execution time, enabling real-time 
detection, and  (iii) it effectively represents the drift trend over time while characterizing and explaining drifting labels.  
We release \XXX\ as an open-source framework broadly applicable across classifiers and data types. Its \textit{per-batch} analysis is versatile and can extend beyond classification, while the \textit{per-label} analysis is specific for classification or can be adapted to other label-based tasks, such as clustering and object detection.
 
\noindent \textbf{Limitations} \
\XXX\ uses the Frechét distance and it can inherit some of its limitations.
(1)~\textit{Noise and selection bias with small datasets}. 
$FDD$ can be affected by noise and selection bias in distance calculation with small datasets. However, we empirically show that \XXX\ performs well even with smaller reference data and window sizes.
(2)~\textit{Limited number of statistics}. 
The $FDD$ relies on a limited set of statistics---mean and covariance---for distance calculation, covering only the first two moments of the distribution and not considering higher moments like skewness and kurtosis.
(3)~\textit{Distance interpretability.} 
The $FDD$ distances are in the $[0, \infty]$ range, but would be more interpretable in the $[0,1]$ range. 
This can be addressed by expressing the distance relative to the threshold.
(4)~\XXX\ \textit{assumes embeddings follow Gaussian distributions}, simplifying mathematical modeling but potentially overlooking nuances in the actual distributions. However, our experiments show that this assumption does not affect drift detection performance while significantly enhancing efficiency.
(5) We experimented with drift in windows with nearly balanced labels, but this assumption may not always hold. 
 
\noindent In \textbf{future work}, we plan to:
(i) integrate drift explanations with data type-specific XAI techniques based on embedding representations or concepts \cite{ventura2022,ventura2023explaining,poeta2023conceptbased},
(ii) address automatic drift adaptation in unsupervised or supervised with limited label, 
(iii) test drift detectors in scenarios with unbalanced data, and
(iv) extend the approach to tasks other than classification.

\bibliographystyle{IEEEtran}
\bibliography{sample}

\vspace{-2mm}
\section{Biography Section}
 




\vspace{-35pt}
\begin{IEEEbiographynophoto}{Salvatore Greco} is a Research Associate at Politecnico di Torino, where he obtained a Ph.D. in computer engineering in 2024. His research interests include explainability, fairness, and robustness in Natural Language Processing.
\end{IEEEbiographynophoto} 

\vspace{-34pt}
\begin{IEEEbiographynophoto}{Bartolomeo Vacchetti} has a PhD in Creativity-Injection into AI-Powered Multimedia Storyboards at Politecnico di Torino. His research focuses on integrating machine and deep learning techniques to image and video editing.
\end{IEEEbiographynophoto}

\vspace{-33pt}
\begin{IEEEbiographynophoto}{Daniele Apiletti} is an Associate Professor at Politecnico di Torino, Italy. He is a spin-off founder, head of the Data Science Engineering Master program, and Steering Committee member of the SmartData@PoliTo research center. His research interests include data science techniques to extract value from data by combining data-driven algorithms with knowledge-based models. He leads funded projects, collaborates with companies, and organizes conferences. 
\end{IEEEbiographynophoto}

\vspace{-32pt}
\begin{IEEEbiographynophoto}{Tania Cerquitelli} is a Full Professor at 
the Politecnico di Torino, Italy. 
She is responsible for aggregate functions to the Deputy Rector for Society, Community, and Program Delivery. 
Her research interests include explainable artificial intelligence, algorithms to democratize data science, and algorithms to detect concept drift early. 
She is a member of the editorial board of several Elsevier journals (i.e., Computer Networks; Future Generation Computer Systems; Expert System with Applications; Engineering Applications of Artificial Intelligence) and Knowledge and Information Systems (Springer). 
She has obtained research funding from the EU, the Piedmont Region, the Ministry of University and Research, and private companies.
\end{IEEEbiographynophoto}

\vfill


\clearpage


\appendix

\renewcommand{\thesection}{\Alph{section}}
The appendix sections are organized as follows. Appendix~\ref{apx:more-experiment} further assesses the drift detection performance of \XXX\ across deep learning classifiers. Appendix~\ref{apx:drift-characterization} further evaluates drift characterization by discussing additional examples of drift trends over time and drift explanations. Appendix~\ref{apx:evaluation-sensitivity} performs a parameter sensitivity evaluation. Appendix~\ref{apx:evaluation-distance-metric} presents an ablation study of the distribution distance metric.

\begin{table*}[h!]
\scriptsize
\centering
\caption{\textbf{Drift detection performance evaluation.} For \XXX, \textsc{MMD} \cite{Rabanser2018FailingLA}, \textsc{KS}~\cite{Rabanser2018FailingLA}, \textsc{LSDD}~\cite{7745962}, \textsc{CVM} \cite{cramer1928composition} detectors \cite{alibi-detect}, and each window size $m_w$ are reported (i) the \textit{accuracy} separately per drift percentage $D_{\%} \in \{0\%, 5\%, 10\%, 15\%, 20\%\}$, and (ii) the \textit{Harmonic Drift Detection} mean $H_{DD}$. Accuracy is computed over 100 windows and averaged by repeating 5 runs. The best-performing detector for each use case based on the overall $H_{DD}$ is highlighted, and per window size is in~bold.}
\label{tab:drift-detection-experiments-appendix}
\begin{subtable}[t]{\textwidth}
\centering
\caption{\textbf{Larger data volumes:} Data stream window sizes $m_w \in \{500, 1000, 2000\}$}
\label{tab:drift-detection-experiments-appendix-larger}
\scalebox{.85}{
\begin{tabular}{clclllllllllllllllllllll}
\toprule
\multirow{5}{*}{\textbf{Use}} & \textbf{} & \multirow{5}{*}{\textbf{Drift}} &  & \multicolumn{20}{c}{} \\
 & & & & \multicolumn{6}{c}{\textit{$m_w = 500$}} & \multicolumn{1}{c}{\textit{}} & \multicolumn{6}{c}{\textit{$m_w = 1000$}} & \multicolumn{1}{c}{\textit{}} & \multicolumn{6}{c}{\textit{$m_w = 2000$}} \\ \cline{5-10} \cline{12-17} \cline{19-24} 
 & & &  & & & & & & & & & & & & & & & & & & \\
 
 \textbf{Case} & & \multicolumn{1}{c}{\textbf{Detector}} &  & \multicolumn{5}{c}{\textbf{Drift Percentage $D_{\%}$}} &  & \multicolumn{1}{c}{\textbf{}} &  \multicolumn{5}{c}{\textbf{Drift Percentage $D_{\%}$}} & \multicolumn{1}{c}{\textbf{}} & & \multicolumn{5}{c}{\textbf{Drift Percentage $D_{\%}$}} \\
 & & & & 0\% & 5\% & 10\% & 15\% & 20\% & $H_{DD}$ & & 0\% & 5\% & 10\% & 15\% & 20\% & $H_{DD}$ & & 0\% & 5\% & 10\% & 15\% & 20\% & $H_{DD}$ \\ \cline{5-9} \cline{12-16} \cline{19-23} 
 & & & & & & & & & & & & & & & & & & & & & & \\ \toprule
\multirow{2}{*}{1.2} 
 & & \multicolumn{1}{l}{MMD} & & 1.00 & 0.00  & 0.05 & 0.91 & 1.00 & \textit{0.66} & & 1.00 & 0.00 & 0.78  & 1.00 & 1.00 & \textit{0.82} & & 1.00 & 0.03 & 1.00 & 1.00 & 1.00 & \textit{0.86} \\
 & & \multicolumn{1}{l}{KS} & & 1.00 & 0.01 & 0.14 & 0.93 & 1.00 & \textit{0.68} & & 1.00 & 0.03 & 0.87 & 1.00 & 1.00   & \textit{0.84} & & 1.00 & 0.25 & 1.00 & 1.00 & 1.00 & \textit{0.89} \\
AG News & & \multicolumn{1}{l}{LSDD} & & 1.00 & 0.00 & 0.06 & 0.90 & 1.00 & \textit{0.66} & & 1.00 & 0.00 & 0.80 & 1.00 & 1.00 & \textit{0.82} & & 1.00 & 0.03 & 0.99 & 1.00 & 1.00 & \textit{0.86} \\
\multirow{2}{*}{DistilBERT} & & \multicolumn{1}{l}{CVM} & & 1.00 & 0.01 & 0.15 & 0.97 & 1.00 & \textit{0.69} & & 1.00 & 0.04 & 0.93 & 1.00 &   1.00 & \textit{0.85} & & 0.99 & 0.31 & 1.00 & 1.00 & 1.00  & \textit{0.90} \\
 & & \multicolumn{1}{l}{\cellcolor{mycellcolor}\XXX} & \cellcolor{mycellcolor} & \cellcolor{mycellcolor}0.99 & \cellcolor{mycellcolor}0.71 & \cellcolor{mycellcolor}1.00 & \cellcolor{mycellcolor}1.00 & \cellcolor{mycellcolor}1.00 & \cellcolor{mycellcolor}\textbf{\textit{0.96}} & \cellcolor{mycellcolor} & \cellcolor{mycellcolor}0.99 & \cellcolor{mycellcolor}0.98 & \cellcolor{mycellcolor}1.00 & \cellcolor{mycellcolor}1.00 & \cellcolor{mycellcolor}1.00 & \cellcolor{mycellcolor}\textit{\textbf{0.99}}  & \cellcolor{mycellcolor} & \cellcolor{mycellcolor}0.99 & \cellcolor{mycellcolor}1.00 & \cellcolor{mycellcolor}1.00 & \cellcolor{mycellcolor}1.00 & \cellcolor{mycellcolor}1.00  &  \cellcolor{mycellcolor}\textit{\textbf{0.99}} \\ \midrule
\multirow{2}{*}{1.3} 
 & & \multicolumn{1}{l}{MMD} &  & 1.00 & 0.00 &  0.18 & 0.98 & 1.00 & \textit{0.70} & & 1.00 & 0.01 &  0.94 & 1.00   & 1.00 & \textit{0.85} & & 1.00 & 0.09 & 1.00 & 1.00 & 1.00 & \textit{0.87} \\
 & & \multicolumn{1}{l}{KS} &  & 1.00 &  0.01 & 0.16 &  0.92 & 1.00 & \textit{0.69} & & 1.00 & 0.02 & 0.85 & 1.00 & 1.00 & \textit{0.84} & & 1.00 & 0.18 & 1.00 & 1.00 & 1.00 & \textit{0.89} \\
AG News & & \multicolumn{1}{l}{LSDD} & & 1.00 & 0.00 & 0.19 & 0.99 & 1.00 & \textit{0.71} & & 1.00 & 0.01 & 0.98 & 1.00 &  1.00 & \textit{0.86} & & 1.00 & 0.26 & 1.00 & 1.00 & 1.00 & \textit{0.90} \\
\multirow{2}{*}{RoBERTa} & & \multicolumn{1}{l}{CVM} & & 1.00 & 0.00 & 0.15 & 0.96 & 1.00 & \textit{0.69} & & 1.00 & 0.02 & 0.90 & 1.00 & 1.00 & \textit{0.84} & & 1.00 & 0.16 & 1.00 & 1.00 & 1.00 & \textit{0.88} \\
 & & \multicolumn{1}{l}{\cellcolor{mycellcolor}\XXX} & \cellcolor{mycellcolor}  & \cellcolor{mycellcolor}1.00 & \cellcolor{mycellcolor}0.09 & \cellcolor{mycellcolor}0.98 & \cellcolor{mycellcolor}1.00 & \cellcolor{mycellcolor}1.00 & \cellcolor{mycellcolor}\textbf{\textit{0.87}}  & \cellcolor{mycellcolor} & \cellcolor{mycellcolor}1.00 & \cellcolor{mycellcolor}0.47 & \cellcolor{mycellcolor}1.00 & \cellcolor{mycellcolor}1.00 & \cellcolor{mycellcolor}1.00 & \cellcolor{mycellcolor}\textbf{\textit{0.93}} & \cellcolor{mycellcolor} & \cellcolor{mycellcolor}1.00 & \cellcolor{mycellcolor}0.96 & \cellcolor{mycellcolor}1.00 & \cellcolor{mycellcolor}1.00 &  \cellcolor{mycellcolor}1.00 & \cellcolor{mycellcolor}\textbf{\textit{1.00}}  \\ \midrule
\multirow{2}{*}{5.2} 
 & & \multicolumn{1}{l}{MMD} &  & 1.00 & 0.01 &  0.16 & 0.86 & 1.00 & \textit{0.67} & & 1.00 & 0.01 &  0.72 & 1.00   & 1.00 & \textit{0.81} & & 1.00 & 0.06 & 1.00 & 1.00 & 1.00 & \textbf{\textit{0.87}} \\
 & & \multicolumn{1}{l}{KS} &  & 1.00 &  0.00 & 0.07 &  0.44 & 0.95 & \textit{0.54} & & 1.00 & 0.01 & 0.36 & 0.99 & 1.00 & \textit{0.74} & & 1.00 & 0.08 & 0.98 & 1.00 & 1.00 & \textbf{\textit{0.87}} \\
MNIST & & \multicolumn{1}{l}{LSDD} & & 0.97 & 0.04 & 0.09 & 0.35 & 0.92 & \textit{0.51} & & 0.97 & 0.05 & 0.28 & 0.95 &  1.00 & \textit{0.72} & & 0.95 & 0.11 & 0.92 & 1.00 & 1.00 & \textit{0.84} \\
\multirow{2}{*}{VGG16} & & \multicolumn{1}{l}{CVM} & & 0.00 & 1.00 & 1.00 & 1.00 & 1.00 & \textit{0.00} & & 0.00 & 1.00 & 1.00 & 1.00 & 1.00 & \textit{0.00} & & 0.00 & 0.16 & 1.00 & 1.00 & 1.00 & \textit{0.00} \\
 & & \multicolumn{1}{l}{\cellcolor{mycellcolor}\XXX} & \cellcolor{mycellcolor}  & \cellcolor{mycellcolor}1.00 & \cellcolor{mycellcolor}0.02 & \cellcolor{mycellcolor}0.36 & \cellcolor{mycellcolor}0.99 & \cellcolor{mycellcolor}1.00 & \cellcolor{mycellcolor}\textbf{\textit{0.74}}  & \cellcolor{mycellcolor} & \cellcolor{mycellcolor}1.00 & \cellcolor{mycellcolor}0.00 & \cellcolor{mycellcolor}0.86 & \cellcolor{mycellcolor}1.00 & \cellcolor{mycellcolor}1.00 & \cellcolor{mycellcolor}\textbf{\textit{0.83}} & \cellcolor{mycellcolor} & \cellcolor{mycellcolor}1.00 & \cellcolor{mycellcolor}0.00 & \cellcolor{mycellcolor}1.00 & \cellcolor{mycellcolor}1.00 &  \cellcolor{mycellcolor}1.00 & \cellcolor{mycellcolor}\textit{0.86}\\ \bottomrule
\end{tabular}
}
\end{subtable}

\vspace{1em}

\begin{subtable}[t]{\textwidth}
\centering
\caption{\textbf{Smaller data volumes:} Data stream window sizes $m_w \in \{250, 500, 1000\}$}
\label{tab:drift-detection-experiments-appendix-smaller}
\scalebox{.85}{
\begin{tabular}{clclllllllllllllllllllll}
\toprule
\multirow{5}{*}{\textbf{Use}} & \textbf{} & \multirow{5}{*}{\textbf{Drift}} &  & \multicolumn{20}{c}{} \\
 & & & & \multicolumn{6}{c}{\textit{$m_w = 250$}} & \multicolumn{1}{c}{\textit{}} & \multicolumn{6}{c}{\textit{$m_w = 500$}} & \multicolumn{1}{c}{\textit{}} & \multicolumn{6}{c}{\textit{$m_w = 1000$}} \\ \cline{5-10} \cline{12-17} \cline{19-24} 
 & & &  & & & & & & & & & & & & & & & & & & \\ 
  \textbf{Case} & & \multicolumn{1}{c}{\textbf{Detector}} &  & \multicolumn{5}{c}{\textbf{Drift Percentage $D_{\%}$}} &  & \multicolumn{1}{c}{\textbf{}} &  \multicolumn{5}{c}{\textbf{Drift Percentage $D_{\%}$}} & \multicolumn{1}{c}{\textbf{}} & & \multicolumn{5}{c}{\textbf{Drift Percentage $D_{\%}$}} \\
 & & & & 0\% & 5\% & 10\% & 15\% & 20\% & $H_{DD}$ & & 0\% & 5\% & 10\% & 15\% & 20\% & $H_{DD}$ & & 0\% & 5\% & 10\% & 15\% & 20\% & $H_{DD}$ \\ \cline{5-9} \cline{12-16} \cline{19-23} 
 & & & & & & & & & & & & & & & & & & & & & & \\ \toprule

\multirow{2}{*}{2.2} 
& & \multicolumn{1}{l}{MMD} & & 1.00 & 0.01 & 0.08 & 0.61 & 0.98 & \textit{0.59} & & 0.95 & 0.12 & 0.63 & 1.00 & 1.00 & \textit{0.80} & & 0.56 & 0.72 & 1.00 & 1.00 & 1.00 & \textit{0.70} \\
 & & \multicolumn{1}{l}{KS} & & 0.42 & 0.68 & 0.89 & 1.00 & 1.00 & \textit{0.57} & & 0.01 & 0.99 & 1.00 & 1.00 & 1.00 & \textit{0.02} & & 0.00 & 1.00 & 1.00 & 1.00 & 1.00 & \textit{0.00} \\
20Newsgroup & & \multicolumn{1}{l}{LSDD} & & 1.00 & 0.00 & 0.01 & 0.05 & 0.25 & \textit{0.14} & & 1.00 & 0.01 & 0.06 & 0.35 & 0.83 & \textit{0.48} & & 0.98 & 0.03 & 0.33 & 0.92 & 0.99 & \textit{0.72} \\
\multirow{2}{*}{DistilBERT} & & \multicolumn{1}{l}{CVM} & & 0.31 & 0.74 & 0.90 & 1.00 & 1.00 & \textit{0.46} & & 0.00 & 1.00 & 1.00 & 1.00 & 1.00 & \textit{0.00} & & 0.00 & 1.00 & 1.00 & 1.00 & 1.00 & \textit{0.00} \\
 & & \multicolumn{1}{l}{\cellcolor{mycellcolor}\XXX} & \cellcolor{mycellcolor} & \cellcolor{mycellcolor}1.00 & \cellcolor{mycellcolor}0.15 & \cellcolor{mycellcolor}0.72 & \cellcolor{mycellcolor}0.99 & \cellcolor{mycellcolor}1.00 & \cellcolor{mycellcolor}\textbf{\textit{0.83}} & \cellcolor{mycellcolor} & \cellcolor{mycellcolor}1.00 & \cellcolor{mycellcolor}0.23 & \cellcolor{mycellcolor}0.95 & \cellcolor{mycellcolor}1.00 & \cellcolor{mycellcolor}1.00 & \cellcolor{mycellcolor}\textbf{\textit{0.89}} & \cellcolor{mycellcolor} & \cellcolor{mycellcolor}1.00 & \cellcolor{mycellcolor}0.40 & \cellcolor{mycellcolor}1.00 & \cellcolor{mycellcolor}1.00 & \cellcolor{mycellcolor}1.00 & \cellcolor{mycellcolor}\textbf{\textit{0.92}} \\ \midrule
\multirow{2}{*}{2.3} 
& & \multicolumn{1}{l}{MMD} & & 1.00 & 0.00 & 0.06 & 0.40 & 0.91 & \textit{0.51} & & 0.96 & 0.22 & 0.78 & 1.00 & 1.00 & \textbf{\textit{0.84}} & & 0.37 & 0.95 & 1.00 & 1.00 & 1.00 & \textit{0.54} \\
& & \multicolumn{1}{l}{KS} & & 0.33 & 0.84 & 0.06 & 0.40 & 0.91 & \textit{0.41} & & 0.00 & 1.00 & 1.00 & 1.00 &  1.00 & \textit{0.00} & & 0.00 & 1.00 & 1.00 & 1.00 & 1.00 & \textit{0.00} \\
20Newsgroup & & \multicolumn{1}{l}{LSDD} & & 1.00 & 0.00 & 0.00 & 0.03 & 0.09 & \textit{0.06} & & 1.00 & 0.00 & 0.03 & 0.17 &   0.51 & \textit{0.30} & & 0.97 & 0.09 & 0.36 & 0.86 & 1.00 &  \textit{0.72} \\
\multirow{2}{*}{RoBERTa} & & \multicolumn{1}{l}{CVM} & & 0.26 & 0.87 & 0.98 & 1.00 & 1.00 & \textit{0.41} & & 0.00 & 1.00 & 1.00 & 1.00 & 1.00 & \textit{0.00} & & 0.00 & 1.00 & 1.00 & 1.00 & 1.00 &  \textit{0.00}  \\
& & \multicolumn{1}{l}{\cellcolor{mycellcolor}\XXX} & \cellcolor{mycellcolor} & \cellcolor{mycellcolor}0.98 & \cellcolor{mycellcolor}0.08 & \cellcolor{mycellcolor}0.26 & \cellcolor{mycellcolor}0.57 & \cellcolor{mycellcolor}0.88 & \cellcolor{mycellcolor}\textbf{\textit{0.61}} & \cellcolor{mycellcolor} & \cellcolor{mycellcolor}0.98 & \cellcolor{mycellcolor}0.07 & \cellcolor{mycellcolor}0.34 & \cellcolor{mycellcolor}0.82 & \cellcolor{mycellcolor}0.99 & \cellcolor{mycellcolor}\textit{0.71} & \cellcolor{mycellcolor} & \cellcolor{mycellcolor}0.99 & \cellcolor{mycellcolor}0.07 & \cellcolor{mycellcolor}0.53 & \cellcolor{mycellcolor}0.98 & \cellcolor{mycellcolor}1.00  & \cellcolor{mycellcolor}\textbf{\textit{0.78}} \\ \midrule
 \multirow{2}{*}{6.2} 
 & & \multicolumn{1}{l}{MMD}  &  & 1.00 & 0.01 &  0.12 & 0.63 & 0.99 & \textbf{\textit{0.61}} & & 1.00 & 0.02 & 0.38 & 1.00 & 1.00 & \textit{0.75} & & 0.99 & 0.10 & 0.97 & 1.00 & 1.00 & \textbf{\textit{0.87}} \\
 & & \multicolumn{1}{l}{KS}   & & 1.00 & 0.00 & 0.00 & 0.01 & 0.18 & \textit{0.09} & & 1.00 & 0.00 & 0.00 & 0.18 & 0.95 & \textit{0.44} & & 1.00 & 0.00 & 0.04 & 0.95 & 1.00 & \textit{0.67} \\
Intel Image & & \multicolumn{1}{l}{LSDD} & & 0.96 & 0.05 & 0.05 & 0.07 & 0.09 & \textit{0.12} & & 0.99 & 0.05 & 0.05 & 0.10 & 0.15 & \textit{0.16} & & 0.96 & 0.03 & 0.07 & 0.17 & 0.37 & \textit{0.27} \\
\multirow{2}{*}{VGG16} & & \multicolumn{1}{l}{CVM}  & & 0.00 & 1.00 & 1.00 & 1.00 & 1.00 & \textit{0.00} & & 0.00 & 1.00 & 1.00 & 1.00 & 1.00 & \textit{0.00} & & 0.00 & 1.00 & 1.00 & 1.00 & 1.00 & \textit{0.00} \\
 & & \multicolumn{1}{l}{\cellcolor{mycellcolor}\XXX} & \cellcolor{mycellcolor} & \cellcolor{mycellcolor}0.95 & \cellcolor{mycellcolor}0.06 & \cellcolor{mycellcolor}0.17 & \cellcolor{mycellcolor}0.61 & \cellcolor{mycellcolor}0.94 & \cellcolor{mycellcolor}\textbf{\textit{0.61}} & \cellcolor{mycellcolor} & \cellcolor{mycellcolor}0.94 & \cellcolor{mycellcolor}0.14 & \cellcolor{mycellcolor}0.66 & \cellcolor{mycellcolor}0.99 & \cellcolor{mycellcolor}1.00 & \cellcolor{mycellcolor}\textbf{\textit{0.80}} & \cellcolor{mycellcolor} & \cellcolor{mycellcolor}0.93 & \cellcolor{mycellcolor}0.31 & \cellcolor{mycellcolor}0.98 & \cellcolor{mycellcolor}1.00 & \cellcolor{mycellcolor}1.00 & \cellcolor{mycellcolor}\textbf{\textit{0.87}} \\ \midrule
  \multirow{2}{*}{7.2} 
 & & \multicolumn{1}{l}{\cellcolor{mycellcolor}MMD}  & \cellcolor{mycellcolor} & \cellcolor{mycellcolor}1.00 & \cellcolor{mycellcolor}0.02 & \cellcolor{mycellcolor}0.22 & \cellcolor{mycellcolor}0.95 & \cellcolor{mycellcolor}1.00 & \cellcolor{mycellcolor}\textit{\textbf{0.71}} & \cellcolor{mycellcolor} & \cellcolor{mycellcolor}0.99 & \cellcolor{mycellcolor}0.04 & \cellcolor{mycellcolor}0.76 & \cellcolor{mycellcolor}1.00 & \cellcolor{mycellcolor}1.00 & \cellcolor{mycellcolor}\textbf{\textit{0.82}} & \cellcolor{mycellcolor} & \cellcolor{mycellcolor}1.00 & \cellcolor{mycellcolor}0.07 & \cellcolor{mycellcolor}1.00 & \cellcolor{mycellcolor}1.00 & \cellcolor{mycellcolor}1.00 & \cellcolor{mycellcolor}\textbf{\textit{0.87}} \\
 & & \multicolumn{1}{l}{KS}   & & 1.00 & 0.00 & 0.00 & 0.13 & 0.61 & \textit{0.31} & & 1.00 & 0.00 & 0.06 & 0.71 & 1.00 & \textit{0.61} & & 1.00 & 0.00 & 0.41 & 1.00 & 1.00 & \textit{0.75} \\
STL-10 & & \multicolumn{1}{l}{LSDD} & & 0.97 & 0.03 & 0.05 & 0.20 & 0.65 & \textit{0.37} & & 0.97 & 0.04 & 0.09 & 0.73 & 1.00 & \textit{0.63} & & 0.97 & 0.01 & 0.49 & 1.00 & 1.00 & \textit{0.76} \\
\multirow{2}{*}{VGG16} & & \multicolumn{1}{l}{CVM}  & & 0.00 & 1.00 & 1.00 & 1.00 & 1.00 & \textit{0.00} & & 0.00 & 1.00 & 1.00 & 1.00 & 1.00 & \textit{0.00} & & 0.00 & 1.00 & 1.00 & 1.00 & 1.00 &  \textit{0.00} \\
 & & \multicolumn{1}{l}{\XXX} & & 0.98 & 0.00 & 0.05 & 0.20 & 0.60 & \textit{0.35} & & 1.00 & 0.04 & 0.12 & 0.77 & 1.00 & \textit{0.65} & & 1.00 & 0.07 & 0.70 & 1.00 & 1.00 & \textit{0.82} \\ 
 \bottomrule

\end{tabular}
}
\end{subtable}

\end{table*}

\setcounter{section}{0}
\refstepcounter{section}
\section*{\thesection. Further Drift Detection Experiments}
\label{apx:more-experiment}
This evaluation further assesses the effectiveness and general applicability of \XXX\ in detecting drifted windows of varying severity across deep learning classifiers. We report here the drift detection results using the same experimental settings and detectors introduced in \S\ref{subsec:experimental-use-cases} while varying the deep learning classifiers (indicated with * in Table~\ref{tab_uses_cases} in \S\ref{subsec:experimental-use-cases}).

\smallskip
\noindent
\textbf{Evaluation Metrics} \ We measure drift detection performance using two metrics: \textit{accuracy}, measured across different drift severity levels, and the \(H_{DD}\) score, as introduced in \S\ref{subsec:experimental-use-cases}. Following the same approach as in previous experiments, for each drift percentage \( D_{\%} \) and window size \( m_w \), we randomly sample 100 windows. Each window contains \( m_w \) samples, with \( D_{\%} \) drawn from the drift dataset. The final accuracy is obtained by averaging the accuracy over five independent runs, each computed on 100 windows of size \( m_w \).\footnote{At each run the threshold of \XXX\ is re-estimated and the reference set of the other detectors is re-sampled (if exceeds the maximum size).} The \( H_{DD} \) score is then calculated based on these final accuracy values.

\smallskip
\noindent
\textbf{Results} \ Table \ref{tab:drift-detection-experiments-appendix}  reports the drift prediction accuracy broken down by drift severity, and the $H_{DD}$ score for all experimental drift detectors across use cases 1, 2, and 5–7, using different classifiers than those reported in  \S\ref{subsec:evaluation-accuracy}. Table~\ref{tab:drift-detection-experiments-appendix-larger} uses larger data windows than Table~\ref{tab:drift-detection-experiments-appendix-smaller}  as datasets contain more samples. 

For \textit{larger} data volumes (Table \ref{tab:drift-detection-experiments-appendix-larger}) \XXX\ is still the most effective in distinguishing between windows with and without drift independently of the window size. 
For the AG News dataset (use cases 1.2 and 1.3), \XXX\ remains highly effective overall on the RoBERTa and DistilBERT classifiers (instead of BERT, analyzed in use case 1.1), achieving an $H_{DD} \geq 0.93$, except in use case 1.3 with a window size of 500, where $H_{DD} = 0.87$.
Similarly, for the MNIST dataset (use case 5.2), \XXX\ remains the most effective detector overall when using a VGG16 classifier, instead of the Vision Transformer analyzed in use case 5.1.  
In these use cases with larger datasets and window sizes, \textsc{DriftLens} achieves an average $H_{DD}$ score of $0.91$, outperforming MMD ($0.79$), KS ($0.78$), LSDD ($0.76$), and CVM ($0.54$), confirming that \XXX\ is the most effective with large amount of data across different classifiers.

For \textit{smaller} data volumes (Table \ref{tab:drift-detection-experiments-appendix-smaller}) \XXX\ is still the most effective detector overall. For the 20 Newsgroups dataset (use cases 2.2 and 2.3), \XXX\ remains the most effective overall in detecting drift on the RoBERTa and DistilBERT classifiers (instead of BERT, analyzed in use case 2.1). The only exception is use case 2.3 with a window size of 500, where MMD achieves a higher $H_{DD}$ score. In contrast, KS and CVM are almost unreliable in these use cases, as they tend to consistently predict the presence of drift, aligning with results in use case 2.1.
Similarly, for the Intel Image dataset (use case 6.2), \XXX\ remains the most effective detector overall on the VGG16 classifier, confirming results 
of use case 6.1. In contrast, on the STL-10 dataset (use case 7.2) with the VGG16 model, instead of the Vision Transformer used in use case 7.1, MMD becomes the most effective detector, with \XXX\ being the second most effective.
In these use cases with smaller datasets and window sizes, \textsc{DriftLens} achieves an average $H_{DD}$ score of $0.74$, outperforming MMD ($0.72$), KS ($0.32$), LSDD ($0.39$), and CVM ($0.07$), confirming that \XXX\ is the most effective also with small amount of data across different classifiers. 

\smallskip
\noindent
\textbf{Summary of findings} \ These results confirm findings in~\S\ref{subsec:evaluation-accuracy}. \XXX\ is the most effective overall in distinguishing between drifted and non-drifted windows independently of the data type, deep learning classifier, and data dimensionality.

\refstepcounter{section}
\section*{\thesection. Drift Characterization: a Qualitative Analysis}
\label{apx:drift-characterization}
This section discusses additional qualitative examples of drift patterns over time and explanations produced by \XXX\ to evaluate its ability to characterize drift, for some of the use cases introduced in Table~\ref{tab_uses_cases} in~\S\ref{subsec:experimental-use-cases}.

\smallskip
\noindent
\textbf{Drift Patterns Settings} \  Drift patterns are generated by randomly sampling $100$ windows containing $1{,}000$ samples each. For the \textit{sudden} pattern, drift occurs after 50 windows and remains constant with $D_{\%}\!=\!40\%$ percentage. For the \textit{incremental} pattern, drift occurs after 50 windows with $D_{\%}\!=\!20\%$ percentage and increases by $\Delta D_{\%}\!=\!1\%$ after each window. For the \textit{periodic} pattern, 20 windows without drift and 20 windows containing $D_{\%}\!=\!40\%$ of drift reoccur periodically.

\smallskip
\noindent
\textbf{Drift Explanation Settings} \ Drift explanations are generated using a window of $m_w\!=\!2000$ samples and $20\%$ of drift. K-Means is executed multiple times for $k \in [2, 10]$ on the embedding vectors predicted with the explained label, selecting the best $k$ based on the Silhouette score. 
The prototypes are then identified by selecting the $top - n$ samples closest to each cluster centroid based on the Euclidean distance.

\vspace{-3mm}
\subsection*{Use Case 1.1}
In this use case, a BERT classifier has been trained for topic classification using the AG News \cite{ag-news} dataset. The model is trained on three classes: \textit{World}, \textit{Business}, and \textit{Sport}. To simulate drift, samples from a previously unseen topic (\textit{Science/Technology}) are introduced.

\begin{figure*}
    \centering
    \begin{subfigure}[b]{0.325\textwidth}
        \centering
        \includegraphics[width=\textwidth]{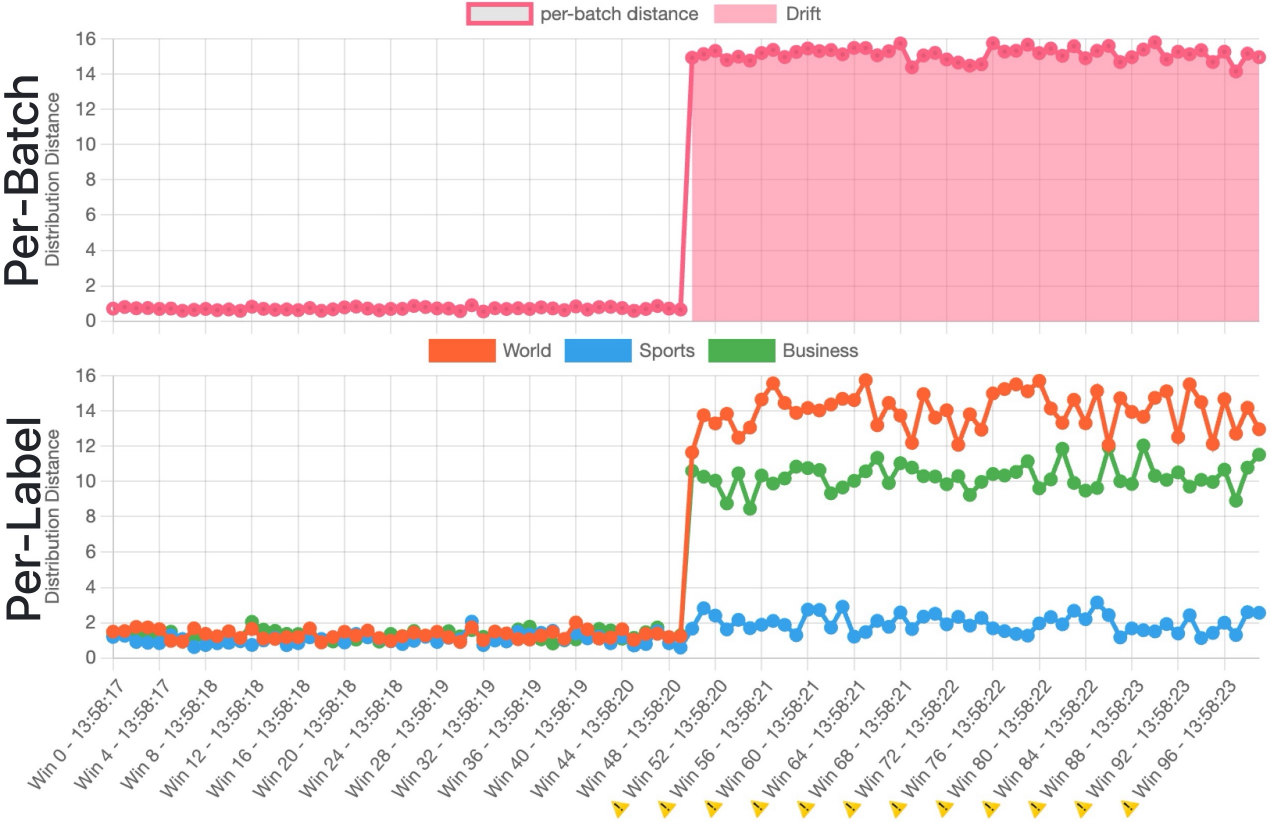}
        \caption{Sudden drift.}
        \label{fig:sub1}
    \end{subfigure}
    \hfill 
    \begin{subfigure}[b]{0.325\textwidth}
        \centering
        \includegraphics[width=\textwidth]{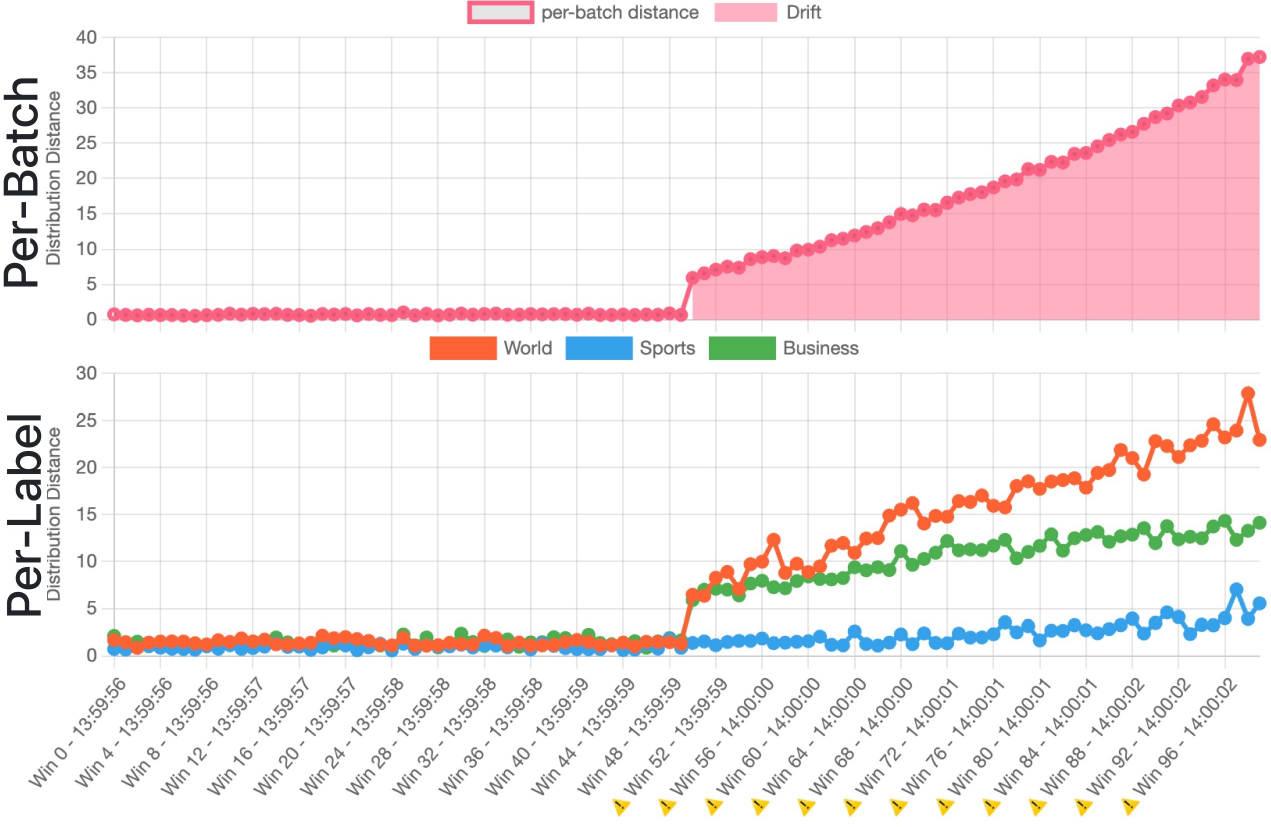}
        \caption{Incremental drift.}
        \label{fig:sub2}
    \end{subfigure}
    \hfill 
    \begin{subfigure}[b]{0.325\textwidth}
        \centering
        \includegraphics[width=\textwidth]{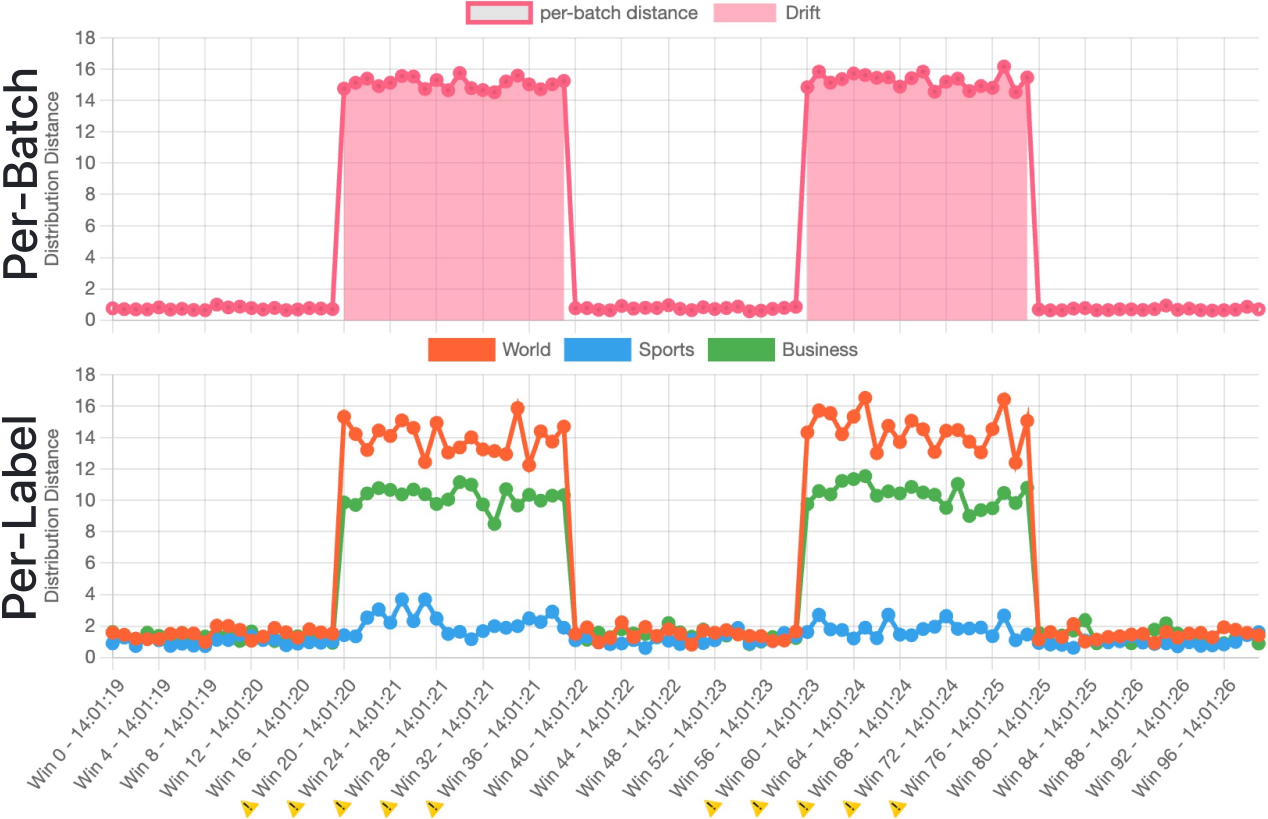}
        \caption{Periodic drift.}
        \label{fig:sub3}
    \end{subfigure}
    \caption{Drift patterns qualitative evaluation for \textit{use case 1.1} (BERT - AG News). Drift simulated with a new class (\textit{Science/Tech}).}
    \label{fig:drift-patterns-usecase1}
\end{figure*}

\smallskip
\noindent
\textbf{Drift Pattern} \
Figure \ref{fig:drift-patterns-usecase1} shows the (a) \textit{sudden}, (b) \textit{incremental}, and (c) \textit{periodic} drift patterns generated using the settings described above. 
The \textit{per-batch} and \textit{per-label} drift curves produced align with the generated patterns. 
In addition, the \textit{per-label} curves show that the labels \textit{World} (red) and \textit{Business} (green) are the most affected by drift, likely because the newly introduced \textit{Science/Technology} samples are predominantly misclassified under these categories. In contrast, drift has a minimal impact on the \textit{Sport} label.

\smallskip
\noindent
\textbf{Drift Explanation} \ Table \ref{tab:explanations-agnews-world} reports the \textit{per-label} drift explanation for the \textit{World} class. For each cluster, the four samples closest to the centroids are extracted as prototypes.
When drift is absent (Table \ref{tab:explanations-agnews-world-historical}), prototypes from the \textit{historical} data indicate that the \textit{World} class is primarily characterized by two distinct groups of samples: (1) Articles covering global economic and political events, such as oil exports, trade agreements, and multinational business operations (Cluster 1), and (2) Articles related to international diplomacy and conflicts (Cluster 2).
In contrast, the drift explanations generated from a \textit{drifted window} (Table~\ref{tab:explanations-agnews-world-drift}) reveal significant changes in the composition of the \textit{World} class due to the introduction of samples from a new class label (\textit{Science/Technology}). Although one group (Cluster 1) still consists of articles aligned with the original \textit{World} category, a second group of articles (Cluster~2) exhibits a substantial difference. This drifted cluster is now dominated by science and technology-related articles misclassified as \textit{World}, particularly those discussing advancements in nanotechnology, cybersecurity, and scientific research. Users can easily understand that the presence of these samples is likely the cause of the drift for this label. This may be driven by an overlap in terminology and contextual similarities between world events and scientific developments, leading to misclassification. Notice that true labels and drift labels are not available in the data stream. However, prototypes can facilitate human understanding of drift by reducing the research space of samples to analyze. In this case, the model can be adapted to new distributions by collecting and annotating additional data labeled as \textit{Science/Tech}, which is then used to retrain the model.

\vspace{-2mm}
\subsection*{Use Case 2.1}
In this use case, a BERT classifier has been trained for topic classification using a subset of the macro-topic from the 20 Newsgroups~\cite{misctwentynewsgroups113} dataset. The model is trained on the following classes: \textit{Technology}, \textit{Sale-Ads}, \textit{Politics}, \textit{Religion}, and \textit{Science}.
Drift is simulated by introducing samples from a previously unseen macro-class, \textit{Recreation}, which includes subtopics such as autos, sports, baseball, and hockey.

\begin{figure*}
    \centering
    \begin{subfigure}[b]{0.325\textwidth}
        \centering
        \includegraphics[width=\textwidth]{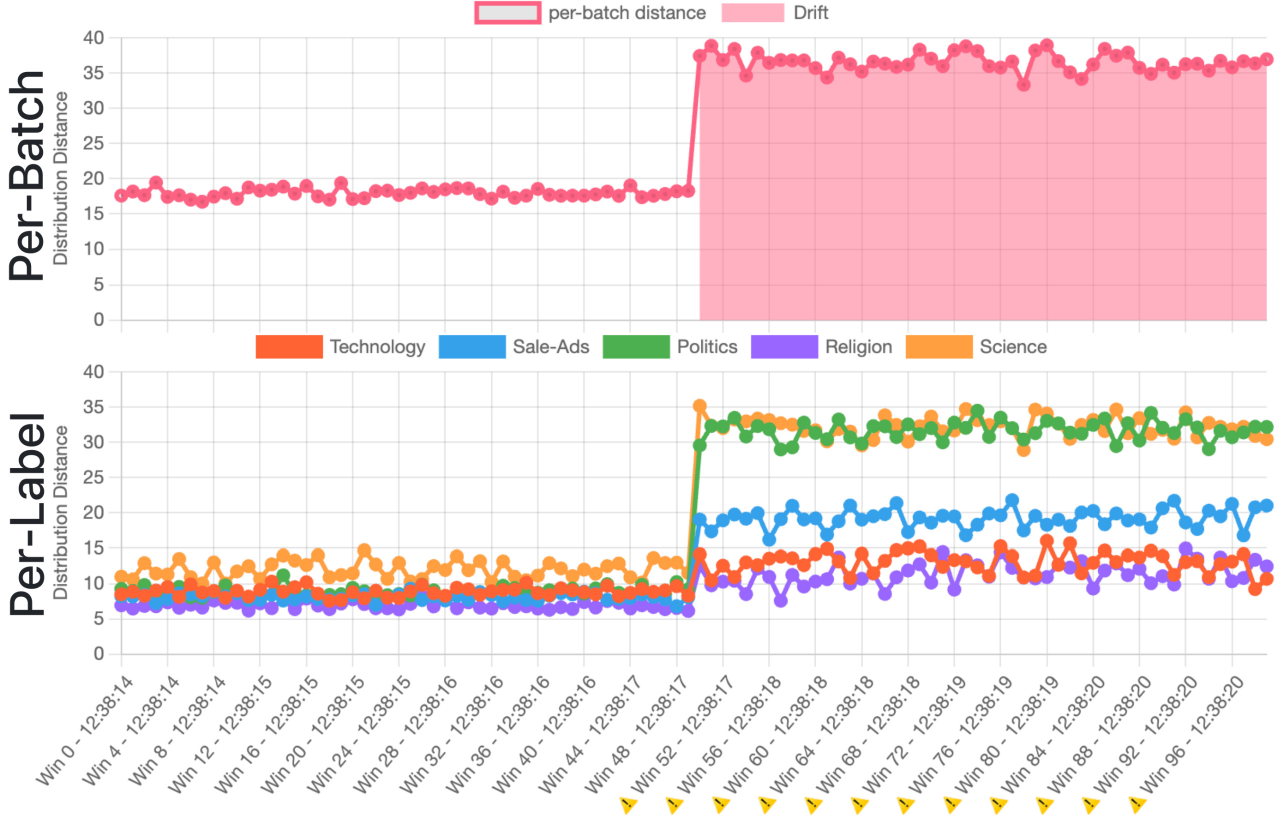}
        \caption{Sudden drift.}
        \label{fig:sub1}
    \end{subfigure}
    \hfill 
    \begin{subfigure}[b]{0.325\textwidth}
        \centering
        \includegraphics[width=\textwidth]{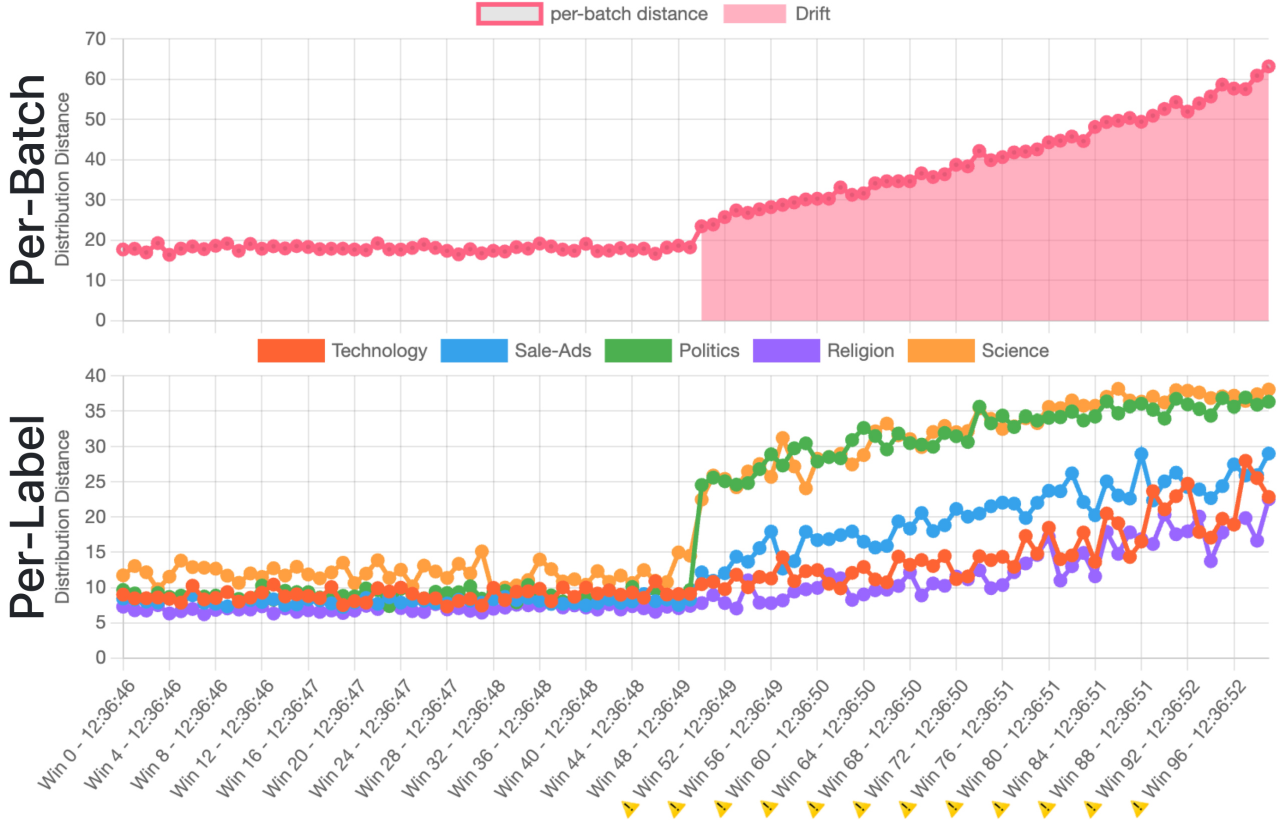}
        \caption{Incremental drift.}
        \label{fig:sub2}
    \end{subfigure}
    \hfill 
    \begin{subfigure}[b]{0.325\textwidth}
        \centering
        \includegraphics[width=\textwidth]{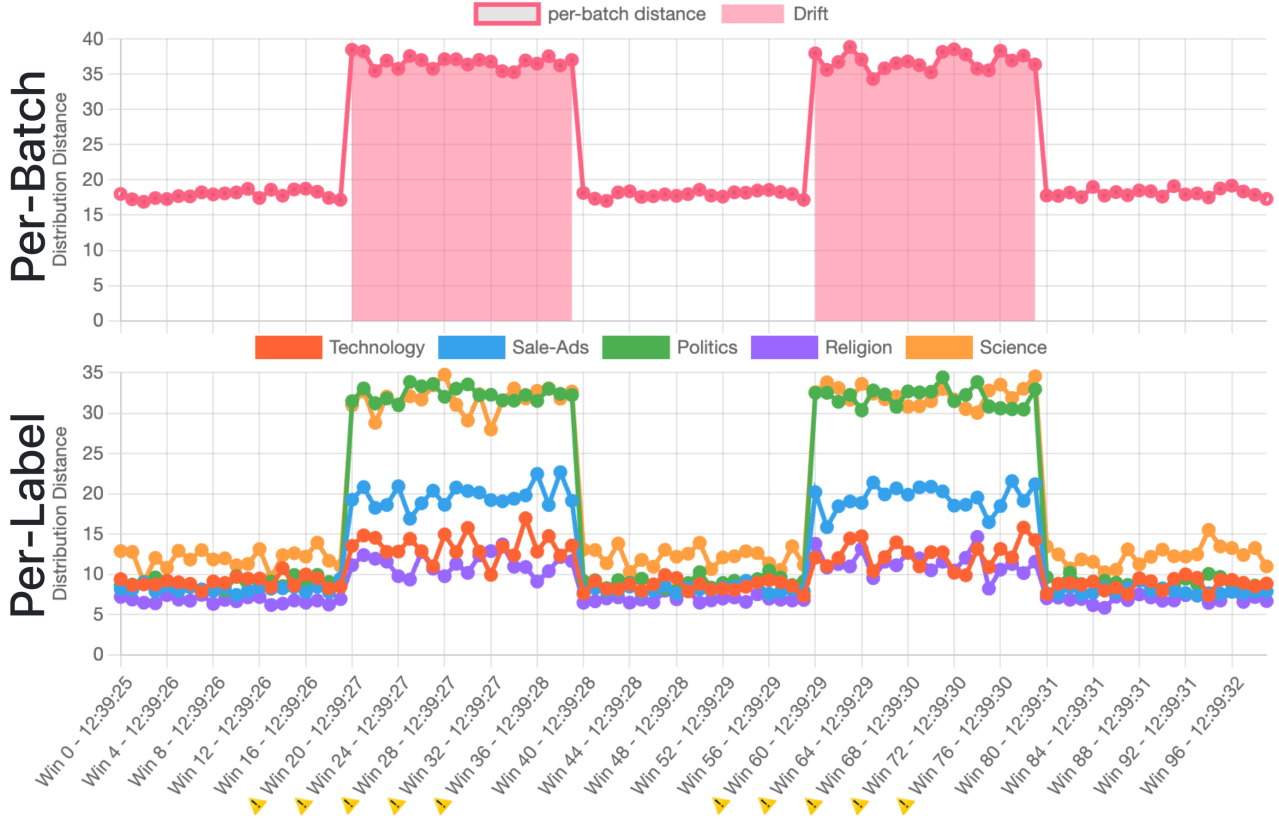}
        \caption{Periodic drift.}
        \label{fig:sub3}
    \end{subfigure}
    \caption{Drift patterns qualitative evaluation for \textit{use case 2.1} (BERT - 20 News). Drift simulated with a new class (\textit{Recreation}).}
    \label{fig:drift-patterns-usecase2}
\end{figure*}

\smallskip
\noindent
\textbf{Drift Pattern} \ Figure \ref{fig:drift-patterns-usecase2} shows the (a) \textit{sudden}, (b) \textit{incremental}, and (c) \textit{periodic} drift patterns, generated using the previously described settings. These curves confirm that the drift trend over time modeled by \XXX\ aligns with the generated pattern while highlighting the labels most affected by drift. In this case, the labels most affected by drift are \textit{Science} (yellow) and \textit{Politics} (green), while \textit{Sale-Ads} (blue) experiences a moderate impact. In contrast, \textit{Technology} (red) and \textit{Religion} (purple) are the least affected.

\smallskip
\noindent
\textbf{Drift Explanation} \ Table \ref{tab:explanations-20news-science} presents the \textit{per-label} drift explanation for the \textit{Science} class. For each cluster, the two closest samples to the centroid are selected as prototypes.
Prototypes extracted from \textit{historical} data (Table \ref{tab:explanations-20news-science-historical}) suggest that, in the absence of drift, the \textit{Science} class primarily consists of three groups of samples: (1) Technical discussions on hardware and electronic components, such as microcontrollers, memory modules, and programmable logic devices (Cluster~1); (2) Conversations about cryptography, security, and governmental policies (Cluster 2); and (3) Broader discussions on technology-related policy and activism, including digital rights and online political engagement (Cluster 3).
In contrast, prototypes generated from a \textit{drifted window} (Table \ref{tab:explanations-20news-science-drift}) reveal that while the first cluster remains largely unchanged, the second cluster experiences drift. Specifically, new content from the \textit{Recreation} category—such as discussions on sports broadcasting and hockey management—introduces topics that were not originally associated with the \textit{Science} class. This semantic change leads to misclassification, causing the observed drift.

\vspace{-1mm}
\subsection*{Use Case 4}
In this use case, a BERT classifier has been trained on the Bias in Bios dataset \cite{10.1145/3287560.3287572} to predict occupations based on short biographies. Target occupations include \textit{Professor}, \textit{Physician}, \textit{Attorney}, \textit{Photographer}, \textit{Journalist}, and \textit{Nurse}.
To simulate a biased classifier, the training data includes only female individuals for the \textit{Nurse} occupation and only male individuals for the other occupations. Drift is introduced by presenting the model with examples of individuals of the opposite gender—i.e., female \textit{Professors}, \textit{Physicians}, \textit{Attorneys}, \textit{Photographers}, and \textit{Journalists}, as well as male \textit{Nurses}. When exposed to these samples, the classifier's performance decreases significantly, highlighting its reliance on biased gendered patterns from data.

\begin{figure*}
    \centering
    \begin{subfigure}[b]{0.325\textwidth}
        \centering
        \includegraphics[width=\textwidth]{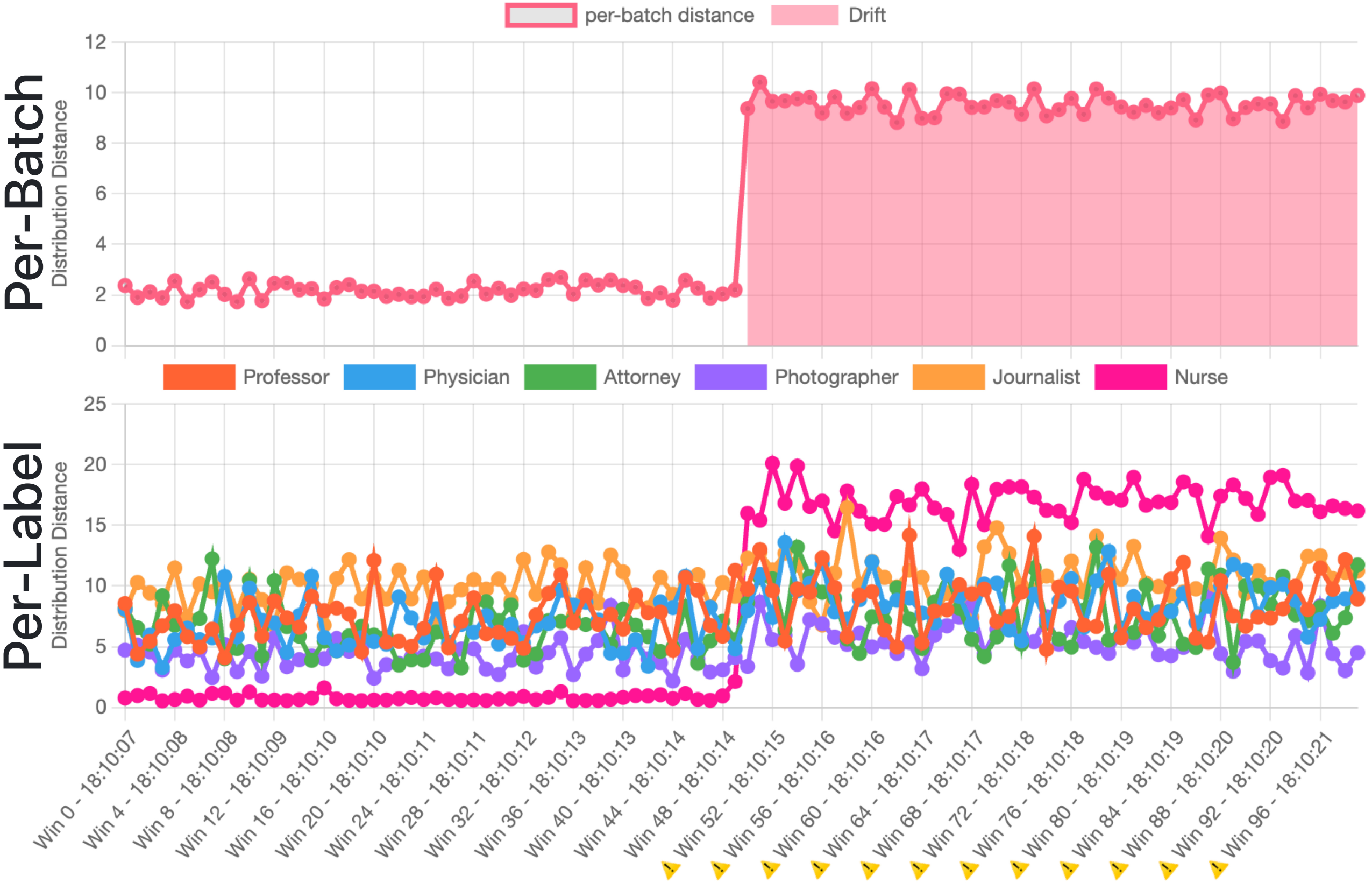}
        \caption{Sudden drift.}
        \label{fig:sub1}
    \end{subfigure}
    \hfill 
    \begin{subfigure}[b]{0.325\textwidth}
        \centering
        \includegraphics[width=\textwidth]{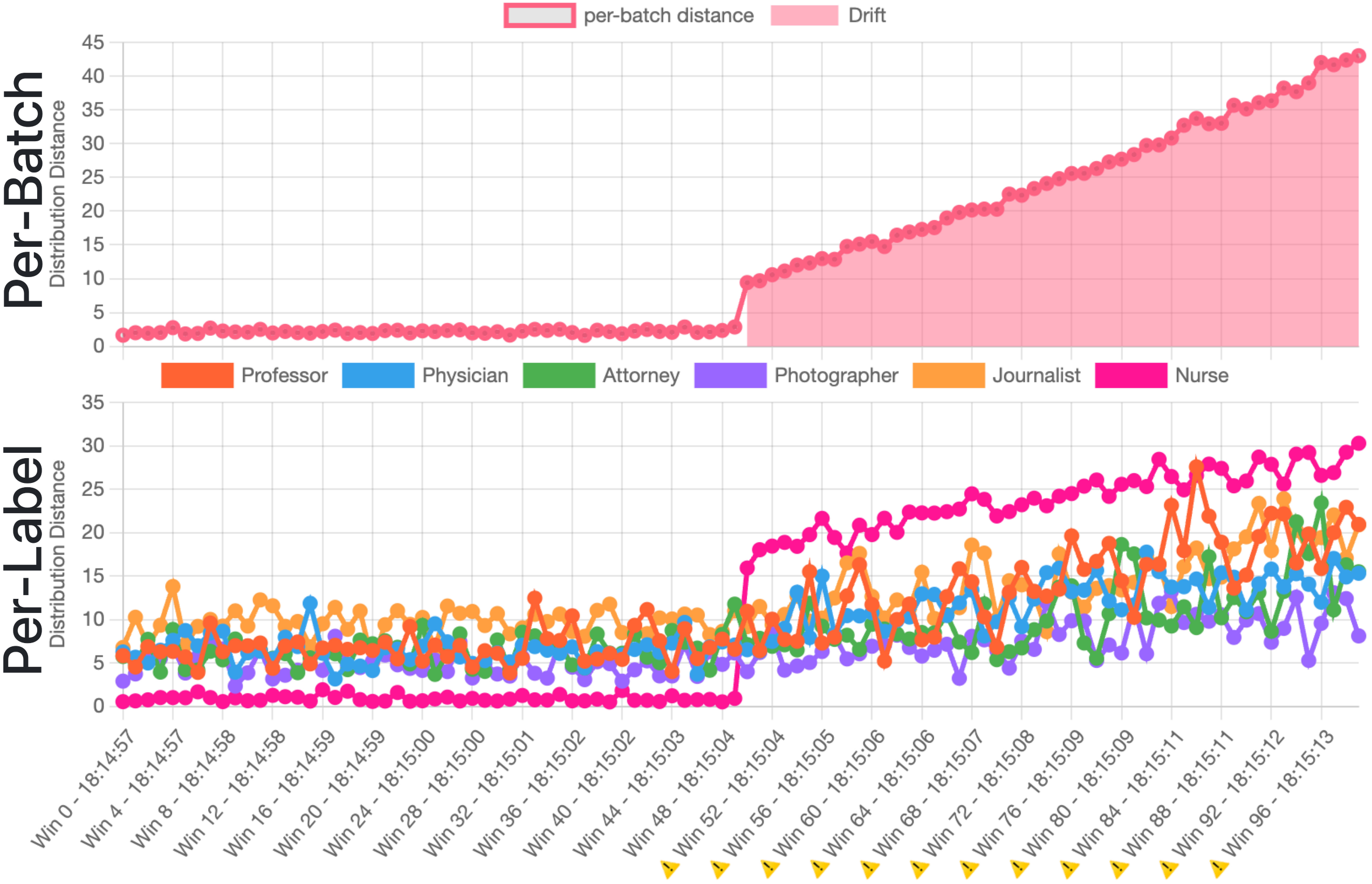}
        \caption{Incremental drift.}
        \label{fig:sub2}
    \end{subfigure}
    \hfill 
    \begin{subfigure}[b]{0.325\textwidth}
        \centering
        \includegraphics[width=\textwidth]{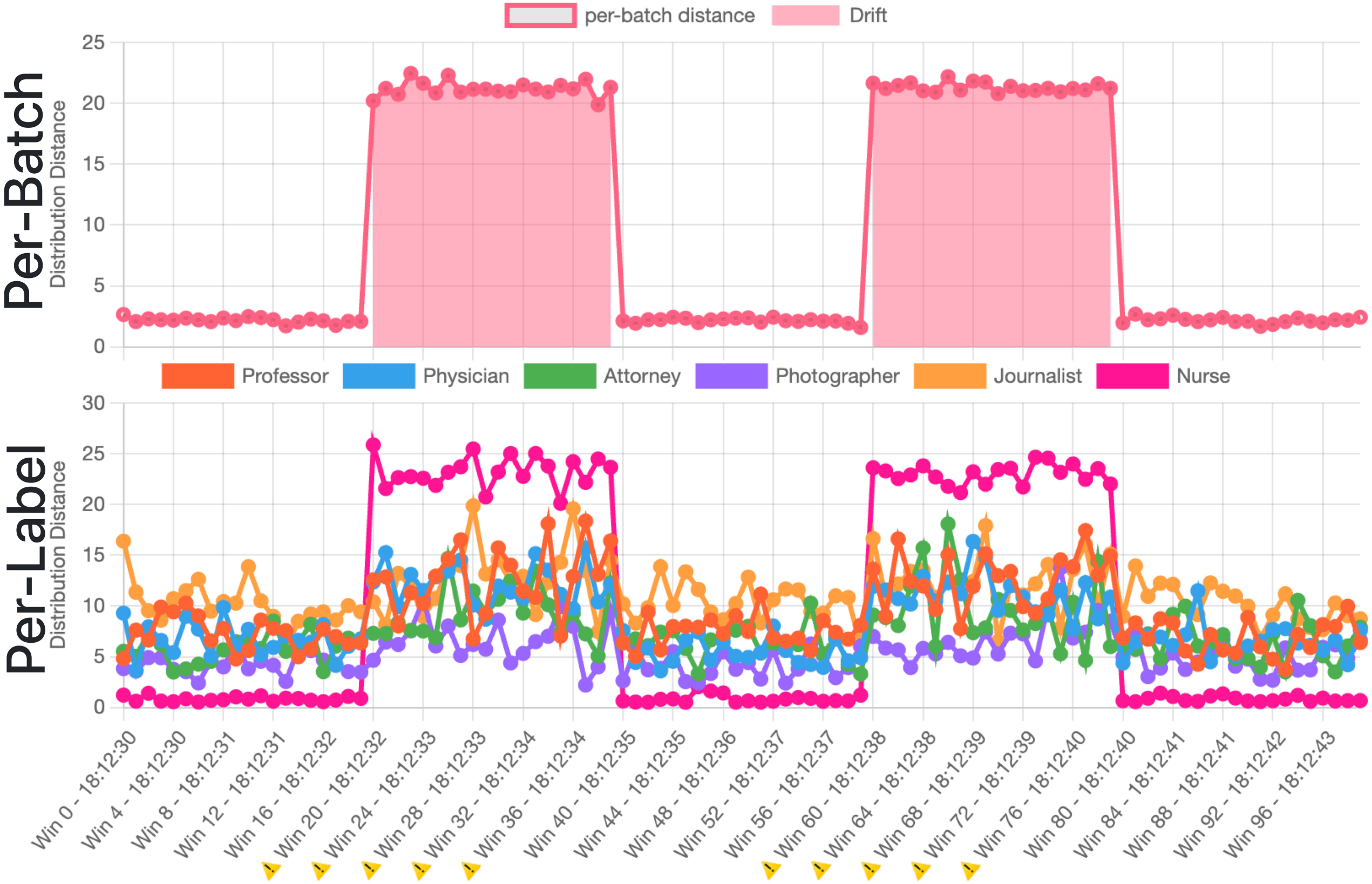}
        \caption{Periodic drift.}
        \label{fig:sub3}
    \end{subfigure}
    \caption{Drift patterns qualitative evaluation for \textit{use case 4} (BERT - BiasInBios). Drift simulated with opposite gender.}
    \label{fig:drift-patterns-usecase-biasinbios}
\end{figure*}

\smallskip
\noindent
\textbf{Drift Pattern} \ Figure \ref{fig:drift-patterns-usecase-biasinbios} shows the (a) \textit{sudden}, (b) \textit{incremental}, and (c) \textit{periodic} drift patterns. 
When drift occurs, the \textit{Nurse} label (pink) is the most affected, likely due to the misclassification of female-gendered bios from other occupations as nurses. In contrast, the impact on other occupations remains minimal. Additionally, the figure shows that, in the absence of drift, the distribution distance for the nurse occupation is typically smaller than for other occupations. This suggests the classifier has likely overfitted to this label, relying heavily on gender patterns in the bios to make predictions.

\smallskip
\noindent
\textbf{Drift Explanation} \ 
Table \ref{tab:explanations-biasinbios-nurse} reports the \textit{per-label} drift explanation for the \textit{Nurse} class. For each cluster, the two closest samples to the centroids are extracted as prototypes.
Prototypes generated from the \textit{historical} data (Table \ref{tab:explanations-biasinbios-nurse-historical}) reveal that the \textit{Nurse} class is characterized by two distinct groups of bios: (1) Bios describing nurse practitioners with extensive clinical experience and affiliations with multiple hospitals (Cluster~1); and (2) Bios that highlight alternative aspects of nursing, such as therapeutic practices and roles in medical information services (Cluster 2).
In contrast, prototypes extracted in a \textit{drifted window} (Table \ref{tab:explanations-biasinbios-nurse-drift}) indicate significant changes in the concepts associated with samples predicted as \textit{Nurse}. While one cluster (Cluster 5) still contains bios correctly classified as \textit{Nurse}, the remaining clusters predominantly consist of samples from other occupations—\textit{Professor}, \textit{Journalist}, \textit{Photographer}, and \textit{Attorney}. However, these misclassified bios share a common pattern: they belong to female individuals. This suggests that the classifier is biased, tending to misclassify samples as \textit{Nurse} when they belong to female individuals. Notably, these prototypes provide insight into the drift phenomenon: the classifier's exposure to bios of individuals with different gender distributions than those seen during training likely causes the drift. As in previous cases, true labels are unavailable due to the unsupervised setting. However, humans can still readily identify the underlying cause of the drift with these explanations, which, in this case, highlight a bias.

\vspace{-1mm}
\subsection*{Use Case 7.1}
In this use case, a Vision Transformer model is trained on the STL-10 dataset \cite{stl} to classify images into one of the following labels: \textit{Airplane}, \textit{Bird}, \textit{Car}, \textit{Cat}, \textit{Deer}, \textit{Dog}, \textit{Horse}, \textit{Monkey}, and \textit{Ship}. Drift is introduced by adding images from a new, previously unseen class label (\textit{Truck}).

\begin{figure*}
    \centering
    \begin{subfigure}[b]{0.325\textwidth}
        \centering
        \includegraphics[width=\textwidth]{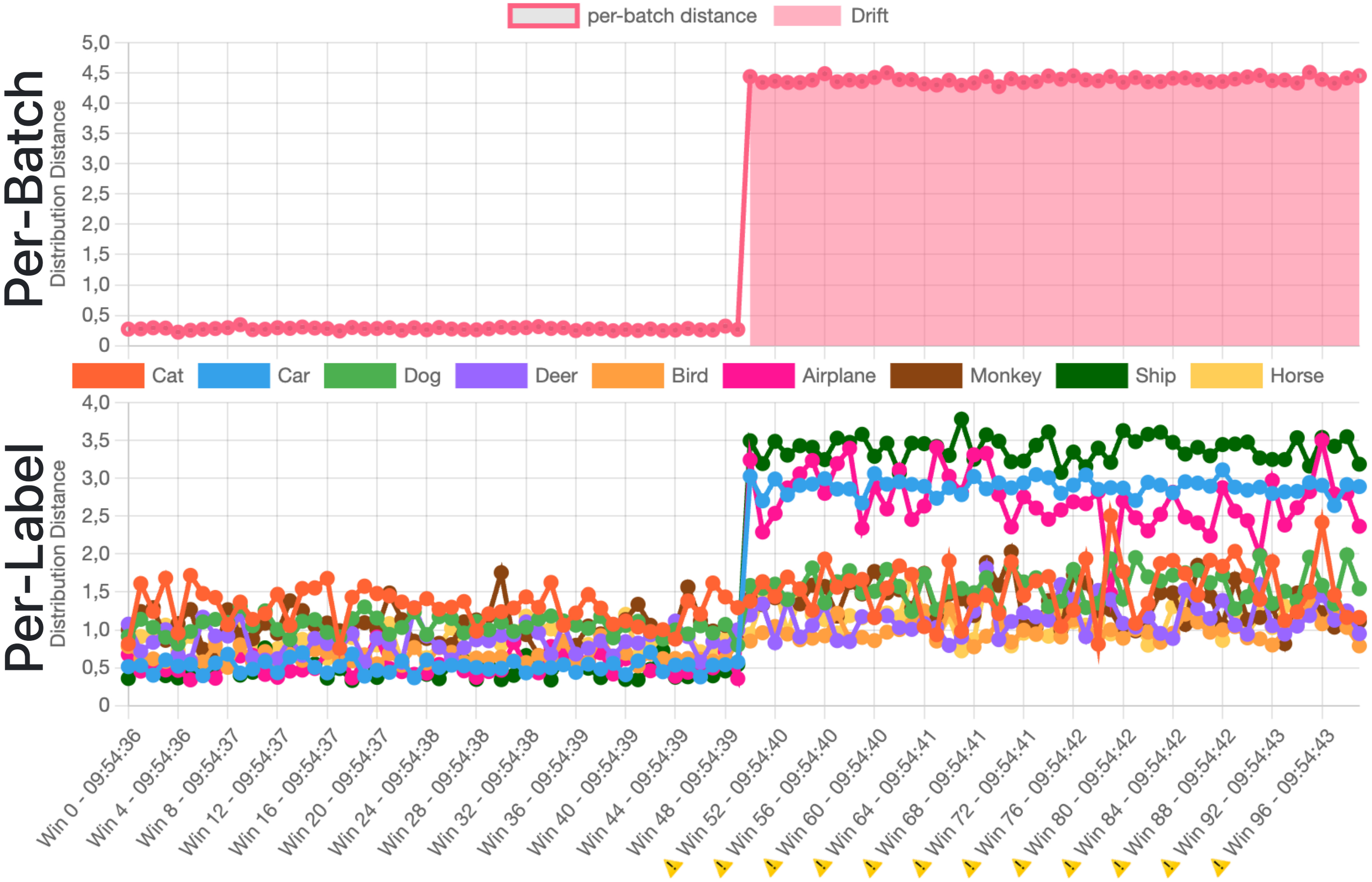}
        \caption{Sudden drift.}
        \label{fig:sub1}
    \end{subfigure}
    \hfill 
    \begin{subfigure}[b]{0.325\textwidth}
        \centering
        \includegraphics[width=\textwidth]{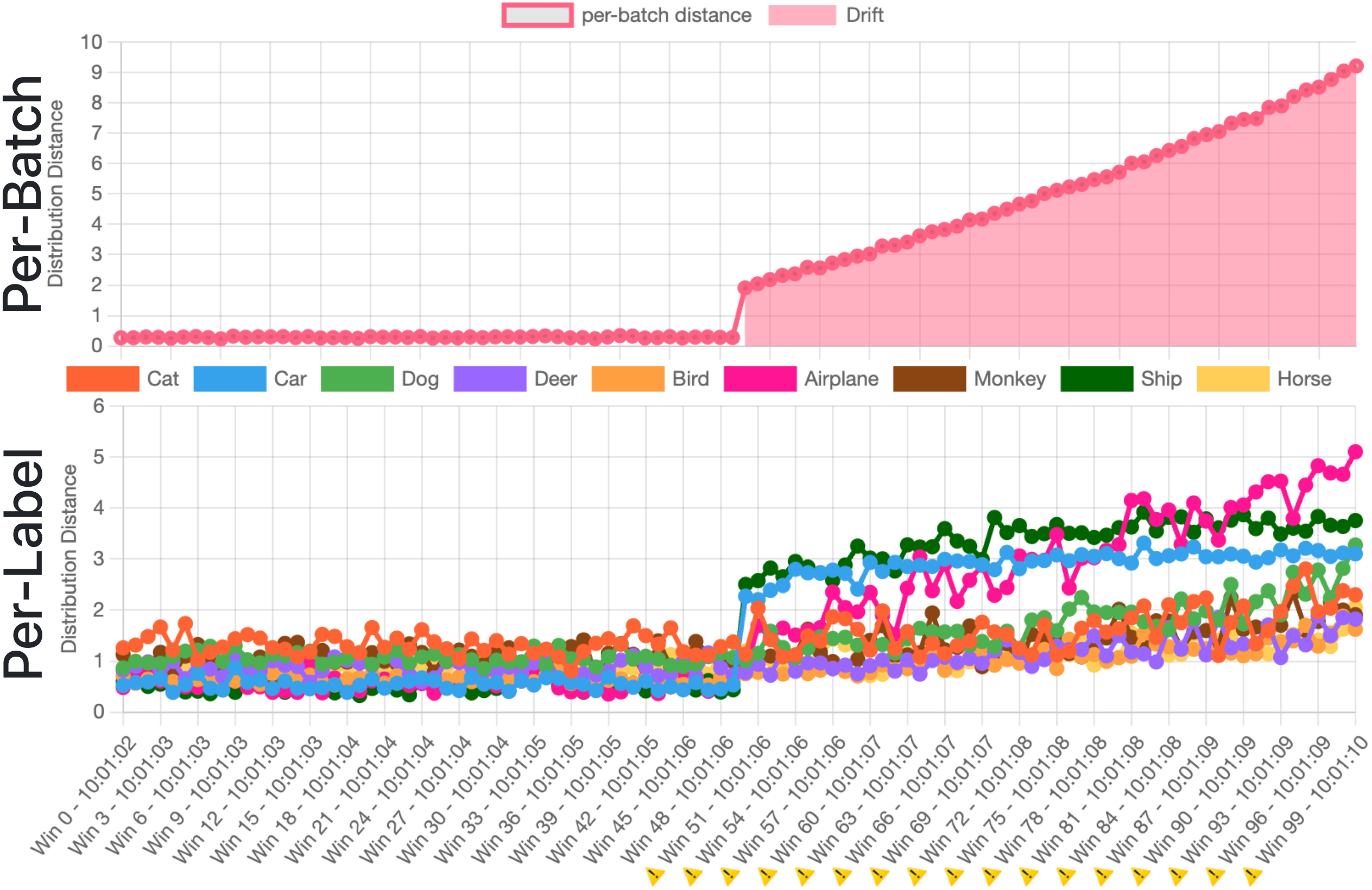}
        \caption{Incremental drift.}
        \label{fig:sub2}
    \end{subfigure}
    \hfill 
    \begin{subfigure}[b]{0.325\textwidth}
        \centering
        \includegraphics[width=\textwidth]{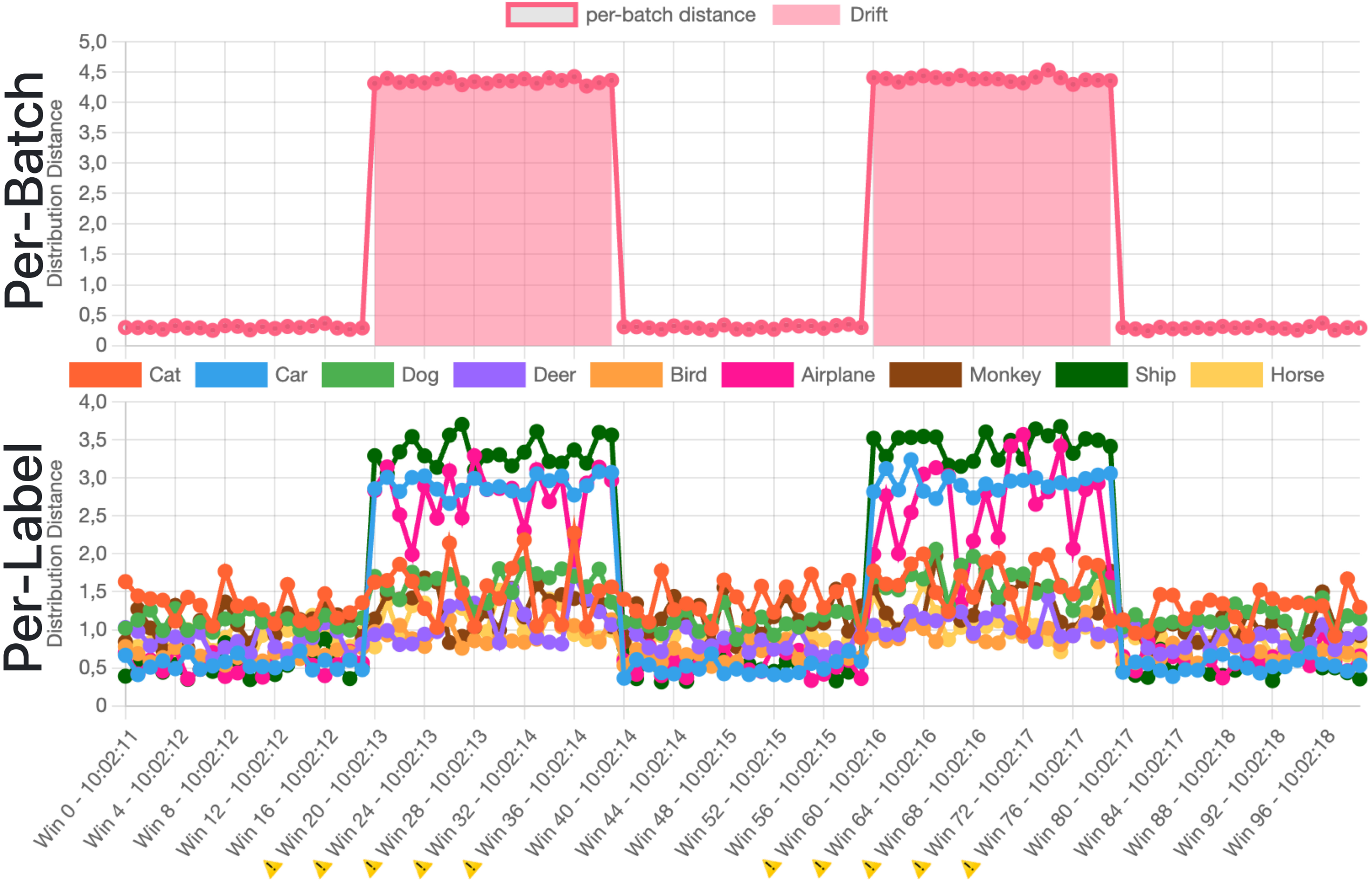}
        \caption{Periodic drift.}
        \label{fig:sub3}
    \end{subfigure}
    \caption{Drift patterns qualitative evaluation for \textit{use case 7.1} (ViT - STL-10). Drift simulated with a new class \textit{Truck}.}
    \label{fig:drift-patterns-usecase-stl}
\end{figure*}

\smallskip
\noindent
\textbf{Drift Pattern} \ Figure \ref{fig:drift-patterns-usecase-stl} shows the (a) \textit{sudden}, (b) \textit{incremental}, and (c) \textit{periodic} drift patterns. The drift curves align with these patterns, confirming their coherence. As expected, the most impacted labels are \textit{Ship} (dark green), \textit{Car} (blue), and \textit{Airplane} (pink), as the classifier tends to misclassify \textit{Truck} images under these categories due to their semantic similarity. In contrast, the remaining labels are almost unaffected by drift.

\begin{figure*}[!ht]
  \centering \includegraphics[width=0.834\textwidth]{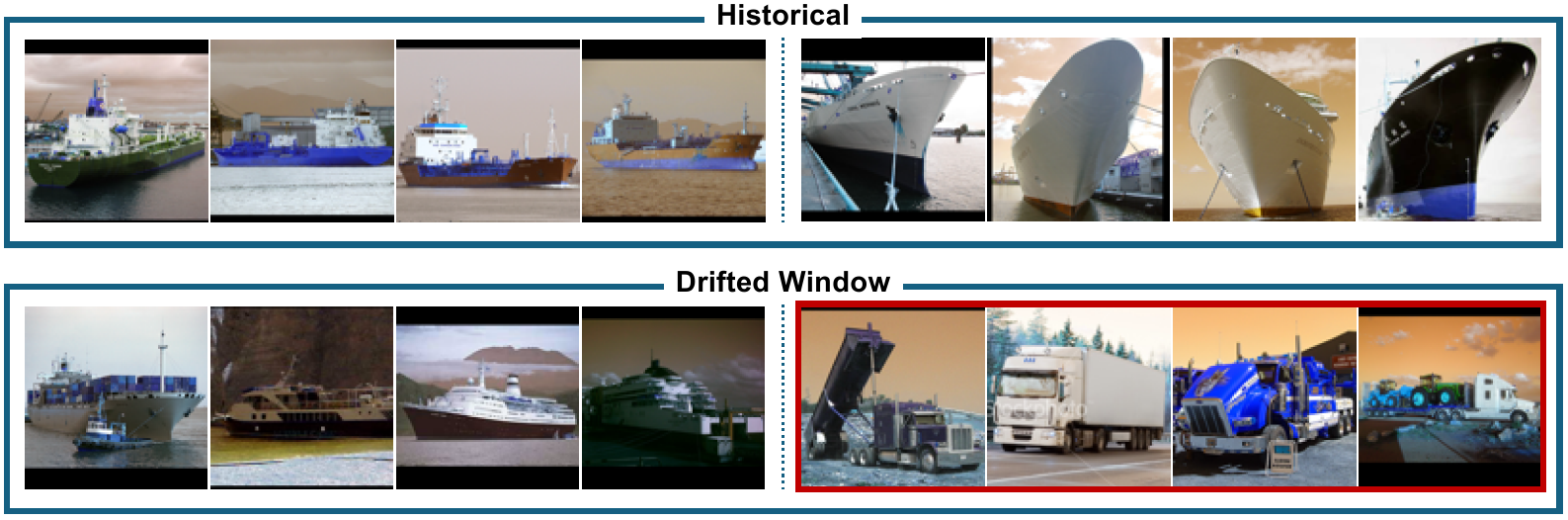}
  \caption{Drift explanation for class \textit{Ship} in use case 7.1.}
  \label{fig:explanation-stl-ship}
\end{figure*}

\begin{figure*}[!ht]
  \centering \includegraphics[width=0.834\textwidth]{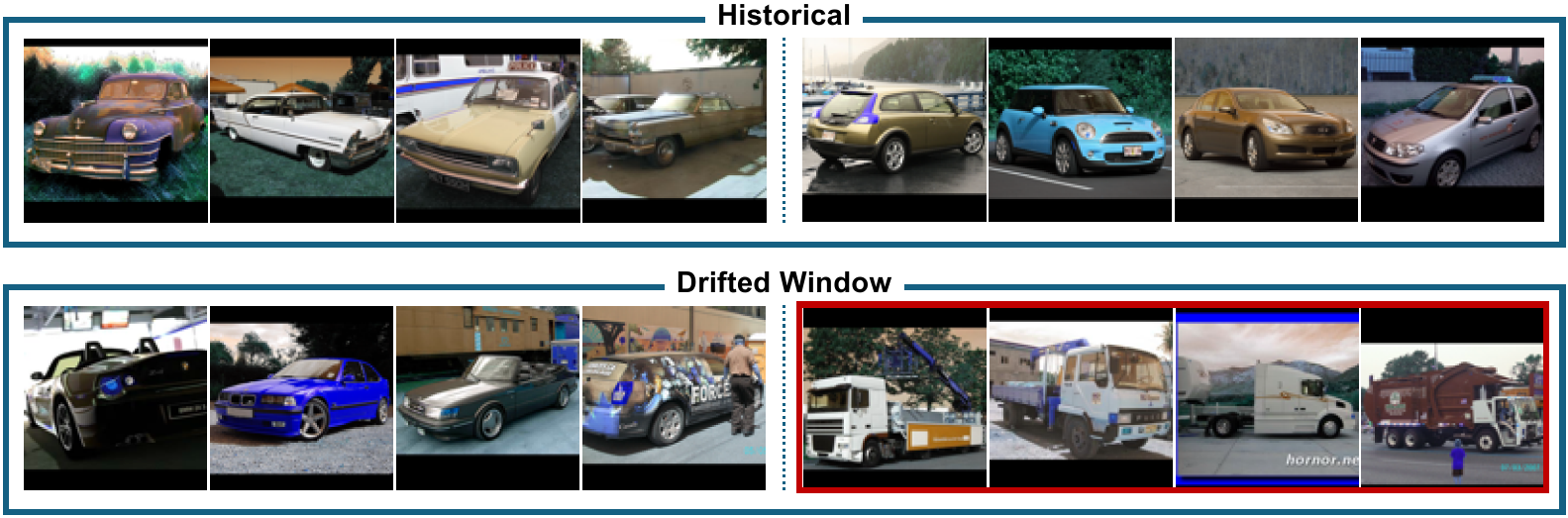}
  \caption{Drift explanation for class \textit{Car} in use case 7.1.}
  \label{fig:explanation-stl-car}
\end{figure*}

\begin{figure*}[!ht]
  \centering \includegraphics[width=0.834\textwidth]{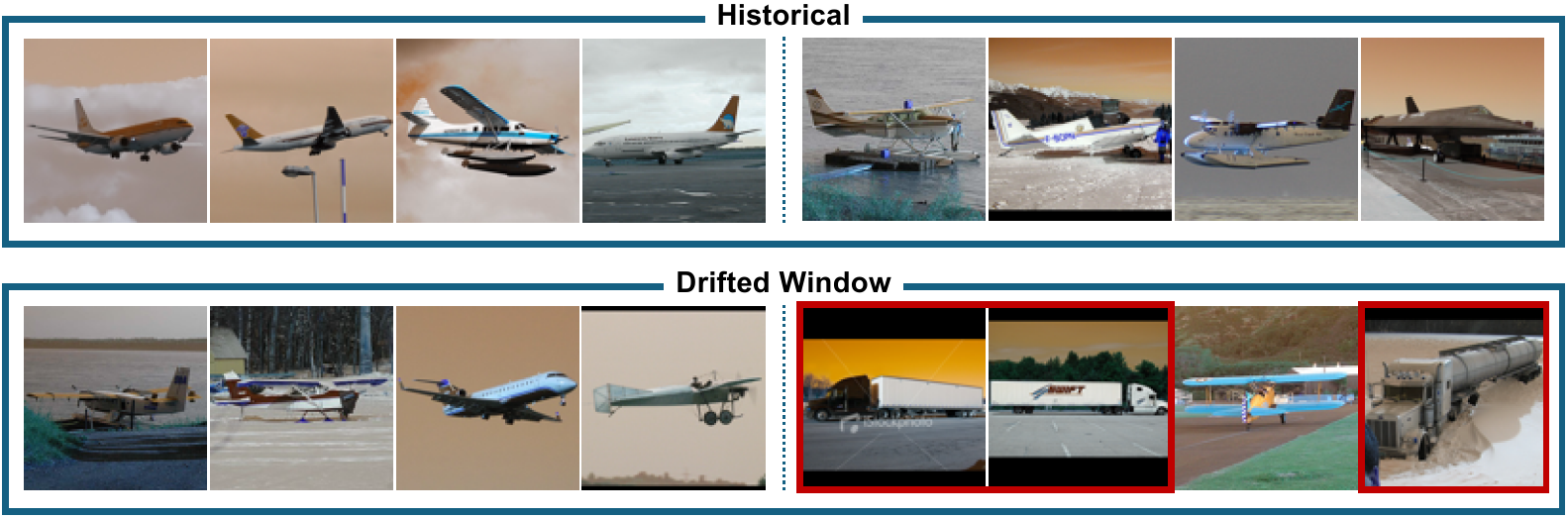}
  \caption{Drift explanation for class \textit{Airplane} in use case 7.1.}
  \label{fig:explanation-stl-airplane}
\end{figure*}

\smallskip
\noindent
\textbf{Drift Explanation} \ Figures \ref{fig:explanation-stl-ship}, \ref{fig:explanation-stl-car} and \ref{fig:explanation-stl-airplane} show the \textit{per-label} drift explanations, generated on the most affected three classes: \textit{Ship}, \textit{Car}, and \textit{Airplane}. 
For each cluster, the four samples closest to the centroids are extracted as prototypes, with drifted samples highlighted in red boxes.
In Figure~\ref{fig:explanation-stl-ship}, prototypes generated from \textit{historical} data for the \textit{Ship} class are represented by two clusters: (1) Large cargo ships (top-left) and (2)~Passenger ships (top-right). In contrast, prototypes generated from a \textit{drifted window} are characterized by a cluster of mixed ships (bottom-left) and a new emerging cluster of particular truck images (bottom-right), which is the cause for drift. This suggests that the model has started misclassifying certain truck images as ships, likely due to their similar background, leading to a distribution drift and potential errors in classification. The behaviour is similar also for the class \textit{Car} (Figure \ref{fig:explanation-stl-car}), where prototypes extracted from the \textit{drifted window} highlight a new concept associated with different types of trucks (bottom-right), but this case, on normal streets background. Finally, the explanation of the class \textit{Airplane} (Figure \ref{fig:explanation-stl-airplane}) identifies a mixed cluster composed of three prototype images picturing particular long tracks and one prototype of a plane (bottom-right).

\smallskip
\smallskip
\smallskip
\noindent
\textbf{Summary of findings} \ These examples further qualitatively demonstrate the effectiveness of \XXX\ in characterizing drift by: (1) modeling drift patterns over time while highlighting drift impact on individual labels; and (2) Providing explanations that help identify and understand the causes of drift. These examples show how drift characterization may facilitate adaptation by providing users with valuable insight to understand and interpret emerging concepts that cause drift and identifying labels that require further analysis. These findings reinforce the results discussed in~\S\ref{subsec:evaluation-characterization} and \S\ref{subsec:evaluation-explanations}.


\refstepcounter{section}
\section*{\thesection. Parameters Sensitivity Evaluation}
\label{apx:evaluation-sensitivity}
This evaluation determines the robustness and sensitivity of \XXX\ to its parameters and the size of reference data. 

\smallskip
\noindent
\textbf{Evaluation metrics} \ We evaluate \XXX' drift detection performance, in terms of accuracy and $H_{DD}$, by varying the values of the following parameters:
(i) the number of randomly sampled windows to estimate the threshold $n_{th}$,
(ii) the threshold sensitivity parameter $T_\alpha$, and (iii) the number of principal components used to reduce the dimensionality of the embedding $d'$.
We additionally measure performance variation by decreasing the amount of reference data $m_b$ to assess how performance would change, even if this is not a real parameter of the framework---the reference set is always used entirely.

\smallskip
\noindent
\textbf{Results} \ Table \ref{tab:experiments-parameters-sensitivity} reports the accuracy and $H_{DD}$ by varying one parameter at a time while keeping the others fixed. 
The default fixed values are underlined and set to the following values: 
$n_{th}\!=\!10{,}000$, 
$T_\alpha\!=\!0.01$, 
$d'\!=\!150$, and $w_b\%\!=\!100\%$ (entire training set used for the baseline). The window size $m_w$ is fixed to $1{,}000$.
The experiments are repeated 5 times and averaged for use cases 1.1 and 7.1 in Table \ref{tab_uses_cases}. 

The results indicate that varying the parameter values has a minimal impact on performance, with a maximum reduction of $0.07$.
 The only exception is the threshold sensitivity \( T_\alpha \) value. As expected, an increase in this parameter causes \textsc{DriftLens} to lower its estimated threshold values, leading to more false positives—normal windows incorrectly classified as affected by drift. This leads to reduced accuracy, especially when there is no actual drift (\( D_{\%} = 0\% \)). However, we can conclude that the performance of \XXX\ is not significantly affected by the choice of parameters, except for the threshold sensitivity.

Interestingly, for use case 1.1, \XXX\ is not affected by the reduction in the number of samples in the reference window (i.e., used for the baseline modeling). This is probably because the training dataset is large (\(\approx 60{,}000\) samples). Therefore, even a \(20\%\) of these samples would be enough to properly model the baseline (reference) distributions. In contrast, for use case 7.1, a reduction in the available samples for the baseline estimation leads to a reduction in the drift detection performance proportional to the decrease in the number of samples. This is likely because, for this use case, a small number of training samples are available to estimate the reference distributions (\(\approx 6{,}000\)). This limited sample size may result in an inadequate representation of the reference data distribution, leading to poorer drift detection performance. 

\smallskip
\noindent
\textbf{Summary of findings} \ \XXX\ is robust to its parameters. The most sensitive parameter is the threshold sensitivity  $T_\alpha$, as increasing its value can lead to a lower threshold and a higher rate of false positive drift predictions. Finally, when the historical dataset is sufficiently large, reducing the size of the reference data does not significantly affect performance.

\begin{table}[]
\scriptsize
\centering
\caption{\textbf{Parameters sensitivity.}}
\label{tab:experiments-parameters-sensitivity}
\scalebox{.85}{
\begin{tabular}{clllllll}
\toprule
\multirow{2}{*}{\textbf{Use}} & \multirow{3}{*}{\textbf{Parameter}} & \multicolumn{5}{c}{\textbf{Drift Percentage $D_{\%}$}} \\
\cmidrule{3-7}
 \textbf{Case} &  & 0\% & 5\% & 10\% & 15\% & 20\% & $H_{DD}$ \\
 \bottomrule  \multicolumn{8}{c}{\textbf{Number of sampled windows for threshold estimation $n_{th}$}} \\ \toprule
\multirow{3}{*}{1.1} 
 & $n_{th} = 1k$   & 1.00 & 1.00 & 1.00 & 1.00 & 1.00 & \textbf{\textit{1.00}} \\
 & $n_{th} = 5k$   & 0.99 & 0.99 & 1.00 & 1.00 & 1.00 & \textit{0.99} \\
 & $n_{th} \in \{100, \underline{10k}, 25k\}$  & 0.99 & 1.00 & 1.00 & 1.00 & 1.00 & \textit{0.99} \\
\midrule
\multirow{2}{*}{7.1}                
 & $n_{th} = 1k$   & 0.93 & 1.00 & 1.00 & 1.00 & 1.00 & \textit{0.96} \\
 & $n_{th} \in \{5k, \underline{10k}, 15k, 20k\}$   & 0.96 & 1.00 & 1.00 & 1.00 & 1.00 & \textbf{\textit{0.98}} \\
 \bottomrule  \multicolumn{8}{c}{\textbf{Threshold sensitivity $T_\alpha$}} \\ \toprule
 \multirow{5}{*}{1.1} 
 & $T_\alpha = 0.00$  & 1.00  & 0.82 & 1.00 & 1.00 & 1.00 & \textit{0.97} \\
 & $T_\alpha = \underline{0.01}$  & 0.99  & 1.00 & 1.00 & 1.00 & 1.00 & \textbf{\textit{0.99}} \\
 & $T_\alpha = 0.05$  & 0.95  & 1.00 & 1.00 & 1.00 & 1.00 & \textit{0.97} \\
 & $T_\alpha = 0.10$  & 0.90  & 1.00 & 1.00 & 1.00 & 1.00 & \textit{0.95} \\
 & $T_\alpha = 0.25$  & 0.75  & 1.00 & 1.00 & 1.00 & 1.00 & \textit{0.86} \\
\midrule
 \multirow{5}{*}{7.1} 
 & $T_\alpha = 0.00$  & 1.00  & 1.00 & 1.00 & 1.00 &  1.00 & \textbf{\textit{1.00}} \\
 & $T_\alpha = \underline{0.01}$  & 0.97  & 1.00 & 1.00 & 1.00 & 1.00 & \textit{0.98} \\
 & $T_\alpha = 0.05$  & 0.90  & 1.00 & 1.00 & 1.00 & 1.00 & \textit{0.95} \\
 & $T_\alpha = 0.10$  & 0.83  & 1.00 & 1.00 & 1.00 & 1.00 & \textit{0.91} \\
 & $T_\alpha = 0.25$  & 0.68  & 1.00 & 1.00 & 1.00 & 1.00 & \textit{0.81} \\
 \bottomrule  \multicolumn{8}{c}{\textbf{Number of principal components $d'$}} \\ \toprule
 \multirow{3}{*}{1.1} 
 & $d' = 50$ & 0.97  & 1.00 & 1.00 & 1.00 & 1.00  & \textit{0.98} \\
 & $d' = 100 $  & 0.99 & 0.99 & 1.00 & 1.00 & 1.00 & \textbf{\textit{0.99}}  \\
 & $d' \in \{\underline{150}, 200, 250\}$  & 0.99 & 1.00 & 1.00 & 1.00 & 1.00 & \textbf{\textit{0.99}}  \\
 \midrule
 \multirow{2}{*}{7.1} 
 & $d' = 50$ & 0.95  & 1.00 & 1.00 & 1.00 & 1.00 &\textit{0.97} \\
 & $d' \in \{100, \underline{150}, 200, 250\}$  & 0.97 & 1.00 & 1.00 & 1.00 & 1.00 & \textbf{\textit{0.98}} \\
\bottomrule  \multicolumn{8}{c}{\textbf{Reference window size $m_b$}} \\ \toprule
\multirow{2}{*}{1.1} 
 & $w_b\% \in \{20\%, 40\%, 60\% \}$ & 0.97  & 1.00 & 1.00 & 1.00 & 1.00  & \textit{0.98} \\
 & $w_b\% \in \{80\%, \underline{100\%} \}$  & 0.99 & 1.00 & 1.00 & 1.00 & 1.00 & \textbf{\textit{0.99}}  \\ \midrule
  \multirow{5}{*}{7.1} 
 & $w_b\% = 20\%$ & 0.58  & 1.00 & 1.00 & 1.00 & 1.00  & \textit{0.73} \\
 & $w_b\% = 40\%$ & 0.68  & 1.00 & 1.00 & 1.00 & 1.00  & \textit{0.81} \\ 
  & $w_b\% = 60\%$ & 0.72  & 1.00 & 1.00 & 1.00 & 1.00  & \textit{0.84} \\
   & $w_b\% = 80\%$ & 0.72  & 1.00 & 1.00 & 1.00 & 1.00  & \textit{0.84} \\
    & $w_b\% = \underline{100\%}$ & 0.97  & 1.00 & 1.00 & 1.00 & 1.00  & \textbf{\textit{0.98}} \\ \bottomrule
\end{tabular}
}
\end{table}

\refstepcounter{section}
\section*{\thesection. Ablation Study: Distribution Distance Metrics}
\label{apx:evaluation-distance-metric}
This evaluation conducts an ablation study on the distribution distance metric used in \XXX.

\begin{table*}[ht]
\scriptsize
\centering

\caption{\textbf{Distribution distance metric ablation study.} Drift detection performance of \XXX\ by varying the distribution distance or divergence metric used. The default metric is the Fréchet distance \cite{DOWSON1982450}, which is compared against the Kullback-Leibler divergence~\cite{kullback1951information}, Jensen-Shannon divergence ~\cite{61115}, Mahalanobis distance~\cite{noauthor_2018-zc}; and Bhattacharyya~distance~\cite{3142ae09-8e70-3b5c-a340-fa8eafc77ee5}. For each metric and window size $m_w$ are reported (i) the \textit{accuracy}, separately per drift percentage $D_{\%}$, and (ii) the $H_{DD}$. 
The best-performing metric for each use case based on the overall $H_{DD}$ is highlighted, and per window size is in~bold. }
\label{tab:experiments-drift-metric-ablation}
\centering
\scalebox{.90}{
\begin{tabular}{ccllllllllllllll}
\toprule

\multirow{5}{*}{\textbf{Use}} & \multirow{6}{*}{\textbf{Metric}} & \multicolumn{7}{c}{\textit{$m_w = 500$}} & \multicolumn{1}{c}{\textit{}} & \multicolumn{6}{c}{\textit{$m_w = 1000$}}   \\ \cline{4-9} \cline{11-16} 
 & &  & & & & & & & & & & &  \\
 &  &  & \multicolumn{5}{c}{\textbf{Drift Percentage $D_{\%}$}} &  & \multicolumn{1}{c}{\textbf{}} &  \multicolumn{5}{c}{\textbf{Drift Percentage $D_{\%}$}} & \multicolumn{1}{c}{\textbf{}}  \\
 \textbf{Case} &  & & 0\% & 5\% & 10\% & 15\% & 20\% & $H_{DD}$ & & 0\% & 5\% & 10\% & 15\% & 20\% & $H_{DD}$  \\ \cline{4-8} \cline{11-15} 
 & & & & & & & & & & & & & &  \\ \toprule
\multirow{2}{*}{1.1 } 
 & \multicolumn{1}{l}{Kullback-Leibler Divergence} & & 0.98 & 0.00 & 0.00 & 0.00 & 0.00 & \textit{0.00} & & 0.98 & 0.00 & 0.00 & 0.03 &  0.39 & \textit{0.19}  \\
 & \multicolumn{1}{l}{Jensen-Shannon Divergence} & & 0.99 & 0.03 & 0.14 & 0.46 & 0.92 & \textit{0.56} & & 1.00 & 0.43 & 1.00 & 1.00 & 1.00 & \textit{0.92}  \\
Ag News & \multicolumn{1}{l}{Mahalanobis Distance} & & 0.98 & 0.61 & 1.00 & 1.00 & 1.00 & \textit{0.94} & & 0.97 & 0.96 & 1.00 & 1.00 &   1.00 & \textit{0.99}  \\
\multirow{2}{*}{BERT} & \multicolumn{1}{l}{Bhattacharyya Distance} & & 0.99 & 0.02 & 0.09 & 0.44 & 0.88 & \textit{0.52} & & 1.00 & 0.40 & 1.00 & 1.00 & 1.00 & \textit{0.92}  \\
 & \multicolumn{1}{l}{\cellcolor{mycellcolor}Frechét Distance} & \cellcolor{mycellcolor} & \cellcolor{mycellcolor}0.99 & \cellcolor{mycellcolor}0.83 & \cellcolor{mycellcolor}1.00 & \cellcolor{mycellcolor}1.00 & \cellcolor{mycellcolor}1.00 & \cellcolor{mycellcolor}\textit{\textbf{0.97}} & \cellcolor{mycellcolor} & \cellcolor{mycellcolor}1.00 & \cellcolor{mycellcolor}0.98 & \cellcolor{mycellcolor}1.00 & \cellcolor{mycellcolor}1.00 &   \cellcolor{mycellcolor}1.00 & \cellcolor{mycellcolor}\textit{\textbf{1.00}}  \\  \midrule
\multirow{2}{*}{2.1 } 
 & \multicolumn{1}{l}{Kullback-Leibler Divergence} & & 0.61 & 0.05 & 0.00 & 0.00 & 0.00 & \textit{0.03} & & 0.00 & 1.00 & 1.00 & 1.00 &  1.00 & \textit{0.00}  \\
 & \multicolumn{1}{l}{Jensen-Shannon Divergence} & & 0.01 & 1.00 & 1.00 & 1.00 & 1.00 & \textit{0.00} & & 0.00 & 1.00 & 1.00 & 1.00 &   1.00 & \textit{0.00}  \\
20Newsgroup & \multicolumn{1}{l}{Mahalanobis Distance} & & 0.00 & 1.00 & 1.00 & 1.00 & 1.00 & \textit{0.00} & & 0.00 & 1.00 & 1.00 & 1.00 &   1.00 & \textit{0.00}  \\
\multirow{2}{*}{BERT} & \multicolumn{1}{l}{Bhattacharyya Distance} & & 0.00 & 1.00 & 1.00 & 1.00 & 1.00 & \textit{0.00} & & 0.00 & 1.00 & 1.00 & 1.00 &   1.00 & \textit{0.00}  \\
 & \multicolumn{1}{l}{\cellcolor{mycellcolor}Frechét Distance} & \cellcolor{mycellcolor} & \cellcolor{mycellcolor}0.89 & \cellcolor{mycellcolor}0.42 & \cellcolor{mycellcolor}0.92 & \cellcolor{mycellcolor}1.00 & \cellcolor{mycellcolor}1.00 & \cellcolor{mycellcolor}\textbf{\textit{0.86}} & \cellcolor{mycellcolor} & \cellcolor{mycellcolor}0.84 & \cellcolor{mycellcolor}0.78 & \cellcolor{mycellcolor}1.00 & \cellcolor{mycellcolor}1.00 & \cellcolor{mycellcolor}1.00 & \cellcolor{mycellcolor}\textbf{\textit{0.89}}  \\  \midrule
\multirow{2}{*}{4} 
 & \multicolumn{1}{l}{Kullback-Leibler Divergence} & & 0.98 & 0.00 & 0.06 & 0.13 & 0.37 & \textit{0.25} & & 0.99 & 0.04 & 0.30 & 0.77 &  0.98 & \textit{0.69}  \\
 & \multicolumn{1}{l}{Jensen-Shannon Divergence} & & 0.98 & 0.06 & 0.43 & 0.86 & 0.99 & \textit{0.73} & & 0.97 & 0.30 & 0.97 & 1.00 & 1.00 & \textit{0.89}  \\
Bias in Bios & \multicolumn{1}{l}{Mahalanobis Distance} & & 1.00 & 0.15 & 0.88 & 1.00 & 1.00 & \textit{0.86} & & 0.98 & 0.43 & 1.00 & 1.00 &   1.00 & \textit{0.91}  \\
\multirow{2}{*}{BERT} & \multicolumn{1}{l}{Bhattacharyya Distance} & & 0.99 & 0.15 & 0.88 & 1.00 & 1.00 & \textit{0.86} & & 0.96 & 0.28 & 0.98 & 1.00 & 1.00 & \textit{0.88}  \\
 & \multicolumn{1}{l}{\cellcolor{mycellcolor}Frechét Distance} & \cellcolor{mycellcolor} & \cellcolor{mycellcolor}0.97 & \cellcolor{mycellcolor}0.53 & \cellcolor{mycellcolor}1.00 & \cellcolor{mycellcolor}1.00 & \cellcolor{mycellcolor}1.00 & \cellcolor{mycellcolor}\textit{\textbf{0.93}} & \cellcolor{mycellcolor} & \cellcolor{mycellcolor}0.98 & \cellcolor{mycellcolor}0.96 & \cellcolor{mycellcolor}1.00 & \cellcolor{mycellcolor}1.00 &   \cellcolor{mycellcolor}1.00 & \cellcolor{mycellcolor}\textit{\textbf{0.99}}  \\  \midrule
 \multirow{2}{*}{6.1} 
 & \multicolumn{1}{l}{Kullback-Leibler Divergence} & & 0.90 & 0.00 & 0.00 & 0.00 & 0.00 & \textit{0.00} & & 0.84 & 0.00 & 0.00 & 0.00 &  0.02 & \textit{0.01} \\
 & \multicolumn{1}{l}{Jensen-Shannon Divergence} & & 0.91 & 0.04 & 0.07 & 0.28 & 0.84 & \textit{0.46} & & 0.82 & 0.37 & 1.00 & 1.00 & 1.00 & \textit{0.83}  \\
Intel Image & \multicolumn{1}{l}{\cellcolor{mycellcolor}Mahalanobis Distance} & \cellcolor{mycellcolor} & \cellcolor{mycellcolor}1.00 & \cellcolor{mycellcolor}0.88 & \cellcolor{mycellcolor}1.00 & \cellcolor{mycellcolor}1.00 & \cellcolor{mycellcolor}1.00 & \cellcolor{mycellcolor}\textbf{\textit{0.98}} & \cellcolor{mycellcolor} & \cellcolor{mycellcolor}1.00 & \cellcolor{mycellcolor}1.00 & \cellcolor{mycellcolor}1.00 & \cellcolor{mycellcolor}1.00 &   \cellcolor{mycellcolor}1.00 & \cellcolor{mycellcolor}\textbf{\textit{1.00}}  \\
\multirow{2}{*}{ViT} & \multicolumn{1}{l}{Bhattacharyya Distance} & & 0.90 & 0.03 & 0.05 & 0.29 & 0.85 & \textit{0.45} & & 0.86 & 0.28 & 1.00 & 1.00 & 1.00 & \textit{0.84}  \\
 & \multicolumn{1}{l}{Frechét Distance} &  & 0.96 & 0.57 & 1.00 & 1.00 &   1.00 & \textit{0.93} &  & 0.96 & 0.75 & 1.00 & 1.00 & 1.00 &  \textit{0.95}  \\  \midrule
  \multirow{2}{*}{7.1} 
 & \multicolumn{1}{l}{Kullback-Leibler Divergence} & & 1.00 & 0.01 & 0.05 & 0.34 & 0.87 & \textit{0.48} & & 0.99 & 0.33 & 1.00 & 1.00 &  1.00 & \textit{0.91}  \\
 & \multicolumn{1}{l}{Jensen-Shannon Divergence} & & 0.98 & 0.30 & 0.96 & 1.00 & 1.00 & \textit{0.89} & & 0.98 & 1.00 & 1.00 & 1.00 & 1.00 & \textbf{\textit{0.99}}  \\
STL-10 & \multicolumn{1}{l}{Mahalanobis Distance} & & 0.96 & 1.00 & 1.00 & 1.00 & 1.00 & \textit{0.97} & & 0.92 & 0.43 & 1.00 & 1.00 &   1.00 & \textit{0.96}  \\
\multirow{2}{*}{ViT} & \multicolumn{1}{l}{Bhattacharyya Distance} & & 1.00 & 0.28 & 0.97 & 1.00 & 1.00 & \textit{0.90} & & 0.98 & 0.28 & 0.98 & 1.00 & 1.00 & \textbf{\textit{0.99}}  \\
 & \multicolumn{1}{l}{\cellcolor{mycellcolor}Frechét Distance} & \cellcolor{mycellcolor} & \cellcolor{mycellcolor}0.96 & \cellcolor{mycellcolor}1.00 & \cellcolor{mycellcolor}1.00 & \cellcolor{mycellcolor}1.00 & \cellcolor{mycellcolor}1.00 & \cellcolor{mycellcolor}\textbf{\textit{0.98}} & \cellcolor{mycellcolor} & \cellcolor{mycellcolor}0.98 & \cellcolor{mycellcolor}1.00 & \cellcolor{mycellcolor}1.00 & \cellcolor{mycellcolor}1.00 & \cellcolor{mycellcolor}1.00 & \cellcolor{mycellcolor}\textbf{\textit{0.99}}   \\ \bottomrule

\end{tabular}
}
\end{table*}

\smallskip
\noindent
\textbf{Distribution Distance Metrics} \ \XXX\ assumes that embeddings follow multivariate normal distributions to simplify data modeling. This assumption enables the estimation of historical and window distributions by only computing mean vectors $\mu$ and covariance matrices $\Sigma$. The default distribution distance metric used is the Fréchet distance \cite{DOWSON1982450} (FDD). However, any distance or divergence metric defined for multivariate normal distributions and computable from $\mu$ and $\Sigma$ may be applied. In this evaluation, we assess the performance of \XXX\ by varying the metric used to compute the difference between historical and window distributions. Specifically, we examine the impact of replacing the Fréchet distance with alternative metrics: (1) Kullback-Leibler divergence~\cite{kullback1951information}; (2) Jensen-Shannon divergence ~\cite{61115}; (3) Mahalanobis distance~\cite{noauthor_2018-zc}; and (4) Bhattacharyya~distance~\cite{3142ae09-8e70-3b5c-a340-fa8eafc77ee5}.

\smallskip
\noindent
\textbf{Evaluation Metrics} \ We evaluate the impact on \XXX\ by varying the distribution distance or divergence metric, on two aspects: (1) drift detection \textit{performance},  measured in terms of \textit{accuracy} and $H_{DD}$; and (2) \textit{efficiency}, measured as the running time in seconds. Similarly to the previous experiments, for each drift percentage \( D_{\%} \) and window size \( m_w \), we compute the accuracy by averaging the accuracy over five independent runs with 100 windows each. The \( H_{DD} \) score is then calculated based on these final accuracy values.

\smallskip
\noindent
\textbf{Results} \ Table \ref{tab:experiments-drift-metric-ablation} reports the accuracy and $H_{DD}$ by varying the distribution distance or divergence metric for some of the experimental use cases in Table \ref{tab_uses_cases}. 

The Frechét Distance outperforms the other metrics in 4 out of 5 use cases. In use case 2.1, however, the other metrics are mostly unreliable as they tend to consistently predict the presence of drift. This behavior is likely due to the highly imbalanced reference set in this use case, which may have affected their drift predictions. Despite this, the Mahalanobis and Bhattacharyya distances remain effective alternatives to the Frechét Distance. In these use cases, using the Frechét Distance, \XXX\ achieves an average $H_{DD}$ score of $0.95$, outperforming the implementation with the Mahalanobis Distance ($0.76$), Bhattacharyya Distance ($0.64$),  Jensen-Shannon Divergence ($0.63$), and Kullback-Leibler Divergence ($0.26$).

We also measured the mean running time (in seconds) required to make predictions within a given window. As discussed in \S\ref{subsec:evaluation-complexity}, the running time of \XXX\ is not affected by the size of the reference dataset $m_b$, since only the pre-computed mean embedding vectors and covariance statistics---characterized by the reduced dimensionalities $d'$ (\textit{per-batch}) and $d'_l$ (\textit{per-label})---are used during the \textit{online} drift detection phase. Furthermore, the dimensionality of the window size $m_w$ has minimal impact on running time, as it only influences the time required to reduce the current window’s embeddings to $d'$ and $d'_l$, a step common to all metrics. Therefore, we evaluated the running time of \XXX\ by varying the choice of distribution distance metrics, using a fixed window size of $m_w = 1000$ samples and default values of $d'_l = 150$ and $d'_l = 75$. These reductions were applied to embeddings of original dimensionality $d = 768$, with three possible predicted labels (corresponding to use case 1). Execution times are reported as the average (in seconds) over five independent runs conducted on an Apple M1 MacBook Pro 13 2020 with 16GB of RAM.

We found that the most efficient metric is Mahalanobis Distance, running in $0.02s$, followed by Bhattacharyya Distance ($0.03s$),  Jensen-Shannon Divergence ($0.03s$), Kullback-Leibler Divergence ($0.03s$), and Frechét Distance ($0.06s$). However, all the metrics enable real-time drift detection as their running time is almost negligible ($\leq0.06s$).

\smallskip
\noindent
\textbf{Summary of findings} \ The Fréchet Distance offers the best balance between drift detection performance and computational efficiency. It consistently outperforms other metrics and demonstrates greater reliability across various use cases, with only a negligible increase in runtime. However, when maximizing efficiency is a priority in specific domains, alternative metrics—such as the Mahalanobis Distance, Bhattacharyya Distance, or Jensen-Shannon Divergence—can be considered. In contrast, the Kullback-Leibler Divergence tends to underperform in these scenarios, likely due to the limited size of the data stream window. As a result, \XXX\ exploits the Fréchet Distance as the default metric for measuring distributions distances, while retaining the flexibility for users to choose other options when needed.

\clearpage

\renewcommand{\ttdefault}{cmvtt}

\begin{table*}[h!]
\footnotesize
\centering
\caption{Drift explanation for class \textit{World} in use case 1.1.}
\label{tab:explanations-agnews-world}
\begin{subtable}[t]{\textwidth}
\centering
\caption{\textbf{Historical}: Prototypes generated from historical data.}
\label{tab:explanations-agnews-world-historical}
\scalebox{0.92}{
\begin{tabular}{cp{14.5cm}cc}
\toprule
 \textbf{Cluster} & \textbf{Prototype example:} (Prototype ID) \texttt{``Input Text''} & \textbf{True Label} & \textbf{Pred. Label} \\ \toprule
\multirow{12}{*}{1} & \textbf{(1)} \texttt{ \ ``Nigerian oil flows despite rebel threat-companies Oil should continue to flow from Nigeria, the world seventh largest exporter, despite a rebel threat to attack foreign oil workers in an quot;all-out war quot; due to start on Friday, multinational energy companies said.''} & \parbox[t]{1cm}{\centering World} & World \\ 
\addlinespace[3pt]
  & \textbf{(2)} \texttt{ \ ``Oil exports flow as strike woes ease A general strike in Nigeria, which has raised fears over oil supply from the world seventh-largest exporter, will likely end its first phase on Thursday quot;all going well quot;, union leaders said.''}  & World & World \\ \addlinespace[3pt]
  & \textbf{(3)} \texttt{ \ ``Tough Talks Ahead After EU Is Criticized Efforts to forge the world's largest free trade zone between the European Union and South America's Mercosur economic bloc are unlikely to be concluded by an Oct. 31 deadline, the EU said Thursday, with both sides declaring each other's trade offers insufficient.''}  & World & World \\ 
  \addlinespace[3pt]
  & \textbf{(4)} \texttt{ \ ``Oil Companies In Nigeria Say They Won Give In To Threats Major oil companies operating in Nigeria oil-rich southern region say they will not give in to threats of attacks on their facilities and employees by militias.''}  & World & World \\ 
 \midrule
\multirow{10}{*}{2} & \textbf{(1)} \texttt{ \ ``No progress in N.Korea, Japan talks on abductees : Talks between Japan and North Korea aimed at resolving a dispute over Japanese nationals abducted by the North decades ago ended Sunday without progress, Japanese officials said.''} & World & World \\ 
\addlinespace[3pt]
  & \textbf{(2)} \texttt{ \ ``Iraq PM to address US Congress Iraqi Prime Minister Iyad Allawi is to address a joint session of the US Congress as well as meeting President Bush.''}  & World & World \\
  \addlinespace[3pt]
  & \textbf{(3)} \texttt{ \ ``Final respects paid to Arafat Palestinians pay their last respects to Yasser Arafat after chaotic scenes at his burial in Ramallah.''}  & World & World \\
  \addlinespace[3pt]
  & \textbf{(4)} \texttt{ \ ``Straw: No British troops to Darfur British Foreign Minister Jack Straw said his country does not plan to deploy forces to Darfur in western Sudan but will provide technical assistance.''}  & World & World \\
  \bottomrule

\end{tabular}
}
\end{subtable}

\vspace{1em}

\begin{subtable}[t]{\textwidth}
\centering
\caption{\textbf{Drifted Window:} Prototypes generated from a window where drift occurs.}
\label{tab:explanations-agnews-world-drift}
\scalebox{0.92}{
\begin{tabular}{cp{14.5cm}cc}
\toprule
 \textbf{Cluster} & \textbf{Prototype example:} (Prototype ID) \texttt{``Input Text''} & \textbf{True Label} & \textbf{Pred. Label} \\ \toprule
\multirow{11}{*}{1} & \textbf{(1)} \texttt{ \ ``Britain Straw to Keep World Pressure on Sudan British Foreign Secretary Jack Straw flew to Sudan on Monday to keep up international pressure on Khartoum to comply with UN demands to end the conflict in Darfur that has already killed up to 50,000 people.''} & World & World \\ 
\addlinespace[3pt]
  & \textbf{(2)} \texttt{ \ ``Aid workers evacuated as violence flares in Darfur; at least 17 ... Fighting near a village in Sudan crisis-plagued Darfur region killed at least 17 people Monday, while helicopters rescued dozens of workers who fled into the bush to escape.''}  & World & World \\ \addlinespace[3pt]
  & \textbf{(3)} \texttt{ \ ``Egypt calls for release of hostages in Iraq Egypt on Saturday called for the immediate release of Egyptians and other nationals taken hostage in violence-ravaged Iraq. quot;Egypt is against the practice of hostage-taking.''}  & World & World \\ 
  \addlinespace[3pt]
  & \textbf{(4)} \texttt{ \ ``Israel To Benefit From Sinai Bombings: Experts A cohort of Egyptian security, political and diplomatic experts have concluded that Israel is the only party to benefit from the blasts.''}  & World & World \\ 
 \midrule
\multirow{10}{*}{2} & \textbf{(1)} \texttt{ \ ``Genetic Material May Help Make Nano-Devices: Study (Reuters) Reuters - The genetic building blocks that form the basis for life may also be used to build the tiny machines of nanotechnology, U.S. researchers said on Thursday.''} & \parbox[t]{1cm}{\centering Science/Tech \\ \textbf{(Drift)}} & World \\ 
\addlinespace[3pt]
  & \textbf{(2)} \texttt{ \ ``UK expertise 'at risk from cuts' Gordon Brown's Whitehall cuts risk damaging the UK's ability to deal with key scientific problems, a trade union says.''}  & \parbox[t]{1cm}{\centering Science/Tech \\ \textbf{(Drift)}} & World \\
  \addlinespace[3pt]
  & \textbf{(3)} \texttt{ \ ``DirecTV hacker sentenced to seven years A Canadian man arrested in the U.S. was allegedly responsible for putting 68,000 hacked smart cards on the street.''}  & \parbox[t]{1cm}{\centering Science/Tech \\ \textbf{(Drift)}} & World \\
  \addlinespace[3pt]
  & \textbf{(4)} \texttt{ \ ``Photo gallery: Bill Gates' home on Lake Washington Webshots users offer their photos of Bill Gates mansion in Medina, Wash.''}  & \parbox[t]{1cm}{\centering Science/Tech \\ \textbf{(Drift)}} & World \\
  \bottomrule
\end{tabular}
}
\end{subtable}

\end{table*}

\clearpage

\begin{table*}[h!]
\footnotesize
\centering
\caption{Drift explanation for class \textit{Science} in use case 2.1.}
\label{tab:explanations-20news-science}
\begin{subtable}[t]{\textwidth}
\centering
\caption{\textbf{Historical}: Prototypes generated from historical data.}
\label{tab:explanations-20news-science-historical}
\scalebox{0.85}{
\begin{tabular}{cp{16cm}cc}
\toprule
 \textbf{Cluster} & \textbf{Prototype example:} (Prototype ID) \texttt{``Input Text''} & \textbf{True Label} & \textbf{Pred. Label} \\ \toprule
\multirow{8}{*}{1} & \textbf{(1)} \texttt{ \ ``How hard would it be to somehow interface them to some of the popular  Motorola microcontrollers. I am a novice at microcontrollers but I am starting to get into them for some of my projects. I have several k SIMMs laying around from upgraded Macs and if I could use them as free memory in one or two of my projects that would be great. One project that comes to mind is a Caller ID device that would require quite a bit of RAM to store several hundered CID records etc.''} & \parbox[t]{1cm}{\centering Science} & Science \\ 
\addlinespace[3pt]
  & \textbf{(2)} \texttt{ \ ``I too would be interested in ANY information on the subject of programing PALS etc.....   Better to know what your on about before you start something I always say. Often saves you a packet as well   Thanks in advance..  Chris      Name  Mr Chris Smith  Twang on that ole guitar    Addrs scst    Uni  Liverpool University Quest To build more and more hardware   Dgree Computer Science       What ever the Sun may be it is certainly not a ball of flaming gas   D.H. Lawrence.''}  & Science & Science \\ 
 \midrule
\multirow{10}{*}{2} & \textbf{(1)} \texttt{ \ ``I wonder if she landed such a fat fee from cooperation with the NSA in the design and propoganda stages that she doesnt care any more    Which is to say is the NSA totally perfidious or does it at least   Of course they take care of their own ... very well ... until the person has outlived hisherundefined usefulness... then elimination becomes a consideration...''} & Science & Science \\ 
\addlinespace[3pt]
  & \textbf{(2)} \texttt{ \ ``If the Clinton Clipper is so very good why not make its algrithm public so many people can exchange ideas and examine it rather than a few isolated respected experts respected by whom for what Perhaps a certain professor who likes key banks would be one of the selected experts... this does seem to expand on some ideas the person was advocating if I recall . How would anybody know that what the  Actually I am completely baffled by why Dorothy Denning has chosen to throw away her academic respectability like this. It looks to me like a major Career Limiting Move. There can be very few people who know what shes been saying who take her seriously any more.  I wonder if she landed such a fat fee from cooperation with the NSA in the design and propoganda stages that she doesnt care any more.''}  & Science & Science \\
 \midrule
\multirow{11}{*}{3} & \textbf{(1)} \texttt{ \ ``The NRA is successful because among a number of things on the drop of a hat they can get a congresspersons office flooded with postcards faxes and phone calls. Certainly with our waycool Internet powers of organization we can act in the same way if such action is appropriate.  As long as we are kept informed of events anyone on this bboard can make a call to action. Hopefully were a strong enough community to act on those calls. I realize this is a little optomistic and Im glad EFF is working in the loop on these issues but dont underestimate the potential of the net for political action.''} & Science & Science \\
\addlinespace[3pt]
  & \textbf{(2)} \texttt{ \ ``Generally an organization has influence in proportion to  The narrowness of its objectives The number of members The strength of belief of its members  This is why the pro and antiabortion groups are so strong narrow objectives lots of interested members who are real passionate.  For this reason mixing with the NRA is probably a bad idea. It diffuses the interests of both groups. It may well diminish the Passion Index of the combined organization. It is not clear it would greatly enlarge the NRA.  So I believe a new organization which may cooperate with NRA where the two organizations interest coincide is the optimum strategy.''}  & Science & Science 
 \\ \bottomrule

\end{tabular}
}
\end{subtable}

\vspace{1em}

\begin{subtable}[t]{\textwidth}
\centering
\caption{\textbf{Drifted Window:} Prototypes generated from a window where drift occurs.}
\label{tab:explanations-20news-science-drift}
\scalebox{0.85}{
\begin{tabular}{cp{16cm}cc}
\toprule
 \textbf{Cluster} & \textbf{Prototype example:} (Prototype ID) \texttt{``Input Text''} & \textbf{True Label} & \textbf{Pred. Label} \\ \toprule
\multirow{8}{*}{1} & \textbf{(1)} \texttt{ \ ``I have a couple reasons why I would be more likely to trust this algorithm: 1. The algorithm will be made totally public, once it is patented. 2. The keys will NOT be escrowed.
Of course if either of these is not true, I will not use this new algorithm. Since I have never seen this new algorithm, I have no idea how secure it is yet. I can't make any judgements about the algorithm itself yet, but I do notice that the creators of this algorithm are being more open about how the thing works, and is willing to make it public, showing that after a bit of scrutiny, any weaknesses will probably be revealed, while we don't know about Clipper.''} & Science & Science \\ 
\addlinespace[3pt]
 & \textbf{(2)} \texttt{ \ ``You can try SGS L6217A, it can achieve 256 current level(microstep), teere is a circuit in the SGS-THOMSON - " Smart Power Applicatio
Manual", order code for the manual is AMSMARTPOWERST/1.''}  & Science & Science \\ 
 \midrule
\multirow{17}{*}{2} & \textbf{(1)} \texttt{ \ ``What the hell are U talking about ESPN showed PensDevils game as advertised.  BUt the morons at ESPN should know that Pens will kick Devils ass and the game will be boring. Id rather see BostonBuffalo game which seems to be an exciting series since noone had expected Buffalo to get past the first round. Well lets hope they change their mind on THUs game and show some other game. The Pens series is really getting boring. I want to see some exciting game no matter who wins. If NHL wants a major network contract  then they better put some brains in ESPN people.''} & \parbox[t]{1cm}{\centering Recreation \\ \textbf{(Drift)}} & Science \\ 
\addlinespace[3pt]
 & \textbf{(2)} \texttt{ \ ``Well the jettison for youth fast strategy was his. Also in hindsight it didnt work all that well but I think it was more because it left the Sharks on a tightrope without a net not that it was inherently flawed. It was the injuries that caused us to fall.   From all indications it wasnt the major factor but the last straw. There were personality conflicts among Shark management and disagreements over how personnel were going to be handled not just who was being traded vs. kept but who was being sent to KC. What Ferriera did if rumors are right was not always what the folks behind the bench wanted or needed.   I think if theyd kept Ferriera they would have lost some of their other management staff. Depending on which sources you trust we might have lost Grillo AND Lombardi AND Murdoch over the summer. Frankly I want to keep those three. we may well ahve also lost Kingston which of course is now a moot point.   The Sharks have been building an organizational staff that is highly consensual and cooperative. Ferriera wanted to run things his way. There were conflicts. Ferriera lost. That says nothing about his skills or accomplishments  at that level a lot is personality and politics.   I think he did some good things for the Sharks but that he never fit in as a Shark person. I hope he succeeds beyond his wildest dreams down in Anaheim too because itll be good for hockey. but I want the Ducks to be doormats for a couple of years so the Sharks succeeed first.''}   & \parbox[t]{1cm}{\centering Recreation \\ \textbf{(Drift)}} & Science \\ \bottomrule

\end{tabular}
}
\end{subtable}

\end{table*}

\clearpage

\begin{table*}[h!]
\scriptsize
\centering
\caption{Drift explanation for class \textit{Nurse} in use case 4.}
\label{tab:explanations-biasinbios-nurse}
\begin{subtable}[t]{\textwidth}
\centering
\caption{\textbf{Historical}: Prototypes generated from historical data.}
\label{tab:explanations-biasinbios-nurse-historical}
\scalebox{0.917}{
\begin{tabular}{cp{15cm}cc}
\toprule
 \textbf{Cluster} & \textbf{Prototype example:} (Prototype ID) \texttt{``Input Text''} & \textbf{True Label} & \textbf{Pred. Label} \\ \toprule
\multirow{9}{*}{1} & \textbf{(1)} \texttt{ \ ``She graduated with honors in 1995. Having more than 21 years of diverse experiences, especially in nurse practitioner, Paula M Crehan affiliates with many hospitals including Ellis Hospital, Albany Medical Center Hospital, St Peter's Hospital, Samaritan Hospital, Mary Imogene Bassett Hospital, and cooperates with other doctors and specialists in medical group Capitalcare Medical Group LLC. Call Paula M Crehan on phone number (518) 452-1337 for more information and advises or to book an appointment.''} & \parbox[t]{1cm}{\centering Nurse} & Nurse \\ 
\addlinespace[3pt]
  & \textbf{(2)} \texttt{ \ ``She graduated with honors in 2007. Having more than 10 years of diverse experiences, especially in nurse practitioner, Tracey L Wilkerson affiliates with many hospitals including St Joseph Mercy Oakland, Mclaren Oakland, and cooperates with other doctors and specialists in many medical groups including Independence Urgent Care, Clarkston Internal Medicine, P.c., Hospitialist Service Of Michigan, LLC. Call Tracey L Wilkerson on phone number (586) 263-2601 for more information and advises or to book an appointment.''}  & Nurse & Nurse \\ 
 \midrule
\multirow{4}{*}{2} & \textbf{(1)} \texttt{ \ ``Myofascial release is a gentle and safe hands-on therapy that is helpful in reducing pain and discomfort. Clients see relief from back pain, fibromyalgia, migraines, TMJ, sciatia and other chronic pain issues and it also helps increase flexibility and energy levels.''} & Nurse & Nurse \\ 
\addlinespace[3pt]
  & \textbf{(2)} \texttt{ \ ``Before joining the firm in 2004, Ms. Egerter was a registered nurse at Hahnemann University Hospital and a Medical Information Specialist at GlaxoSmithKline.''}  & Nurse & Nurse 
 \\ \bottomrule

\end{tabular}
}
\end{subtable}

\vspace{1em}

\begin{subtable}[t]{\textwidth}
\centering
\caption{\textbf{Drifted Window:} Prototypes generated from a window where drift occurs.}
\label{tab:explanations-biasinbios-nurse-drift}
\scalebox{0.917}{
\begin{tabular}{cp{15cm}cc}
\toprule
 \textbf{Cluster} & \textbf{Prototype example:} (Prototype ID) \texttt{``Input Text''} & \textbf{True Label} & \textbf{Pred. Label} \\ \toprule
\multirow{9}{*}{1} & \textbf{(1)} \texttt{ \ ``Her research focuses on biodiversity, invasion ecology, plant-soil interactions, environmental change, and threshold dynamics. She has been involved with wide-ranging ecological research, with a particular emphasis on experimental manipulations and quantitative data analysis, since receiving her PhD from the University of Michigan in 1999. She came to UC Berkeley from UC Irvine in 2009, where she was in the Ecology and Evolutionary Biology Department. At Berkeley, she teaches classes in restoration ecology and is a professor of restoration ecology in the UC Agricultural Experimental Station.''} & \parbox[t]{1cm}{\centering Professor \\ \textbf{(Drift)}} & Nurse \\
\addlinespace[3pt]
  & \textbf{(2)} \texttt{ \ ``Her scholarship focuses on developing, implementing, and researching applications of critical sociocultural theory and practices to the preparation of general education teachers of English Language Learners. She has been awarded four U.S. Department of Education grants focused on ESL teacher quality and published numerous articles, curricula, and multimedia products targeting teachers of English Language Learners.''}  & \parbox[t]{1cm}{\centering Professor \\ \textbf{(Drift)}} & Nurse \\ 
 \midrule
\multirow{8}{*}{2} & \textbf{(1)} \texttt{ \ ``As a journalist she writes for a wide range of publications, businesses and web sites. As a travel/tourism writer Angela has an interest in luxury travel, destinations, re-enactments, historic and family travel, adventure/sports, food tourism, dark tourism and much more. Food writing focuses on fresh produce, innovations, vegan/vegetarism, food manufacturing \& supply.''} & \parbox[t]{1cm}{\centering Journalist \\ \textbf{(Drift)}} & Nurse \\ 
\addlinespace[3pt]
  & \textbf{(2)} \texttt{ \ ``She is the manager of the Data Journalism Awards, an international data journalism competition organised by the Global Editors Network. She is also the founder and director of HEI-DA, a data storytelling nonprofit promoting digital innovation, the future of data journalism and open data. Before launching HEI-DA, Marianne spent 10 years in London where she worked as a web producer, data journalism and graphics editor for Bloomberg News. She also created the Data Journalism Blog in 2011 and gives lectures at journalism schools, in the UK and in France.''}  & \parbox[t]{1cm}{\centering Journalist \\ \textbf{(Drift)}} & Nurse \\ \midrule
\multirow{7}{*}{3} & \textbf{(1)} \texttt{ \ ``Since she began photoblogging in 2004 on Flickr under the name, Sesame Ellis, Rachel developed an entire brand of visual storytelling that she share on her blog: sesameellis.com. As half of the team behind Beyond Snapshots (Random House/Amphoto books 2012), she teaches people how to photograph their lives through workshops. Just named Australia's newest Tamron Ambassador, there are bigger plans now to share the joy of daily life photography.''} & \parbox[t]{1cm}{\centering Photographer \\ \textbf{(Drift)}} & Nurse \\ 
\addlinespace[3pt]
 & \textbf{(2)} \texttt{ \ ``Her books On the Sixth Day (2005) and the upcoming The Adventures of Guille and Belinda (2010) are published by Nazraeli Press. She is the recipient of a Guggenheim Fellowship, the National Geographic Magazine Grant for Photography, and the Hasselblad Foundation Grant. A solo exhibition of her work is currently on view at Kulturhuset in Stockholm.''}  & \parbox[t]{1cm}{\centering Photographer \\ \textbf{(Drift)}} & Nurse \\ \midrule
 \addlinespace[3pt]
\multirow{8}{*}{4} & \textbf{(1)} \texttt{ \ ``Ms. West began her career as corporate attorney over ten years ago and eventually opened West Law Firm. She began with one practice area, Bankruptcy Law, and later expanded to include representation of Business, Intellectual Property, Insurance and Personal Injury clients.''} & \parbox[t]{1cm}{\centering Attorney \\ \textbf{(Drift)}} & Nurse \\ 
\addlinespace[3pt]
  & \textbf{(2)} \texttt{ \ ``She has experience working through the various stages of litigation, including researching and writing memoranda on a variety of legal topics, drafting pretrial motions, compiling data, completing document review in preparation for trial, and attending various depositions and pretrial matters. Prior to joining Segal McCambridge as an associate, Ms. Laurel worked as a summer associate and law clerk at the firm. She has additional experience working for the City of Chicago as a law clerk in the Collections, Ownership, and Administrative Litigation Division.''}  & \parbox[t]{1cm}{\centering Attorney \\ \textbf{(Drift)}} & Nurse
\\ \midrule
\multirow{10}{*}{5} & \textbf{(1)} \texttt{ \ ``She graduated with honors in 1998. Having more than 18 years of diverse experiences, especially in nurse practitioner, Nancy G Dagefoerde affiliates with many hospitals including Saint Anthony Medical Center, Swedish American Hospital, Fhn Memorial Hospital, Rochelle Community Hospital, Rockford Memorial Hospital, and cooperates with other doctors and specialists in many medical groups including Cardiovascular Institute At Osf LLC, Osf Multi-specialty Group. Call Nancy G Dagefoerde on phone number (815) 398-3000 for more information and advises or to book an appointment.''} & Nurse & Nurse \\ 
\addlinespace[3pt]
  & \textbf{(2)} \texttt{ \ ``She graduated with honors in 1990. Having more than 27 years of diverse experiences, especially in nurse practitioner, Dorothy L Rhoads affiliates with many hospitals including Abington Memorial Hospital, Lansdale Hospital, Chestnut Hill Hospital, Doylestown Hospital, Holy Redeemer Hospital And Medical Center, and cooperates with other doctors and specialists in medical group Abington Memorial Hospital. Call Dorothy L Rhoads on phone number (215) 366-1160 for more information and advises or to book an appointment.''}  & Nurse & Nurse \\ \bottomrule

\end{tabular}
}
\end{subtable}

\end{table*}

\clearpage

\end{document}